%% file: main.tex
\documentclass[pdflatex,sn-basic]{sn-jnl}% Springer Nature, basic author-year reference style

\usepackage{graphicx}%
\usepackage{multirow}%
\usepackage{amsmath,amssymb,amsfonts}%
\usepackage{amsthm}%
\usepackage{mathrsfs}%
\usepackage[title]{appendix}%
\usepackage{xcolor}%
\usepackage{textcomp}%
\usepackage{manyfoot}%
\usepackage{booktabs}%
\usepackage{algorithm}%
\usepackage{algorithmicx}%
\usepackage{algpseudocode}%
\usepackage{bm}%

\theoremstyle{thmstyleone}%
\newtheorem{theorem}{Theorem}%
\newtheorem{proposition}{Proposition}%
\newtheorem{lemma}{Lemma}%
\newtheorem{corollary}{Corollary}%

\theoremstyle{thmstyletwo}%
\newtheorem{example}{Example}%
\newtheorem{remark}{Remark}%

\theoremstyle{thmstylethree}%
\newtheorem{definition}{Definition}%

\providecommand{\mathbold}[1]{\bm{#1}}
\providecommand{\Upsigma}{\mathrm{\Sigma}}

\makeatletter
\gdef\orcidlogo{\textsuperscript{\scriptsize ORCID}}
\newsavebox{\fitsb}
\newcommand{\fitbox}[1]{\sbox{\fitsb}{#1}\ifdim\wd\fitsb>\textwidth\resizebox{\textwidth}{!}{\usebox{\fitsb}}\else\usebox{\fitsb}\fi}
\makeatother

\begin{document}

\title[A Temporal Algebraic Framework for Data Stream Evaluation]{A Temporal Algebraic Framework for Data Stream Evaluation Under Delayed Labels and Distribution Shifts}

\author*[1]{\fnm{Ronaldo C.} \sur{Prati}\orcid{https://orcid.org/0000-0001-8597-4987}}\email{ronaldo.prati@ufabc.edu.br}

\author[2]{\fnm{Elaine Ribeiro} \spfx{de} \sur{Faria}\orcid{https://orcid.org/0000-0001-5242-9026}}\email{elaine@ufu.br}

\affil*[1]{\orgdiv{Center for Mathematics, Computing and Cognition (CMCC)}, \orgname{Universidade Federal do ABC (UFABC)}, \orgaddress{\city{Santo Andr\'e}, \state{SP}, \country{Brazil}}}

\affil[2]{\orgdiv{Faculdade de Computa\c{c}\~ao (FACOM)}, \orgname{Universidade Federal de Uberl\^andia (UFU)}, \orgaddress{\city{Uberl\^andia}, \state{MG}, \country{Brazil}}}

\abstract{In data streams, labels often arrive late or only for some instances, class prevalence drifts, and rare classes vanish from short windows. Standard evaluation then confuses a degraded decision boundary with a change in prevalence, and is biased when labels are delayed or selected by the model. We extend an algebraic framework that obtains pooled (micro) and per-class (macro) measures from one confusion matrix to streams, weighting the counts in time by exponential fading or a sliding window; the update costs $O(m^2)$ per step, whatever the stream length, and each pooled measure stays a weighted combination of the per-class ones. Delayed and selective feedback enter as an observation mask: until its label arrives, an instance is counted with the model's calibrated class probabilities, and the bias of this interim estimate is driven by their calibration error. Audited conventions report no value for a class without verified instances. We separate prior shift from concept drift by monitoring prior-free coordinates (recall and specificity) and attribute a change in a measure to the two sources with a first-order Jacobian decomposition. On synthetic streams and four real datasets, a detector on these coordinates with a memory per class raises no excess false alarms under label shift, unlike the Jacobian alarm and error-rate detectors. The correction for delayed labels recovers recall in some real settings and is only partial for precision, specificity and $F_1$. Twenty-five statements are checked in Lean~4, and the code is in the \texttt{aggmetrics} package.}


\maketitle

\section{Introduction}
\label{sec:intro}

Models deployed on data streams process data at high velocity while the distribution changes, and the evaluation protocol is what tells an adaptive system whether the model is still healthy. A protocol that cannot separate a degraded decision boundary from a transient shift can trigger needless adaptation or hide a real failure.

Classification evaluation was developed for batch settings with independent and identically distributed observations, complete and immediate labels, and every class present in each evaluation sample. Three of these assumptions break in streams.

First, streams need continuous temporal aggregation. Under the prequential (test-then-train) paradigm, predictions are made sequentially and the history is discounted by exponential fading ($\alpha$) or a sliding window ($W$). Standard packages~\citep{bifet2010moa,montiel2018scikitmultiflow,montiel2021river} compute streaming metrics as ad hoc functions, with no guarantee on how temporal weighting interacts with micro- and macro-averaging across classes, and embedded applications bound the cost of an update and the memory they may use.

Second, labels are not available immediately. Predictions $\mathbold{p}_t$ are made at time $t$ but the true labels $\mathbold{y}_t$ arrive after an uncertain latency $\Delta_t$ (a fraudulent chargeback is confirmed after weeks), and in credit underwriting or medical screening labels are revealed only for accepted candidates, so the model's own decisions censor the rest. Evaluating only the verified or uncensored instances can then be biased.

Third, diagnostics under distribution shift are unreliable. Accuracy, Precision and $F_1$ depend on the class prior $P_t(Y)$, so a change in class prevalence moves them even when the decision boundary $P_t(Y \mid X)$ is intact and can make drift detectors such as ADWIN~\citep{bifet2007adwin} or DDM~\citep{gama2004learning} raise spurious alarms. Under extreme imbalance, a short window may contain no instance of a rare class, which leaves recall and related measures undefined ($0/0$, a zero-division void), and the default conventions ($0/0 \rightarrow 0.0$ or $1.0$) create artificial steps in monitoring dashboards.

\begin{figure}[!ht]
\centering
\includegraphics[width=\textwidth]{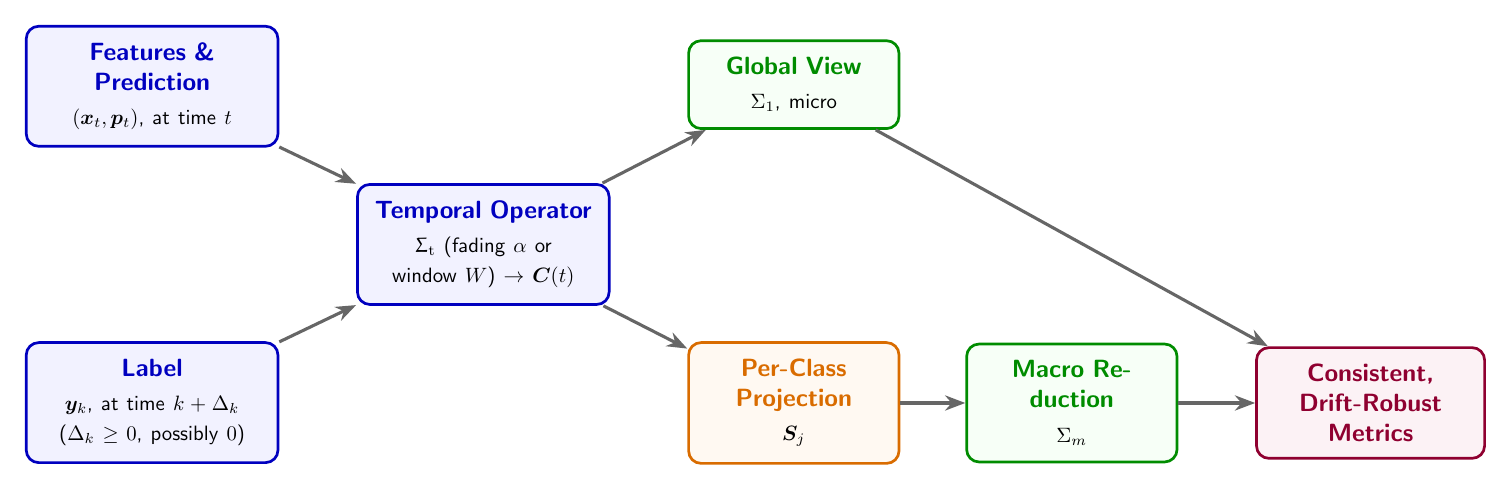}
\caption{Pipeline of the framework. Features and predictions arrive at time $t$, and each label arrives separately at time $k+\Delta_k$ with latency $\Delta_k \ge 0$. Both flows feed the temporal operator $\Upsigma_{\mathrm{t}}$, which keeps one evolving state $\mathbold{C}(t)$. The global view reduces $\mathbold{C}(t)$ directly, and the per-class view first extracts each class's table and then averages across classes. Figure~\ref{fig:hierarchy} gives the algebraic architecture.}
\label{fig:overview}
\end{figure}

This paper extends the unified algebraic framework of indicator matrices \citep{prati2026unified} to continuous, resource-constrained and partially verified data streams. That framework builds a confusion matrix from indicator matrices of the true and the predicted classes, and obtains pooled (micro) and per-class (macro) performance measures from it by two reductions (\S\ref{sec:temporal_im} recalls them). Figure~\ref{fig:overview} shows the pipeline and Algorithm~\ref{alg:update} in \S\ref{sec:product_tnorm} gives its update step by step.

Our contributions are as follows.
\begin{enumerate}
    \item A temporal reduction operator $\Upsigma_{\mathrm{t}}$ that weights the confusion counts in time, in exponential-fading and sliding-window forms. It updates in $O(m^2)$ operations per step independent of the stream length, and each pooled measure remains a weighted combination of the per-class measures (Theorem~\ref{thm:temporal_decomposability}).
    \item A model of delayed and selective feedback as an observation mask, which contains the four label-availability settings of \citet{gomes2022survey}, and an interim estimate of missing labels, which counts a pending instance with the model's calibrated class probabilities combined by the product $t$-norm, with a queue bounded by an analytic bound (Algorithm~\ref{alg:update}) and a bias that follows the calibration error of those probabilities.
    \item Prior-free canonical coordinates (recall and specificity for two classes, the confusion rows for several), with the conditions under which they are prior-free in the binary and the multiclass case, and a scheme that detects on these coordinates and attributes a metric change by a first-order Jacobian decomposition, evaluated on synthetic and real streams with thresholds calibrated for the run-level false alarm.
    \item Audited conventions for undefined (zero-division) cases, a formula for the probability of an undefined window, and Pareto dominance over time for comparing online learners without choosing metric weights.
\end{enumerate}

Section~\ref{sec:related} reviews related work. Section~\ref{sec:temporal_im} introduces the temporal operator, per-class projections, complexity bounds and zero-division auditing. Section~\ref{sec:delayed_labels} presents observation masks for latency and censoring and the interim estimate. Section~\ref{sec:drift_decomposition} treats prior shift and concept drift: prior-free coordinates, attribution and monitoring. Section~\ref{sec:experiments} reports the experiments and Section~\ref{sec:conclusions} concludes. Examples~\ref{ex:running} to~\ref{ex:jacobian} follow one small stream through the sections, and Appendix~\ref{app:running} lists their values. Appendix~\ref{app:notation} collects the notation.

%% ---------------------------------------------------------------------------
\section{Related Work and Background}
\label{sec:related}

Prequential error estimation, pioneered by Dawid \citep{dawid1984present} and popularized in streaming literature by Gama et al. \citep{gama2014survey} and Bifet et al. \citep{bifet2010moa}, evaluates streaming models by testing on incoming instances prior to model updating. Exponential fading factors $\alpha \in (0, 1)$ and sliding windows $W$ are widely used and implemented in standard streaming toolkits such as MOA \citep{bifet2010moa}, scikit-multiflow \citep{montiel2018scikitmultiflow}, and River \citep{montiel2021river}, but whether temporal weighting stays consistent with pooling the classes (micro) and averaging over them (macro) has remained unformalized: these packages compute streaming metrics as empirical, per-metric functions rather than as instances of one aggregation rule.

A related strand of work characterizes and detects the non-stationarity such packages are meant to track. Distinguishing covariate shift $P(X)$, prior probability shift $P(Y)$ and concept drift $P(Y \mid X)$ is an established problem \citep{lu2018learning,webb2016characterizing,morenotorres2012unifying}. The detectors built to monitor drift (DDM~\citep{gama2004learning}, ADWIN~\citep{bifet2007adwin}, HDDM~\citep{friasblanco2015hddm}) track a scalar error rate, which moves with the class prior whenever the classifier's errors differ between classes, and label-shift methods address the prior itself. What \S\ref{sec:drift_decomposition} adds is a monitor on prior-free coordinates and an attribution of the change of a given metric to its prior-dependent and concept-dependent parts. A complementary line of work targets a different question given a known ground-truth drift point: how accurately and promptly a detector fires. \citet{cerqueira2026framework} recently formalized this with an acceptable detection window and timing-aware metrics (F1 detection score, Normalized Detection Time); their framework evaluates detector timing once a drift signal exists, while \S\ref{sec:drift_decomposition} concerns the signal itself and keeps it free of prior shift.

A different failure mode arises even before drift needs to be detected at all: ground truth itself may not be available in time. In practical stream mining, immediate label feedback is an unrealistic assumption. \citet{gomes2022survey} survey semi-supervised learning for delayed, partially labelled streams and state the problem in four settings (immediate or delayed labels, fully or partially labelled), noting that the last, delayed and partially labelled, is the least explored; on the evaluation side they review benchmarking practice and propose a label-removal protocol, and they note that prequential evaluation of a partially labelled stream computes the loss only where labels are known, which is the naive estimator whose selection bias \S\ref{sec:product_tnorm} corrects. \citet{grzenda2020delayed} propose continuous re-evaluation for delayed labels, which follows the predictions made for an instance over time until its label arrives and assumes that every label arrives after a finite delay; \citet{gomes2022survey} note that it could be adapted to partially labelled streams by using only the instances whose label arrives. \citet{plasse2016handling} study the handling of delayed labels in learning, and \citet{zliobaite2014active} address the complementary problem of \emph{where} to spend a limited labelling budget under drift, without formalising the selection bias when evaluation is restricted to the actively queried or decision-accepted subset. What we add is an estimate of the current performance from the unverified instances and a correction of selection bias under decision-dependent censoring.

Rare-class evaluation compounds the same problem: online class imbalance learning \citep{wang2018systematic} and the broader survey of imbalanced-stream methods by \citet{aguiar2024survey} both document how existing algorithms and metrics degrade under skewed, drifting class ratios, but neither formalizes the zero-division singularities that arise when a rare class disappears from a short evaluation window, the failure mode addressed in \S\ref{sec:zero_division}.

A third, largely orthogonal limitation concerns the metrics themselves, independent of how or when labels arrive. Beyond the measures covered by Theorem~\ref{thm:temporal_decomposability}, which are ratios of linear combinations of confusion-matrix cells (recall, precision, $F_1$ and similar), ranking-based metrics pose their own streaming challenges. \citet{brzezinski2017prequential} extend AUC to the prequential setting via an efficient sliding-window sorted-tree structure, addressing for that one metric the same consistency gap this paper generalizes to the whole family of such ratios under $\Upsigma_{\mathrm{t}}$. Model comparison under conflicting objectives has a parallel history in the Pareto-optimal (ROC convex hull) classifiers of \citet{provost2001robust}; \S\ref{sec:exp4} extends this static, batch notion of Pareto dominance to continuously-updated temporal contingency states.

%% ---------------------------------------------------------------------------
\section{Temporal Indicator Matrices and the $\Upsigma_{\mathrm{t}}$ Operator}
\label{sec:temporal_im}

This section establishes the formal foundations of prequential indicator matrices by unifying temporal evaluation with the matrix aggregation algebra of the core framework \citep{prati2026unified}. We express temporal aggregation as a weighted bilinear matrix operator, formalize multiclass per-class projections via linear selection matrices, and define the composition algebra between the new temporal operator $\Upsigma_{\mathrm{t}}$ and the core reduction operators $(\Sigma_1, \Sigma_m)$.

For readers unfamiliar with \citet{prati2026unified}, we briefly recall the two objects this paper
extends. The core framework represents a batch of $n$ classified instances as a pair of indicator
matrices $\mathbold{Y}, \hat{\mathbold{Y}} \in \mathbb{R}^{n \times m}$ (one row per instance, one
column per class), with $\mathbold{C} = \mathbold{Y}^T\hat{\mathbold{Y}} \in \mathbb{R}^{m \times m}$
the resulting confusion matrix. A performance measure $M$ is then evaluated in one of two ways: the
\emph{micro} reduction $\Sigma_1$ pools all $nm$ cell contributions into a single global confusion
table before evaluating $M$ once, while the \emph{macro} reduction $\Sigma_m$ evaluates $M$
separately on each class's own $2\times2$ table and averages the $m$ results. The two coincide
for every choice of the per-class numerators exactly when every class's denominator is equal (Corollary~1 of \citealp{prati2026unified}), and
generally differ otherwise. The same micro/macro distinction reappears, with a temporal
twist, in \S\ref{sec:composition} below. This paper's contribution is to show that both reductions,
and the relationship between them, survive once the batch matrices
$\mathbold{Y}, \hat{\mathbold{Y}}$ are replaced by the temporally weighted state $\mathbold{C}(t)$
defined next.

\subsection{Notation and the Streaming Bilinear Formulation}

Let $\mathcal{S} = \{ (\mathbold{x}_t, \mathbold{y}_t, \mathbold{p}_t) \}_{t \in \mathbb{N}^+}$ represent an ordered sequence of streaming instances arriving at discrete timestamps $t \in \{1, 2, \dots\}$, where $\mathbold{y}_t$ denotes instance $t$'s true label as soon as it exists, whether or not it is yet observable to the evaluator, subject to a possibly random verification latency; \S\ref{sec:delayed_labels} formalizes this observability gap with the per-instance mask $\mathcal{O}_k(t)$ and its latency $\Delta_k$. At time $t$:
\begin{itemize}
    \item $\mathbold{x}_t \in \mathcal{X}$ denotes the input feature vector drawn from dynamic distribution $P_t(X)$;
    \item $\mathbold{y}_t = [y_{t1}, \dots, y_{tm}]^T \in \Delta^{m-1}$ is the true target vector on the standard probability simplex $\Delta^{m-1} = \{ \mathbold{u} \in [0, 1]^m : \sum_{j=1}^m u_j = 1 \}$, representing deterministic labels ($\mathbold{y}_t = \mathbold{e}_j$, the $j$-th standard basis vector, i.e.\ a one-hot vector with a $1$ in the true-class position and $0$ elsewhere) or soft distributions $P_t(Y \mid \mathbold{x}_t)$;
    \item $\mathbold{p}_t = [p_{t1}, \dots, p_{tm}]^T \in \Delta^{m-1}$ is the model output, representing discrete decisions ($\mathbold{p}_t = \mathbold{e}_k$, the same one-hot construction as $\mathbold{e}_j$ above, but with the $1$ in the \emph{predicted}-class position $k$) or calibrated posterior probabilities.
\end{itemize}

Let $\mathbold{Y}_{1:t} = [\mathbold{y}_1, \dots, \mathbold{y}_t]^T \in \mathbb{R}^{t \times m}$ and $\mathbold{P}_{1:t} = [\mathbold{p}_1, \dots, \mathbold{p}_t]^T \in \mathbb{R}^{t \times m}$ denote the accumulated label and prediction matrices up to time $t$. In stationary batch evaluation, the sample reduction operator aggregates instances uniformly via $\mathbold{C} = \mathbold{Y}^T \mathbold{P} = \sum_{i=1}^n \mathbold{y}_i \mathbold{p}_i^T$.

In non-stationary streams, we formalize temporal aggregation through a diagonal temporal weighting matrix $\mathbold{W}(t) = \operatorname{diag}(w_1(t), w_2(t), \dots, w_t(t)) \in \mathbb{R}^{t \times t}$ with $\mathbold{W}(t) \ge 0$ (i.e., $w_k(t) \ge 0$), normalized such that $\operatorname{Tr}(\mathbold{W}(t)) = \sum_{k=1}^t w_k(t) = 1$ (exactly for the sliding window once it has filled, $t \ge W$, with $W$ the window size, a scalar that is not the weighting matrix $\mathbold{W}(t)$; the exponential-fading realization below satisfies it asymptotically, $\operatorname{Tr}(\mathbold{W}(t)) = 1 - \alpha^t \to 1$, and the residual mass $\alpha^t$ is accounted for in Proposition~\ref{prop:variance_scaling}).

\begin{definition}[The Temporal Reduction Operator $\Upsigma_{\mathrm{t}}$]
The temporal reduction operator $\Upsigma_{\mathrm{t}}$ maps the historical matrix trajectory $(\mathbold{Y}_{1:t}, \mathbold{P}_{1:t})$ into the time-aggregated contingency matrix $\mathbold{C}(t) \in \mathbb{R}^{m \times m}$ (with $\mathbold{C}(t) \ge 0$) via the weighted bilinear form:
\begin{equation}
\label{eq:bilinear_temporal}
\mathbold{C}(t) = \Upsigma_{\mathrm{t}}(\mathbold{Y}_{1:t}, \mathbold{P}_{1:t}) = \mathbold{Y}_{1:t}^T \, \mathbold{W}(t) \, \mathbold{P}_{1:t} = \sum_{k=1}^t w_k(t) \mathbold{y}_k \mathbold{p}_k^T
\end{equation}
\end{definition}

In classification evaluation, performance metrics are computed by reducing data along specific dimensions or axes. In the static indicator matrix framework \citep{prati2026unified}, three primary reduction operators govern this process: (i) the sample operator $\Sigma_n$ sums across individual data instances to form contingency tables; (ii) the macro operator $\Sigma_m$ averages per-class metrics to produce an unweighted, balanced summary; and (iii) the micro operator $\Sigma_1$ pools all class cells globally before computing a single overall rate. In dynamic streaming environments, the new operator $\Upsigma_{\mathrm{t}}$ introduces the \emph{temporal axis}. The operator $\Sigma_n$ weights all past samples equally, which assumes an unchanging world, and $\Upsigma_{\mathrm{t}}$ weights them by recency. Thus $\Upsigma_{\mathrm{t}}$ generalizes $\Sigma_n$: both build $\mathbold{C}(t)$ from the raw instance stream and differ only in whether the weighting is uniform or time-dependent. It is not a fourth operator composed alongside it, which is why only $(\Sigma_1, \Sigma_m)$ appear in the composition tuples of \S\ref{sec:composition} onward. By progressively fading out stale historical observations (via fading parameter $\alpha$) or restricting focus to a recent moving window ($W$), $\Upsigma_{\mathrm{t}}$ continuously reflects the model's present health while composing with spatial class reductions $(\Sigma_1, \Sigma_m)$.

The two canonical stream weighting matrices are:
\begin{enumerate}
    \item Exponential fading ($\Upsigma_{\mathrm{t}}^\alpha$) assigns the exponentially decaying weights $w_k(t) = (1 - \alpha)\,\alpha^{t-k}$ for a fading parameter $\alpha \in (0, 1)$. This yields the recursive update:
    \begin{equation}
    \label{eq:fading_update}
    \mathbold{C}_\alpha(t) = \alpha \mathbold{C}_\alpha(t-1) + (1 - \alpha) \mathbold{y}_t \mathbold{p}_t^T, \qquad n_{\mathrm{eff}} = \frac{1}{1 - \alpha}
    \end{equation}
    The implementation stores the unnormalised state $\tilde{\mathbold{C}}_\alpha = \mathbold{C}_\alpha/(1-\alpha)$, updated as $\alpha\tilde{\mathbold{C}}_\alpha + \mathbold{y}_t\mathbold{p}_t^T$; every metric is a ratio of cells and is the same under both conventions, but statements that read a cell as a probability use the normalised state. Here $n_{\mathrm{eff}}$ is the memory horizon of the weights; the effective sample size that governs variance is $(1+\alpha)/(1-\alpha)$ (Proposition~\ref{prop:variance_scaling}). As an example, for two classes and $\alpha=0.9$, after a correct prediction of class~1 the state $\mathbold{C}_\alpha$ gains $0.1$ in the cell $(1,1)$ and every older cell is multiplied by $0.9$; the recall of class~1 is $\mathrm{TP}/(\mathrm{TP}+\mathrm{FN})$ of the resulting cells, a weighted average of the recent outcomes of that class.
    \item The sliding window ($\Upsigma_{\mathrm{t}}^W$) assigns uniform weights over a moving historical horizon of length $W \in \mathbb{N}^+$ with weighting matrix $\mathbold{W}(t) = \frac{1}{W} \operatorname{diag}(\mathbold{0}_{t-W}, \mathbold{1}_W)$. For $t \ge W$ this yields the windowed aggregation, whose recursive form holds for $t > W$ (at $t = W$ no instance exits) (before the window fills, the state is the sum of the $t$ instances divided by $W$, and ratios are unaffected):
    \begin{equation}
    \label{eq:window_update}
    \mathbold{C}_W(t) = \frac{1}{W} \sum_{k=0}^{W-1} \mathbold{y}_{t-k} \mathbold{p}_{t-k}^T = \mathbold{C}_W(t-1) + \frac{1}{W} \left( \mathbold{y}_t \mathbold{p}_t^T - \mathbold{y}_{t-W} \mathbold{p}_{t-W}^T \right)
    \end{equation}
\end{enumerate}

\begin{example}[Running example: the stream and its temporal state]
\label{ex:running}
Throughout the paper we follow one stream of $8$ instances and $m=3$ classes $A,B,C$, with true labels $A\,B\,A\,C\,B\,A\,B\,B$ and predictions $A\,B\,B\,C\,B\,A\,C\,B$, so that the classifier errs at $t=3$ ($A$ predicted as $B$) and at $t=7$ ($B$ predicted as $C$). With exponential fading $\alpha=\tfrac12$ the state after the first four instances is
\[
\mathbold{C}_\alpha(4)=\begin{pmatrix}\tfrac1{16}&\tfrac14&0\\[2pt]0&\tfrac18&0\\[2pt]0&0&\tfrac12\end{pmatrix},
\qquad \text{total mass } 1-\alpha^4=\tfrac{15}{16}.
\]
With a sliding window $W=4$ the state at $t=8$ covers instances $5$ to $8$,
\[
\mathbold{C}_W(8)=\begin{pmatrix}\tfrac14&0&0\\[2pt]0&\tfrac12&\tfrac14\\[2pt]0&0&0\end{pmatrix}.
\]
Appendix~\ref{app:running} lists every intermediate value of the running example, and the script \texttt{experiments/\allowbreak running\_example.py} reproduces them.
\end{example}

\begin{figure}[!ht]
\centering
\includegraphics[width=\textwidth]{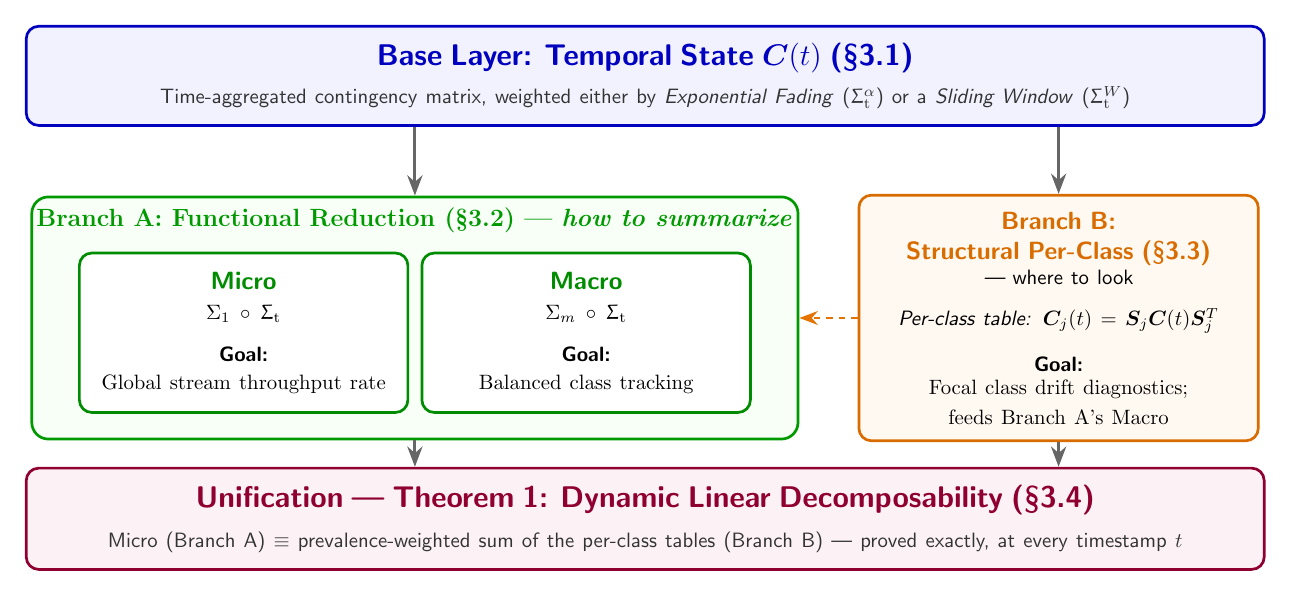}
\caption{Algebraic architecture of data stream evaluation. The temporal state $\mathbold{C}(t)$ (base layer, \S\ref{sec:temporal_im}) feeds two evaluation branches: functional reduction compositions ($\Sigma \circ \Upsigma_{\mathrm{t}}$, Branch~A, \S\ref{sec:composition}) and structural per-class projections ($\mathbold{S}_j$, Branch~B, \S\ref{sec:ovr}). Theorem~\ref{thm:temporal_decomposability} (\S\ref{sec:theorems}) relates them. The dashed arrow shows that the per-class matrices $\mathbold{C}_j(t)$ of Branch~B are the input of the macro reduction $\Sigma_m \circ \Upsigma_{\mathrm{t}}$ of Branch~A.}
\label{fig:hierarchy}
\end{figure}

Figure~\ref{fig:hierarchy} summarises the architecture: the temporal state, the two evaluation branches built on it (functional reductions, \S\ref{sec:composition}, and structural per-class projections, \S\ref{sec:ovr}) and the theorem that relates them (Theorem~\ref{thm:temporal_decomposability}, \S\ref{sec:theorems}).

Both weightings give non-negative states through the bilinear form~\eqref{eq:bilinear_temporal} and keep every downstream composition and the decomposability property. The choice between $\mathbold{C}_\alpha(t)$ and $\mathbold{C}_W(t)$ sets the memory profile and the responsiveness to drift:
\begin{itemize}
    \item The sliding window $\mathbold{C}_W(t)$ has a finite horizon over exactly the last $W$ observations, so it keeps no memory of concepts older than $W$ steps. It needs a FIFO buffer of size $O(W \cdot m)$ to subtract exiting samples, and its step-like impulse response produces abrupt metric shifts when past bursts leave the window.
    \item Exponential fading $\mathbold{C}_\alpha(t)$ forgets continuously with memory horizon $n_{\mathrm{eff}} = \frac{1}{1-\alpha}$, uses $O(m^2)$ space and time per update, and stores no historical instances. Its trajectories are smooth and have no boundary artifacts.
\end{itemize}

\subsection{Composition Algebra with Core Reduction Operators}
\label{sec:composition}
Having established the global temporal matrix $\mathbold{C}(t)$ as the streaming state representation, we formalize how the temporal operator $\Upsigma_{\mathrm{t}}$ composes functionally with the spatial reduction operators of the core framework ($\Sigma_1, \Sigma_m$) to generate scalar streaming evaluations. Each reduction operator targets a distinct analytical goal in data stream monitoring:

\begin{enumerate}
    \item The micro-temporal composition ($\Sigma_1 \circ \Upsigma_{\mathrm{t}}$) pools the $m$ one-vs-rest tables $\mathbold{S}_j\mathbold{C}(t)\mathbold{S}_j^T$ (\S\ref{sec:ovr}) into one table over time before evaluating the scalar metric $M$ once:
    \begin{equation}
    (\Sigma_1 \circ \Upsigma_{\mathrm{t}} M)(t) = M\left( \sum_{j=1}^m \mathbold{S}_j \mathbold{C}(t) \mathbold{S}_j^T \right)
    \end{equation}
    Every arrival contributes equally to the total cell mass, so this composition tracks the overall service level, which suits settings where an error costs the same whatever the class. A drop in a minority class can be masked if the majority class stays accurate. The pooled accuracy is the accuracy of the pooled one-vs-rest cells, $1-2(1-\mathrm{acc})/m$ for multiclass accuracy $\mathrm{acc}$, not multiclass accuracy itself.

    \item The macro-temporal composition ($\Sigma_m \circ \Upsigma_{\mathrm{t}}$) evaluates the scalar metric $M$ on each class and averages the $m$ values without weights:
    \begin{equation}
    (\Sigma_m \circ \Upsigma_{\mathrm{t}} M)(t) = \frac{1}{m} \sum_{j=1}^m M\left( \mathbold{C}_j(t) \right)
    \end{equation}
    The macro operator gives each class the weight $1/m$ whatever its prevalence at the time, so a degradation on a rare or emerging class is not hidden by accurate predictions on a prevalent one.

\end{enumerate}

\subsection{Multiclass Selection Matrices and Per-Class Projections}
\label{sec:ovr}
In contrast to the functional composition algebra of \S\ref{sec:composition}, which produces scalar summaries, evaluating non-linear metrics and isolating drift requires the structure of the state matrix itself. While the global temporal state $\mathbold{C}(t) \in \mathbb{R}^{m \times m}$ captures the full joint allocation of predictions and ground truth across all $m$ classes, standard classification measures (such as Recall, Specificity, Precision, and $F_1$) are intrinsically defined on binary $2 \times 2$ contingency tables. Furthermore, effective operational monitoring in data streams requires isolating which classes are experiencing concept drift.

To achieve this without procedural slicing, structural per-class projections are formalized via linear selection matrices $\mathbold{S}_j \in \mathbb{R}^{2 \times m}$ defined for each class $j \in \{1, \dots, m\}$:
\begin{equation}
\mathbold{S}_j = \begin{bmatrix} \mathbold{e}_j^T \\ \mathbold{1}_m^T - \mathbold{e}_j^T \end{bmatrix}
\end{equation}
where $\mathbold{e}_j$ is the $j$-th standard basis vector and $\mathbold{1}_m$ is the all-ones vector. The $2 \times 2$ per-class temporal contingency matrix for class $j$ at time $t$ is obtained via the matrix congruence transformation:
\begin{equation}
\label{eq:ovr_projection}
\mathbold{C}_j(t) = \mathbold{S}_j \, \mathbold{C}(t) \, \mathbold{S}_j^T = \mathbold{S}_j \left( \mathbold{Y}_{1:t}^T \mathbold{W}(t) \mathbold{P}_{1:t} \right) \mathbold{S}_j^T = \begin{bmatrix}
\mathrm{TP}_j(t) & \mathrm{FN}_j(t) \\
\mathrm{FP}_j(t) & \mathrm{TN}_j(t)
\end{bmatrix}
\end{equation}
The four entries of $\mathbold{C}_j(t)$ correspond directly to the temporally discounted counts for class $j$:
\begin{equation}
\begin{gathered}\mathrm{TP}_j(t) = [\mathbold{C}(t)]_{jj}, \quad \mathrm{FN}_j(t) = \sum_{k \ne j} [\mathbold{C}(t)]_{jk}, \\ \mathrm{FP}_j(t) = \sum_{l \ne j} [\mathbold{C}(t)]_{lj}, \quad \mathrm{TN}_j(t) = \sum_{l \ne j} \sum_{k \ne j} [\mathbold{C}(t)]_{lk}
\end{gathered}
\end{equation}

The structural per-class projection serves two primary analytical and operational goals:
\begin{itemize}
    \item When an aggregate alarm triggers (a drop in macro-$F_1$ or micro-accuracy), global metrics cannot identify the cause. Tracking the per-class metrics $M_j(t) = M(\mathbold{C}_j(t))$ separates a loss of class recall, $\mathrm{TP}_j(t) / (\mathrm{TP}_j(t) + \mathrm{FN}_j(t))$, from a rise in false alarms $\mathrm{FP}_j(t)$.
    \item The congruence $\mathbold{S}_j \mathbold{C}(t) \mathbold{S}_j^T$ is the input of the macro-temporal operator $\Sigma_m \circ \Upsigma_{\mathrm{t}}$ and needs no separate per-class data structures: all $m$ binary tables $\mathbold{C}_j(t)$ are extracted from the single state $\mathbold{C}(t)$ in $O(m^2)$ total time.
\end{itemize}

\begin{remark}[Pairwise projections are a diagnostic, not a result]
\label{rem:ovo}
The per-class projection collapses all other classes into a single rest, so $\mathrm{FN}_j(t)$ and $\mathrm{FP}_j(t)$ show that class $j$ is losing recall or precision but not which competing class is responsible. For a pair $j,k$ the selection matrix $\mathbold{S}_{jk} = [\mathbold{e}_j, \mathbold{e}_k]^T$ extracts the confusion restricted to $\{j,k\}$. This congruence alone is not a conditional rate, because instances whose true or predicted label lies outside $\{j,k\}$ must be masked from the denominator, and no analogue of Theorem~\ref{thm:temporal_decomposability} recombines the pairwise tables. The base framework likewise treats the One-vs-One decomposition as partial~\citep{prati2026unified}. We do not develop $\mathbold{S}_{jk}$ here.
\end{remark}

\subsection{Algebraic Theorems and Complexity Bounds}
\label{sec:theorems}
Having structured the dual evaluation branches, we state the results on the exactness, computational cost and statistical stability of the temporal state ($t \to \infty$).

\begin{proposition}[Resource Complexity of Exponential Fading]
\label{prop:complexity}
The recursive fading operator $\Upsigma_{\mathrm{t}}^\alpha$ requires $O(m^2)$ memory to store $\mathbold{C}_\alpha(t)$ and $O(m^2)$ floating-point operations per stream update, both independent of the stream length $t$.
\end{proposition}
\begin{proof}
Evaluating Equation~\eqref{eq:fading_update} requires scalar multiplication $\alpha \mathbold{C}_\alpha(t-1)$, outer product $(1-\alpha)\mathbold{y}_t \mathbold{p}_t^T$, and matrix addition ($3m^2$ floating-point operations). No historical tuples $(\mathbold{y}_k, \mathbold{p}_k)$ for $k < t$ are retained, which bounds the state storage by $m^2$ scalar values.
\end{proof}

\begin{proposition}[Sliding-Window Resource Complexity]
\label{prop:complexity_window}
The sliding-window operator $\Upsigma_{\mathrm{t}}^W$, maintained via a FIFO queue of the $W$ most
recent raw observation pairs $(\mathbold{y}_k,\mathbold{p}_k)$, requires $O(W\cdot m)$ memory
for the queue plus the $O(m^2)$ state matrix $\mathbold{C}_W(t)$, and executes each update in
$O(m^2)$ floating-point operations, independent of the stream length $t$. The per-update cost is the same as for exponential fading, but the memory grows linearly in $W$.
\end{proposition}
\begin{proof}
Updating $\mathbold{C}_W(t) = \mathbold{C}_W(t-1) + \mathbold{y}_t\mathbold{p}_t^T -
\mathbold{y}_{t-W}\mathbold{p}_{t-W}^T$ requires: (i) enqueuing $(\mathbold{y}_t,\mathbold{p}_t)$
and dequeuing $(\mathbold{y}_{t-W},\mathbold{p}_{t-W})$ from the FIFO queue, both $O(1)$ operations
independent of $W$; (ii) forming the new outer product and subtracting the evicted one, $2m^2$
floating-point operations. Unlike exponential fading, which discards each instance immediately,
the queue must retain the $W$ most recent raw pairs to know what to subtract when they exit the
window, giving $O(W\cdot m)$ additional memory that grows with $W$, as measured in Table~\ref{tab:exp7}.
\end{proof}

\noindent Both operators thus meet the same $O(m^2)$ per-update time bound, which does not depend on $t$, and differ only in
memory profile. What makes either a safe choice for the aggregation machinery that follows is that
both preserve linear decomposability across time, established next.

\begin{theorem}[Dynamic Temporal Preservation of Linear Decomposability]
\label{thm:temporal_decomposability}
Let $M = \frac{\langle \mathbold{a}, \mathbold{v} \rangle}{\langle \mathbold{b}, \mathbold{v} \rangle}$ be a linear-fractional performance measure, a ratio of two linear functions of the four cells TP, FN, FP, TN of a class table (accuracy, recall, precision and $F_1$ are of this form), with coefficient vectors satisfying $0 \le \mathbold{a} \le \mathbold{b}$ entrywise, and let $\mathbold{v}_{\mathrm{micro}}(t) = \sum_j \mathbold{v}_j(t)$ with $\langle \mathbold{b}, \mathbold{v}_{\mathrm{micro}}(t) \rangle > 0$. Under any nonnegative temporal weighting of nonnegative label and prediction vectors, in particular exponential fading $\Upsigma_{\mathrm{t}}^\alpha$ and sliding window $\Upsigma_{\mathrm{t}}^W$, the micro-temporal composition equals the denominator-weighted macro-temporal composition at every timestamp $t$, where classes with $\langle \mathbold{b}, \mathbold{v}_j(t) \rangle = 0$ get weight $0$ and are omitted:
\begin{equation}
(\Sigma_1 \circ \Upsigma_{\mathrm{t}} M)(t) = \sum_{j=1}^m \omega_j(t) M\left( \mathbold{S}_j \mathbold{C}(t) \mathbold{S}_j^T \right), \qquad \text{where} \quad \omega_j(t) = \frac{\langle \mathbold{b}, \mathbold{v}_j(t) \rangle}{\sum_{l=1}^m \langle \mathbold{b}, \mathbold{v}_l(t) \rangle}
\end{equation}
\end{theorem}

\begin{proof}
Let $\mathbold{v}_j(t) = \operatorname{vec}(\mathbold{S}_j \mathbold{C}(t) \mathbold{S}_j^T) = [\mathrm{TP}_j(t), \mathrm{FN}_j(t), \mathrm{FP}_j(t), \mathrm{TN}_j(t)]^T$ denote the vectorized per-class contingency cells. Because $\Upsigma_{\mathrm{t}}$ is a linear operator on the space of matrices, it commutes with the linear selection matrices $\mathbold{S}_j$ and global micro-summation $\mathbold{v}_{\mathrm{micro}}(t) = \sum_{j=1}^m \mathbold{v}_j(t)$.

The micro-temporal metric evaluated on the global pooled stream is given by:
\begin{equation}
(\Sigma_1 \circ \Upsigma_{\mathrm{t}} M)(t) = \frac{\langle \mathbold{a}, \mathbold{v}_{\mathrm{micro}}(t) \rangle}{\langle \mathbold{b}, \mathbold{v}_{\mathrm{micro}}(t) \rangle} = \frac{\sum_{j=1}^m \langle \mathbold{a}, \mathbold{v}_j(t) \rangle}{\sum_{l=1}^m \langle \mathbold{b}, \mathbold{v}_l(t) \rangle}
\end{equation}
A class with $\langle \mathbold{b}, \mathbold{v}_j(t) \rangle = 0$ also has $\langle \mathbold{a}, \mathbold{v}_j(t) \rangle = 0$, because $\mathbold{a} \le \mathbold{b}$ and $\mathbold{v}_j(t) \ge 0$, so it contributes to neither sum. Multiplying and dividing each remaining summand by its positive local denominator $\langle \mathbold{b}, \mathbold{v}_j(t) \rangle$:
\begin{equation}
(\Sigma_1 \circ \Upsigma_{\mathrm{t}} M)(t) = \sum_{j=1}^m \left( \frac{\langle \mathbold{b}, \mathbold{v}_j(t) \rangle}{\sum_{l=1}^m \langle \mathbold{b}, \mathbold{v}_l(t) \rangle} \right) \left( \frac{\langle \mathbold{a}, \mathbold{v}_j(t) \rangle}{\langle \mathbold{b}, \mathbold{v}_j(t) \rangle} \right) = \sum_{j=1}^m \omega_j(t) M_j(t)
\end{equation}
which establishes the equivalence for all timestamps $t \ge 1$. The argument uses only the linearity of $\mathbold{C} \mapsto \mathbold{v}_j$ and the nonnegativity of the weights, so it holds for any temporal weighting and not only for the two operators of this paper.
\end{proof}

Theorem~\ref{thm:temporal_decomposability} applies to every linear-fractional measure. For instance, in stream Accuracy ($\mathbold{a} = [1, 0, 0, 1]^T, \mathbold{b} = [1, 1, 1, 1]^T$), the weights $\omega_j(t) = 1/m$ reduce to the standard uniform average; in stream Recall ($\mathbold{a} = [1, 0, 0, 0]^T, \mathbold{b} = [1, 1, 0, 0]^T$), the weight $\omega_j(t)$ equals the dynamic class prevalence $P_t(Y=j) = \sum_{k} [\mathbold{C}(t)]_{jk}$; and in stream Precision ($\mathbold{a} = [1, 0, 0, 0]^T, \mathbold{b} = [1, 0, 1, 0]^T$), $\omega_j(t)$ reflects the model's positive prediction rate $P_t(\hat{Y}=j) = \sum_{l} [\mathbold{C}(t)]_{lj}$.

\begin{example}[Running example: pooled and per-class measures]
\label{ex:running_decomp}
The three tables $\mathbold{S}_j\mathbold{C}_W(8)\mathbold{S}_j^T$ of the window state have $(\mathrm{TP}_j,\mathrm{FN}_j,\mathrm{FP}_j,\mathrm{TN}_j)=(\tfrac14,0,0,\tfrac34)$, $(\tfrac12,\tfrac14,0,\tfrac14)$ and $(0,0,\tfrac14,\tfrac34)$ for $A$, $B$ and $C$, and the pooled table is $(\tfrac34,\tfrac14,\tfrac14,\tfrac74)$. The per-class precisions are $(1,1,0)$ with denominators $\mathrm{TP}_j+\mathrm{FP}_j=(\tfrac14,\tfrac12,\tfrac14)$, so the weights of Theorem~\ref{thm:temporal_decomposability} are $\omega=(\tfrac14,\tfrac12,\tfrac14)$. The pooled precision is $\tfrac34$ and the denominator-weighted macro is $\tfrac14\cdot1+\tfrac12\cdot1+\tfrac14\cdot0=\tfrac34$, whereas the unweighted macro is $\tfrac23$.
\end{example}

Theorem~\ref{thm:temporal_decomposability} also gives a class/prevalence attribution of a change in the pooled metric. When it shifts by $\Delta M_{\mathrm{micro}}(t) = M_{\mathrm{micro}}(t) - M_{\mathrm{micro}}(t-1)$, the aggregate delta decomposes linearly into two distinct operational sources:
\begin{equation}
\label{eq:drift_decomposition}
\Delta (\Sigma_1 \circ \Upsigma_{\mathrm{t}} M)(t) = \sum_{j=1}^m \bar{\omega}_j(t) \, \Delta M_j(t) + \sum_{j=1}^m \Delta \omega_j(t) \, \bar{M}_j(t)
\end{equation}
where $\bar{\omega}_j(t)$ and $\bar{M}_j(t)$ denote mid-point values. The first term quantifies the contribution of \emph{intrinsic performance drift} localized within specific classes, whereas the second term isolates the impact of \emph{prevalence drift} (dynamic class prior shifts) on aggregate accuracy. A formalization of drift taxonomies, dynamic response timescales, and continuous differential Jacobian decompositions is established in \S\ref{sec:drift_decomposition}.

\begin{proposition}[Asymptotic Expectation and Variance Scaling]
\label{prop:variance_scaling}
Let the pairs $(\mathbold{y}_t, \mathbold{p}_t)$ be independent and identically distributed with population state matrix $\mathbold{C}_{\mathrm{pop}} = \mathbb{E}[\mathbold{y}_t \mathbold{p}_t^T]$ and elementwise variance $\mathbold{V}_{\mathrm{inst}} = \mathrm{Var}(\mathbold{y}_t \mathbold{p}_t^T)$ (for instance, a classifier that is not updated, so that predictions do not depend on past labels). The exponential fading estimator $\Upsigma_{\mathrm{t}}^\alpha$ satisfies:
\begin{gather}
\mathbb{E}[\mathbold{C}_\alpha(t)] = (1-\alpha^t)\,\mathbold{C}_{\mathrm{pop}} \xrightarrow{t \to \infty} \mathbold{C}_{\mathrm{pop}}, \\
\mathrm{Var}([\mathbold{C}_\alpha(t)]_{jk}) = \left( \frac{1 - \alpha}{1 + \alpha} \right) (1-\alpha^{2t}) [\mathbold{V}_{\mathrm{inst}}]_{jk} \xrightarrow{t \to \infty} \frac{[\mathbold{V}_{\mathrm{inst}}]_{jk}}{n_{\mathrm{ess}}}, \qquad n_{\mathrm{ess}} = \frac{1+\alpha}{1-\alpha}
\end{gather}
\end{proposition}
\begin{proof}
Unrolling the recurrence~\eqref{eq:fading_update} yields $\mathbold{C}_\alpha(t) = (1-\alpha) \sum_{s=0}^{t-1} \alpha^s \mathbold{y}_{t-s} \mathbold{p}_{t-s}^T$. Taking expectations gives $\mathbb{E}[\mathbold{C}_\alpha(t)] = (1-\alpha) \frac{1-\alpha^t}{1-\alpha} \mathbold{C}_{\mathrm{pop}} = (1-\alpha^t)\,\mathbold{C}_{\mathrm{pop}}$. By independence the variance is
\begin{equation}
\mathrm{Var}([\mathbold{C}_\alpha(t)]_{jk}) = (1-\alpha)^2 \sum_{s=0}^{t-1} \alpha^{2s} [\mathbold{V}_{\mathrm{inst}}]_{jk} = (1-\alpha)^2 \left( \frac{1 - \alpha^{2t}}{1 - \alpha^2} \right) [\mathbold{V}_{\mathrm{inst}}]_{jk},
\end{equation}
and $1 - \alpha^2 = (1-\alpha)(1+\alpha)$ gives the stated form.
\end{proof}

Proposition~\ref{prop:variance_scaling} gives the variance half of a bias-variance trade-off (the bias half is the tracking lag of \S\ref{sec:drift_taxonomy}). A larger $\alpha$ gives a lower-variance estimate and a slower response to drift, and a smaller $\alpha$ the opposite.

\begin{remark}[Autocorrelated entries]
\label{rem:autocorr}
Proposition~\ref{prop:variance_scaling} assumes independent entries. If the entries are second-order stationary with variance $v$ and autocorrelation $\rho_d$ at lag $d$, the weights of $\Upsigma_{\mathrm{t}}^\alpha$ have autocorrelation $\sum_i w_i w_{i+d} \to (1-\alpha)^2\alpha^d/(1-\alpha^2)$, and the asymptotic variance of the fading estimator is
\begin{equation}
\label{eq:variance_inflation}
\mathrm{Var} = v\,\frac{1-\alpha}{1+\alpha}\,\kappa(\alpha), \qquad \kappa(\alpha) = 1 + 2\sum_{d \ge 1} \alpha^d \rho_d ,
\end{equation}
so that the effective sample size $(1+\alpha)/(1-\alpha)$ is divided by $\kappa(\alpha)$. For $\rho_d = \rho^d$ this is $(1+\alpha\rho)/(1-\alpha\rho)$, but real autocorrelations decay more slowly and the lag-one value alone is a poor guide. For the frozen gradient-boosting classifier of R1 (Elec2) and R2 (Airlines) at $\alpha = 0.98$, after removing a moving average of $1{,}001$ steps, the value of $\kappa$ computed from the sample autocorrelation up to lag $1{,}500$ is $8.8$ and $1.04$ for the error indicator. For the class-time sequences that the detector of \S\ref{sec:monitoring} averages it is $20.8$ (recall) and $13.3$ (specificity) on Elec2, and $2.4$ and $1.6$ on Airlines, against $5.3$, $4.0$, $3.1$ and $2.4$ from the AR(1) formula with the lag-one autocorrelation. On Elec2 the sample variance of the fading estimate of the error indicator is $9.0$ times the value for independent entries, against $8.8$ from Eq.~\eqref{eq:variance_inflation}. The factor measures any variation of the mean that the moving average leaves in, so on a stream with real drift it includes that drift, and we did not separate the two.
\end{remark}

\subsection{Explicit Auditing of Zero-Division Voids}
\label{sec:zero_division}
In data streams with strong class imbalance or emerging classes ($P_t(Y=j) \ll 1$), evaluation windows frequently experience intervals with zero class support ($\mathrm{TP}_j(t) + \mathrm{FN}_j(t) = 0$) or zero positive predictions ($\mathrm{TP}_j(t) + \mathrm{FP}_j(t) = 0$). In such states, metric denominators $\langle \mathbold{b}, \mathbold{v}_j(t) \rangle = 0$ create mathematical singularities.

Conventional batch libraries (such as scikit-learn) typically apply silent default substitutions (e.g., $\mathtt{zero\_division=0}$), which distort stream evaluation: a rare class that does not appear in a moving window is scored as $0.0$ without notice and can trigger false drift alarms in macro-aggregated indicators.

To prevent silent distortion while maintaining uninterrupted streaming pipelines, the framework formalizes a dynamic support indicator vector $\mathbold{u}(t) \in \{0, 1\}^m$, where, without latency, $u_j(t) = \mathbb{I}[\langle \mathbold{b}, \mathbold{v}_j(t) \rangle > \varepsilon]$ (the case with latency is treated after Example~\ref{ex:running_void}), with $\varepsilon=0$ under a sliding window and a small positive tolerance under exponential fading, where cell masses decay but never reach exactly zero. The macro-temporal composition $\Sigma_m \circ \Upsigma_{\mathrm{t}}$ supports three auditable policies:
\begin{enumerate}
    \item With $\mathtt{policy = exclude}$ the macro composition averages metrics over the classes with non-zero support in the current temporal state (and is undefined, as for $\mathtt{nan}$, if no class has support):
    \begin{equation}
    \label{eq:policy_exclude}
    (\Sigma_m \circ \Upsigma_{\mathrm{t}} M)_{\mathrm{exclude}}(t) = \frac{1}{\sum_{j=1}^m u_j(t)} \sum_{j: u_j(t)=1} M\left( \mathbold{C}_j(t) \right)
    \end{equation}
    \item With $\mathtt{policy = zero}$ an unobserved class gets $0.0$, a conservative assumption:
    \begin{equation}
    \label{eq:policy_zero}
    (\Sigma_m \circ \Upsigma_{\mathrm{t}} M)_{\mathrm{zero}}(t) = \frac{1}{m} \sum_{j=1}^m u_j(t) M\left( \mathbold{C}_j(t) \right)
    \end{equation}
    \item With $\mathtt{policy = nan}$ the composition emits $\mathrm{NaN}$ whenever any class has undefined support, which forces explicit handling upstream.
\end{enumerate}
In all cases, the framework emits an explicit diagnostic audit tuple $(\mathbold{u}(t), m_{\mathrm{active}}(t))$, where $m_{\mathrm{active}}(t) = \sum_{j=1}^m u_j(t)$ is the number of classes with non-zero support (the same quantity already appearing as the denominator of Eq.~\eqref{eq:policy_exclude}), alongside every evaluation update, ensuring full transparency in automated monitoring dashboards.

\begin{example}[Running example: a void]
\label{ex:running_void}
In the window state of Example~\ref{ex:running} no instance of class $C$ is true, so $\mathrm{TP}_C+\mathrm{FN}_C=0$ and $\mathrm{Recall}_C$ is $0/0$. The support indicator is $\mathbold{u}=(1,1,0)$ and $m_{\mathrm{active}}=2$. The per-class recalls are $(1,\tfrac23,\text{undefined})$, so the macro recall is $\tfrac56$ under $\mathtt{exclude}$, $\tfrac59$ under $\mathtt{zero}$ and undefined under $\mathtt{nan}$. The pooled recall is $\tfrac34$, equal to the weighted macro with weights $(\tfrac14,\tfrac34,0)$ that omit $C$ (Theorem~\ref{thm:temporal_decomposability}).
\end{example}

Under verification latency or censoring (\S\ref{sec:delayed_labels}) the state $\mathbold{C}(t)$ also holds interim estimates of the unverified instances (the model's calibrated class probabilities standing in for the labels that have not arrived, \S\ref{sec:product_tnorm}), which are spread over the rows of every class the model finds plausible, so a class can have mass in $\mathbold{C}(t)$ without an instance of it in the window. We therefore distinguish the interim support $u^{\mathrm{int}}_j(t) = \mathbb{I}[\langle \mathbold{b}, \mathbold{v}_j(\mathbold{C}(t)) \rangle > \varepsilon]$ and the verified support $u^{\mathrm{ver}}_j(t) = \mathbb{I}[\langle \mathbold{b}, \mathbold{v}_j(\mathbold{C}_{\mathrm{verified}}(t)) \rangle > \varepsilon]$, and the coverage $\kappa_j(t)$, the share of the denominator $\langle \mathbold{b}, \mathbold{v}_j(\mathbold{C}(t)) \rangle$ that comes from verified instances. By marginal prediction consistency (Eq.~\eqref{eq:col_marg}) the column sums of $\mathbold{C}(t)$ do not depend on the interim estimates. If the denominator is a combination of the prediction-side sums $\mathrm{TP}_j+\mathrm{FP}_j$ and $\mathrm{FN}_j+\mathrm{TN}_j$ (precision, accuracy), $u^{\mathrm{int}}_j$ is exact. If it involves the label side (recall, specificity), only $u^{\mathrm{ver}}_j$ is safe. The policies below use $u_j = u^{\mathrm{int}}_j$ for the first kind and $u_j = u^{\mathrm{ver}}_j$ for the second, which coincide without latency, and the audit tuple reports $(\mathbold{u}^{\mathrm{ver}}, \mathbold{u}^{\mathrm{int}}, \mathbold{\kappa})$. A value computed for a class with $0 < \kappa_j < 1$ is an estimate, and one with $\kappa_j = 0$ is not reported as a measurement.

A practical question these policies raise immediately is \emph{how often} they trigger for a given class and a given choice of $W$ or $\alpha$. A practitioner needs this information before deployment, not only as a post-hoc diagnostic.

\begin{proposition}[Undefined-Window Probability Under Stationary Arrivals]
\label{prop:undefined_window}
Let class $j$ arrive i.i.d.\ with per-instance probability $p_j = P_t(Y=j)$, stationary over the relevant horizon. Under a sliding window of size $W$ (steady state, $t \ge W$), the probability that $u_j(t) = 0$, that is, that $\mathrm{Recall}_j(t)$ is undefined because no instance of class $j$ occurred in the window, is exactly
\begin{equation}
\label{eq:undefined_window_iid}
P\bigl(u_j(t) = 0\bigr) = (1 - p_j)^W.
\end{equation}
Under exponential fading with a mass tolerance $\varepsilon$ and $t \ge s^\ast$, $(1-p_j)^{s_\varepsilon} \le P(u_j(t)=0) \le (1-p_j)^{s^\ast}$, where $s^\ast = \min\{s : (1-\alpha)\alpha^s \le \varepsilon\} = O\!\left(n_{\mathrm{eff}} \log \frac{1}{n_{\mathrm{eff}}\varepsilon}\right)$ is the first age at which the weight of a single arrival does not exceed $\varepsilon$ (the eviction age of Eq.~\eqref{eq:truncation_bound} is defined with a strict inequality and differs only when the weight equals $\varepsilon$) and $s_\varepsilon = \lceil \ln(1/\varepsilon)/\ln(1/\alpha) \rceil$ is the number of silent steps after which the whole mass of the class is at most $\varepsilon$.
\end{proposition}
\begin{proof}
Under a sliding window, $u_j(t)=0$ iff none of the $W$ i.i.d.\ arrivals in $(t-W, t]$ has true label $j$, which by independence has probability $(1-p_j)^W$ exactly. Under fading with eviction, $u_j(t)=0$ once the most recent class-$j$ arrival is older than $s^\ast$ steps (Eq.~\eqref{eq:truncation_bound}); the same independence argument gives the two bounds: an arrival within the last $s^\ast$ steps keeps the mass above $\varepsilon$, and no arrival for $s_\varepsilon$ steps brings it to at most $\varepsilon$.
\end{proof}

\begin{corollary}[Design rule for the undefined-window rate]
\label{cor:window_sizing}
Sizing a streaming parameter to guarantee a target confidence level is the same design principle underlying adaptive windowing schemes such as ADWIN~\citep{bifet2007adwin}, applied here to the undefined-window rate specifically: to keep it below a target tolerance $\delta \in (0,1)$, Eq.~\eqref{eq:undefined_window_iid} requires
\begin{equation}
\label{eq:window_sizing_rule}
W \ge \frac{\ln(1/\delta)}{-\ln(1-p_j)} \approx \frac{\ln(1/\delta)}{p_j} \qquad \text{(the approximation tightening as } p_j \to 0\text{)},
\end{equation}
and under fading the upper bound of Proposition~\ref{prop:undefined_window} gives the corresponding rule with $s^\ast$ in place of $W$.
\end{corollary}

\begin{corollary}[Undefined-window probability with latency]
\label{cor:undefined_latency}
Let the delays be independent of the arrivals and of the class, with $F(s)=P(\Delta \le s)$, and let a label be requested with probability $b$ independently of the instance. For a sliding window of size $W$, the probability that no instance of class $j$ in the window is verified at time $t$, so that $u^{\mathrm{ver}}_j(t)=0$, is
\[
P\bigl(u^{\mathrm{ver}}_j(t)=0\bigr) = \prod_{s=0}^{W-1}\bigl(1 - p_j\, b\, F(s)\bigr),
\]
where $s$ is the age of the instance. For a constant delay $\Delta < W$ it is $(1-p_j b)^{W-\Delta}$, that is $(1-p_j)^{W-\Delta}$ for $b=1$, and it is $1$ if $\Delta \ge W$. The window needed for a target tolerance $\delta$ grows by the delay, $W \ge \Delta + \ln(1/\delta)/(-\ln(1-p_j b))$.
\end{corollary}
\begin{proof}
An instance of age $s$ is a verified instance of class $j$ with probability $p_j b F(s)$ and these events are independent across positions, so the probability that none occurs is the product. For a constant delay $F(s)=\mathbb{I}[s \ge \Delta]$.
\end{proof}

\begin{remark}[Bursty arrivals inflate the undefined-window rate]
\label{rem:bursty_undefined}
Proposition~\ref{prop:undefined_window} assumes i.i.d.\ arrivals, and rare events are often bursty. Model the arrivals of class $j$ as a two-state on/off Markov process~\citep{kleinberg2003bursty}, off$\to$on with probability $\beta$ and on$\to$off with probability $\gamma$ per step, with a small off-burst singleton arrival rate $\rho$ (the generator of S8 in \S\ref{sec:exp4}). The stationary undefined-window probability is approximately
\begin{equation}
\label{eq:undefined_window_markov}
P\bigl(u_j(t)=0\bigr) \approx \underbrace{\frac{\gamma}{\beta+\gamma}}_{\pi_{\mathrm{off}}} (1-\beta)^{W-1} (1-\rho)^{W},
\end{equation}
which requires the chain to be off for the whole window, with probability $\pi_{\mathrm{off}}(1-\beta)^{W-1}$ by the memoryless sojourn time and stationarity, and no singleton arrival in it. For $\beta=0.006$, $\gamma=0.03$, $\rho=0.0006$ and $W=60$ it gives $56.4\%$, and a simulation of $4\times10^5$ steps gives $57.6\%$ on average over three seeds and two classes (\S\ref{sec:exp4}). At the marginal prevalence of about $8\%$ of these classes, the i.i.d.\ formula~\eqref{eq:undefined_window_iid} predicts below $1\%$ undefined windows, so burstiness accounts for almost all of the undefined-window rate: the arrival process, and not only its marginal rate, governs how often the audited policies trigger.
\end{remark}

%% ---------------------------------------------------------------------------
\section{Verification Latency and Selective Feedback as Dynamic Masks}
\label{sec:delayed_labels}

Stream evaluation often assumes that the true label is available immediately after each prediction (null latency). Labels often arrive late or only for some instances: fraud is confirmed after a dispute resolves, a diagnosis after laboratory results, a loan default months after the decision, a forecast after its horizon. \citet{parreira2026ferds} call this regime intermediate verification latency.

Latency affects drift detectors such as DDM~\citep{gama2004learning}, ADWIN~\citep{bifet2007adwin} and HDDM~\citep{friasblanco2015hddm}, whose statistics are computed only over the verified sub-population. It also affects evaluation: \citet{parreira2026ferds} report, over eleven datasets and seven latency levels, that measured predictive performance degrades as latency increases. When the delay and the choice of which labels arrive do not depend on the model or on the label, the naive estimate over verified instances is only lagged (Table~\ref{tab:exp12}, setting~(ii)). Selection bias appears when what is observed depends on the model's decision or on the label.

We treat three feedback imperfections in one algebraic framework: temporal latency, endogenous censoring (labels withheld for instances the model rejects) and stochastic auditing (labels sought for a random or policy-driven subset). Each imposes a different structure on the observation mask $\mathcal{O}_k(t)$ defined next.

\subsection{Dynamic Mask Formulation for Latency, Censoring, and Auditing}
\label{sec:mask_formulation}
In operational stream environments labels rarely arrive with the prediction. We model the availability of a verified label for instance~$k$ at evaluation time $t$ by a binary observation mask:
\begin{equation}
    \mathcal{O}_k(t) = \mathbb{I}[k + \Delta_k \le t] \cdot A_k
    \label{eq:mask}
\end{equation}
where $\Delta_k \ge 0$ is the verification latency (possibly random) and $A_k \in \{0,1\}$ says whether a label is ever supplied for instance~$k$ ($A_k = 0$ is the same as $\Delta_k = \infty$). The calligraphic $\mathcal{O}$ is unrelated to the asymptotic $O(\cdot)$ notation of Proposition~\ref{prop:complexity}. The first factor gates temporal availability, since no label can be used before it arrives. The second says whether a label is sought at all. When the system decides which labels to request, $A_k$ is known at arrival; when an external process supplies the labels, $A_k$ is revealed only if the label arrives. Four regimes follow, two for the delay and two for the audit:

\begin{enumerate}
    \item Constant delay ($\Delta_k = \Delta$): all labels arrive with fixed latency, as in batch annotation pipelines or periodic audit cycles. The mask simplifies to $\mathcal{O}_k(t) = \mathbb{I}[k \le t - \Delta]$, creating a clean observation horizon with no stochastic component.

    \item Stochastic latency ($\Delta_k \sim \mathcal{P}(\lambda)$): labels arrive after a random delay drawn from a distribution such as the exponential or Poisson. This regime models asynchronous annotation queues, where the expected backlog at time $t$ is $\mathbb{E}[\text{unverified}] = \lambda$ instances. The effective number of verified instances up to time $t$ is $\sum_{k=1}^t \mathcal{O}_k(t)$, which varies stochastically around $t - \lambda$.

    \item Decision-dependent (endogenous) censoring: the audit probability depends on the model's own prediction,
    \begin{equation}
        \pi_{\mathrm{audit}}(k) = \pi_{\mathrm{audit}}(\mathbold{p}_k, \mathbold{x}_k)
        \label{eq:audit_prob}
    \end{equation}
    This arises in decision systems such as loan approval, where only accepted or flagged instances receive ground-truth feedback. The resulting selection bias couples the model's decision boundary to the observed evaluation set and invalidates the naive estimate.

    \item Stochastic (policy-driven) auditing: the audit decision is independent of the instance's features and prediction and follows an external labelling budget or sampling policy,
    \begin{equation}
        \pi_{\mathrm{audit}}(k) = b, \qquad b \in (0, 1], \qquad \pi_{\mathrm{audit}}(k) \perp (\mathbold{p}_k, \mathbold{x}_k)
        \label{eq:audit_budget}
    \end{equation}
    This models active-learning or manual-review budgets, where only a fraction $b$ of the stream is labelled under a random or round-robin policy rather than by the classifier's own decision. Because the audit gate is independent of $(\mathbold{p}_k, \mathbold{x}_k)$, the resulting verified sub-population is a representative sample of the full stream. Unlike decision-dependent censoring, naive evaluation restricted to it is unbiased up to sampling variance, at the cost of an effective sample size reduced to $n_{\mathrm{ess}}\, b$.
\end{enumerate}

\citet{gomes2022survey} organise label arrival along two axes: whether the labels that arrive do so immediately or after a delay, and whether every instance is eventually labelled. This gives four settings: (i)~immediate and fully labelled, (ii)~delayed and fully labelled, (iii)~immediate and partially labelled, (iv)~delayed and partially labelled. They are the four corners of the mask in Eq.~\eqref{eq:mask}, obtained by switching its two factors on or off (Table~\ref{tab:settings}). The delay axis is the pair of regimes 1 and 2 above; the survey's delay $D$ is a random variable with finite range, and the constant case is degenerate. The labelling axis is the pair 3 and 4.

The survey rests on the assumption that the learner has no influence on which labels arrive, and it describes missing labels with the missing-completely-at-random, missing-at-random and missing-not-at-random vocabulary of the missing-data literature. Policy-driven auditing (regime~4) has a gate independent of the data and is missing completely at random. Decision-dependent censoring (regime~3) lets the audit probability depend on $(\mathbold{x}_k,\mathbold{p}_k)$, so it is missing at random, with the model itself shaping the labelled subset; the survey sets that case aside under active learning. Neither mechanism depends on the true label $\mathbold{y}_k$ given $(\mathbold{x}_k,\mathbold{p}_k)$, which is what allows a calibrated estimate to stand in for the missing label (\S\ref{sec:product_tnorm}). The estimate needs the audit probability to be positive wherever it is used.

Setting~(iv) is the product of the two axes: any delay regime combined with any audit regime, the factors multiplying in Eq.~\eqref{eq:mask}. An unaudited instance has $\Delta_k=\infty$, which is how the survey describes the labels that never arrive; we keep the audit indicator as a separate factor because its mechanism, not its timing, determines the selection bias. One convention differs: the survey's ``immediate'' is a verification latency of one time unit, while here $\Delta_k=0$ means that the label is usable at evaluation time $k$, so the two coincide up to a unit offset. Algorithm~\ref{alg:update} handles the four settings without change, and \S\ref{sec:exp2} reports all four in one experiment.

\begin{table}[!ht]
\centering
\footnotesize
\setlength{\tabcolsep}{3pt}
\begin{tabular}{@{}llll@{}}
\toprule
\textbf{Setting} & \textbf{Delay factor $\mathbb{I}[k+\Delta_k\le t]$} & \textbf{Audit factor} & \textbf{Regimes used} \\
\midrule
(i) immediate, fully labelled & $\Delta_k=0$ & $\pi_{\mathrm{audit}}=1$ & none (Section~\ref{sec:theorems}) \\
(ii) delayed, fully labelled & $\Delta_k$ constant or random & $\pi_{\mathrm{audit}}=1$ & 1, 2 \\
(iii) immediate, partially labelled & $\Delta_k=0$ & $\pi_{\mathrm{audit}}<1$ & 3 or 4 (or both, Eq.~\eqref{eq:audit_composite}) \\
(iv) delayed, partially labelled & $\Delta_k$ constant or random & $\pi_{\mathrm{audit}}<1$ & (1 or 2) with (3 or 4) \\
\bottomrule
\end{tabular}
\caption{The four label-availability settings of \citet{gomes2022survey} as corners of the observation mask $\mathcal{O}_k(t)=\mathbb{I}[k+\Delta_k\le t]\cdot\mathbb{I}[k\text{ audited}]$.}
\label{tab:settings}
\end{table}

The censoring and auditing regimes often compose. \S\ref{sec:exp2} simulates this combination in a credit/risk stream: writing $\hat{c}_k \in \{0,1\}$ for the model's realized scalar decision on instance $k$ (distinct from the full prediction vector $\mathbold{p}_k$), instances the model approves ($\hat{c}_k=0$) are always eventually observed, while instances it rejects ($\hat{c}_k=1$) are censored unless separately selected by an independent audit budget $b$,
\begin{equation}
    \pi_{\mathrm{audit}}(k) = \mathbb{I}[\hat{c}_k = 0] + \mathbb{I}[\hat{c}_k = 1] \cdot U_k, \qquad U_k \overset{\mathrm{iid}}{\sim} \mathrm{Bernoulli}(b)
    \label{eq:audit_composite}
\end{equation}
composing Eq.~\eqref{eq:audit_prob}'s decision-dependent gate with Eq.~\eqref{eq:audit_budget}'s stochastic budget. The resulting evaluation set is neither purely censored nor an unbiased sample: it is fully observed for approved instances and $b$-sampled for rejected ones, and \S\ref{sec:product_tnorm} shows how the product $t$-norm interim estimator corrects the selection bias contributed by both components simultaneously.

The verified streaming contingency matrix at time $t$ integrates only observed, labelled instances:
\begin{equation}
    \mathbold{C}_{\mathrm{verified}}(t) = \sum_{k=1}^t \mathcal{O}_k(t)\, w_k(t)\, \mathbold{y}_k \mathbold{p}_k^T
    \label{eq:verified_cm}
\end{equation}
where $w_k(t)$ is the temporal weight assigned by $\Upsigma_{\mathrm{t}}$ to instance~$k$. Note that $\mathbold{C}_{\mathrm{verified}}(t)$ is sub-populational: it represents only a censored fraction of the stream, and any performance metric derived directly from it inherits the selection bias of the audit policy.

To manage latency without reprocessing history, the evaluator keeps a \emph{verification queue} $\mathcal{Q}(t)$ of the instances whose labels have not arrived and may still arrive, with what the correction needs: $\mathcal{Q}(t) = \{(k, \tilde{\mathbold{y}}_k, \mathbold{p}_k) : \mathcal{O}_k(t) = 0,\; \mathcal{E}_k = 1,\; k \text{ not yet evicted}\}$. The flag $\mathcal{E}_k \in \{0,1\}$ records what the evaluator knows when instance~$k$ arrives: $\mathcal{E}_k = 0$ if it already knows that no label will be requested for~$k$, and $\mathcal{E}_k = 1$ otherwise. Which instances are eligible depends on who controls the labelling:
\begin{itemize}
    \item When the system decides (active learning, decision-dependent censoring), the model's decision $\hat{c}_k$ and the audit draw $U_k$ of Eq.~\eqref{eq:audit_composite} are known at arrival, and the instances that are not selected have $\mathcal{E}_k = 0$. They contribute their interim term to $\mathbold{C}(t)$ at arrival (\S\ref{sec:product_tnorm}) and never enter the queue, since no correction can follow. The queue holds only the selected instances still waiting for a label, so its size is set by the latency and the audit budget, not by the stream length.
    \item When labels are supplied by an external process for a subset that is not known in advance (the partially labelled streams of \citet{gomes2022survey}, or fraud confirmations that may never come), eligibility cannot be known at arrival, so $\mathcal{E}_k = 1$ for every instance. Every instance is held until its label arrives or it is evicted, and eviction (\S\ref{sec:product_tnorm}) is what keeps the queue bounded when some labels never arrive.
\end{itemize}

\subsection{Marginally Consistent Interim Estimates via the Product $t$-norm}
\label{sec:product_tnorm}
Instances whose true labels have not arrived (those in the queue $\mathcal{Q}(t)$ and, when $\mathcal{E}_k=0$, those that never will) cannot contribute to $\mathbold{C}_{\mathrm{verified}}(t)$. Ignoring them leaves a hole in the state that grows with the latency. We fill it with an estimate $\tilde{\mathbold{y}}_k \in [0,1]^m$ of the label, obtained as described next, in place of the observed label $\mathbold{y}_k \in \{0,1\}^m$. We call it the interim estimate or, in the experiments, the stand-in. We write $\tilde{\mathbold{y}}_k$, not $\hat{\mathbold{y}}_k$, to avoid clashing with the prediction notation of \S\ref{sec:temporal_im}: $\tilde{\mathbold{y}}_k$ stands in for the unavailable true label, while $\mathbold{p}_k$ remains the model's prediction being evaluated.

The estimate is a calibrated function of the model's score and needs no data beyond the stream. Let $\mathbold{s}_k$ be the score vector the model produces for instance~$k$, from which its decision $\mathbold{p}_k$ is taken (for instance the class probabilities of a Hoeffding tree), and let $g_t$ be a calibration map fitted at time~$t$. Then $\tilde{\mathbold{y}}_k = g_t(\mathbold{s}_k)$ is computed once, at arrival, and needs no label of instance~$k$. The map $g_t$ is the only learned component. It is fitted on the pairs $(\mathbold{s}_j, \mathbold{y}_j)$ of the instances whose labels have arrived by time~$t$ (the verified set of Eq.~\eqref{eq:verified_cm}), with $\mathbold{s}_j$ the score computed at the time of prediction, and it is updated online as each label arrives, optionally with the same fading or window weights as $\Upsigma_{\mathrm{t}}$ so that it follows drift. Platt scaling, isotonic regression or temperature scaling can serve as $g_t$. Four points follow.
\begin{enumerate}
    \item The raw score is not a safe substitute. A classifier's own probabilities are often over- or under-confident, and Eq.~\eqref{eq:bias} below ties the bias of the estimator to that error.
    \item The verified set is a selected sample. Under regimes 1 to 4 the mechanism that decides which labels arrive does not depend on the true label $\mathbold{y}_k$ given $(\mathbold{x}_k,\mathbold{p}_k)$. Because $g_t$ estimates $P(Y \mid \mathbold{s})$, it is consistent on the verified set only if the audit probability depends on the instance through the score $\mathbold{s}$, as it does when it depends on the decision $\mathbold{p}_k$, and is positive wherever the estimate is used. If it depends on other features of $\mathbold{x}_k$, $g_t$ is biased. Under decision-dependent censoring the flagged region reaches $g_t$ only through the audit budget $b$, so with $b=0$ that region is never observed and the correction cannot be validated there.
    \item Under delay, $g_t$ lags the stream by the latency, since it only sees labels that have arrived. If the delay depends on the label (the asymmetric regime of \S\ref{sec:exp2}), the labels that have arrived over-represent the class confirmed first, $g_t$ inherits that bias, and the correction is only partial; \S\ref{sec:exp2} measures it.
    \item The value $\tilde{\mathbold{y}}_k$ used at arrival must be stored, not recomputed from the current $g_t$, so that the correction step subtracts exactly the term it added.
\end{enumerate}

The soft contribution of an unverified instance $(\mathbold{x}_k, \tilde{\mathbold{y}}_k, \mathbold{p}_k)$ to the contingency matrix is computed by a bivariate aggregation operator $T: [0,1]^2 \to [0,1]$ applied entry-wise:
\begin{equation}
    \mathbold{C}_{\mathrm{interim}, k} = T(\tilde{\mathbold{y}}_k,\, \mathbold{p}_k) \in [0,1]^{m \times m}
    \label{eq:soft_cm}
\end{equation}
We ask for \emph{marginal consistency}: the row and column sums of the soft contribution must match $\tilde{\mathbold{y}}_k$ and $\mathbold{p}_k$,
\begin{align}
    \sum_{j'} [T(\tilde{\mathbold{y}}_k,\, \mathbold{p}_k)]_{jj'} &= \tilde{y}_{kj} \qquad (\text{marginal label consistency}) \label{eq:row_marg}\\
    \sum_{j} [T(\tilde{\mathbold{y}}_k,\, \mathbold{p}_k)]_{jj'} &= p_{kj'} \qquad (\text{marginal prediction consistency}) \label{eq:col_marg}
\end{align}
so that the soft matrix contributes the expected totals to the TP, FP, TN and FN cells. Among $t$-norms, only the product $T_P(a, b) = a \cdot b$ satisfies both conditions for soft predictions~\citep{prati2026unified}, and then
\begin{equation}
    [T_P(\tilde{\mathbold{y}}_k,\, \mathbold{p}_k)]_{jj'} = \tilde{y}_{kj}\, p_{kj'},
    \label{eq:product_tnorm}
\end{equation}
the outer product $\tilde{\mathbold{y}}_k \mathbold{p}_k^T$. The minimum $T_M(a,b) = \min(a,b)$ and the \L{}ukasiewicz $T_L(a,b) = \max(0, a+b-1)$ $t$-norms violate at least one condition, with marginal defect
\begin{equation}
    \mathrm{defect}(T) = \left\| \mathbold{1}^T [T(\tilde{\mathbold{y}}_k,\mathbold{p}_k)] - \mathbold{p}_k^T \right\|_1 + \left\| [T(\tilde{\mathbold{y}}_k,\mathbold{p}_k)] \mathbold{1} - \tilde{\mathbold{y}}_k \right\|_1
    \label{eq:defect}
\end{equation}
that is positive for soft $\mathbold{p}_k$ (Table~\ref{tab:exp2}) and zero for $T_P$. For hard one-hot decisions $\mathbold{p}_k$ every $t$-norm gives the same matrix, because $T(a,1)=a$ and $T(a,0)=0$, so the choice only matters for soft predictions. The product is also the right choice for a second reason: the prediction $\mathbold{p}_k$ is a function of $\mathbold{x}_k$ once the model is fixed, so $\mathbb{E}[\mathbold{y}_k \mathbold{p}_k^T \mid \mathbold{x}_k] = P(Y \mid \mathbold{x}_k)\,\mathbold{p}_k^T$, which is an outer product.

The unified streaming contingency matrix combines verified labels with interim estimates:
\begin{equation}
    \mathbold{C}(t) = \underbrace{\sum_{k=1}^t \mathcal{O}_k(t)\, w_k(t)\, \mathbold{y}_k \mathbold{p}_k^T}_{\text{verified contribution}} \;+\; \underbrace{\sum_{k=1}^t (1 - \mathcal{O}_k(t))\, w_k(t)\, \tilde{\mathbold{y}}_k \mathbold{p}_k^T}_{\text{interim contribution}}
    \label{eq:unified_cm}
\end{equation}
It recovers $\mathbold{C}_{\mathrm{verified}}(t)$ in the limiting case of zero latency ($\Delta_k = 0$, $A_k = 1$) and is built from outer products with the prescribed marginals. The zero marginal defect is a property of that construction. Correcting a selection bias is a separate matter that needs a mechanism that is missing at random given the score and a consistent calibration map $g_t$ (point~2 above), and it fails when the delay depends on the label.

Equation~\eqref{eq:unified_cm} is not recomputed from scratch. Algorithm~\ref{alg:update} credits the interim term at arrival and swaps it for the verified term if the label later arrives, so that both live in one running matrix.

\begin{algorithm}[!ht]
\caption{Prequential state update on arrival of instance $t+1$ (exponential fading)}
\label{alg:update}
\begin{algorithmic}[1]
\Require State $\mathbold{C}(t)$, queue $\mathcal{Q}(t)$, score $\mathbold{s}_{t+1}$, prediction $\mathbold{p}_{t+1}$, eligibility flag $\mathcal{E}_{t+1}$, and the labels $\mathbold{y}_k$ that arrive at time $t+1$
\Ensure Updated state $\mathbold{C}(t+1)$ and queue $\mathcal{Q}(t+1)$
\State $\tilde{\mathbold{y}}_{t+1} \gets g_t(\mathbold{s}_{t+1})$ \Comment{interim estimate, stored with the entry}
\State $\mathbold{C}(t+1) \gets \alpha\,\mathbold{C}(t) + (1-\alpha)\, T_P(\tilde{\mathbold{y}}_{t+1}, \mathbold{p}_{t+1})$ \Comment{decay, then credit the interim term}
\If{$\mathcal{E}_{t+1} = 1$} \State enqueue $(t{+}1, \tilde{\mathbold{y}}_{t+1}, \mathbold{p}_{t+1})$ into $\mathcal{Q}$ \EndIf
\ForAll{labels $\mathbold{y}_k$ arriving at time $t+1$ with $k \in \mathcal{Q}$}
    \State $\mathbold{C}(t+1) \gets \mathbold{C}(t+1) + (1-\alpha)\,\alpha^{\,t+1-k}\bigl(\mathbold{y}_k \mathbold{p}_k^T - T_P(\tilde{\mathbold{y}}_k, \mathbold{p}_k)\bigr)$ \Comment{swap interim for verified}
    \State remove $k$ from $\mathcal{Q}$
\EndFor
\While{the oldest entry $k$ of $\mathcal{Q}$ has $(1-\alpha)\,\alpha^{\,t+1-k} < \varepsilon$}
    \State remove $k$ from $\mathcal{Q}$ \Comment{its interim term stays in $\mathbold{C}$ and keeps decaying}
\EndWhile
\State \Return $\mathbold{C}(t+1)$, $\mathcal{Q}(t+1)$
\end{algorithmic}
\end{algorithm}

Each step costs $O(m^2)$ operations, plus $O(m^2)$ for each arriving label, independent of $t$ and of $|\mathcal{Q}|$: the weight of an entry is computed from its age when a label arrives, and eviction only inspects the oldest entry. Under a sliding window the weights are $1/W$ and the update is $\mathbold{C}(t+1) \gets \mathbold{C}(t) + \frac{1}{W}\bigl(T_P(\tilde{\mathbold{y}}_{t+1}, \mathbold{p}_{t+1}) - \text{term of the instance leaving the window}\bigr)$. A FIFO of the last $W$ instances stores each instance's current term, interim or verified, so that a label arriving within the window replaces the stored term and adds $\frac{1}{W}(\mathbold{y}_k \mathbold{p}_k^T - T_P(\tilde{\mathbold{y}}_k, \mathbold{p}_k))$ to $\mathbold{C}$. Eviction is then automatic, $|\mathcal{Q}(t)| \le W$. The window variant and the eviction step are implemented in the package (\S\ref{sec:software}) and run in \S\ref{sec:exp2} (Table~\ref{tab:window}).

Eviction bounds the queue. Writing the weight of an entry of age $s=t-k$ as $w(s) = (1-\alpha)\alpha^s$, an entry is evicted once $w(s) < \varepsilon$, that is after $s^* = \min\{s : w(s) < \varepsilon\}$ steps. The entries that can no longer be corrected carry a share of the state of at most
\begin{equation}
    \sum_{s \ge s^*} w(s) \;=\; \alpha^{s^*} \;<\; \varepsilon\, n_{\mathrm{eff}},
    \qquad s^* = \min\{s : w(s) < \varepsilon\},
    \label{eq:truncation_bound}
\end{equation}
an exact geometric tail (with $n_{\mathrm{eff}} = 1/(1-\alpha)$) reached after $s^* = O\!\left(n_{\mathrm{eff}} \log\frac{1}{n_{\mathrm{eff}}\varepsilon}\right)$ steps. Hence $|\mathcal{Q}(t)| \le s^\ast \le n_{\mathrm{eff}} \ln\frac{1}{n_{\mathrm{eff}}\varepsilon} + 2$ at any time when $n_{\mathrm{eff}}\varepsilon < 1$, independent of $t$ and of how many instances are permanently censored. The wait limit comes from the temporal operator and not from the labelling process: a label that arrives after the eviction point would be multiplied by a weight below $\varepsilon$ and changes $\mathbold{C}(t)$ by less than $\varepsilon$, although it can still be used elsewhere, for instance to train the model and $g_t$. The correction therefore pays off only when the latency is short against the memory of the operator, $\Delta_k \ll W$ under a window or $\Delta_k \ll n_{\mathrm{eff}}$ under fading. If $\Delta_k \ge W$ for most instances, every instance leaves the window unverified and the windowed state consists of interim terms only, that is, of the model's own calibrated belief. Table~\ref{tab:exp12} is in the first regime ($\lambda=25$, $n_{\mathrm{eff}}=100$), and the effect of delay there is correspondingly small. The queue costs $O(|\mathcal{Q}|\,m)$ memory on top of the $O(m^2)$ state of Proposition~\ref{prop:complexity}.

Let $\mathbold{C}_{\mathrm{oracle}}(t)$ be the state that would be obtained if every label were available at arrival. The bias of the unified estimator~\eqref{eq:unified_cm} relative to it is
\begin{equation}
    \mathbold{C}_{\mathrm{oracle}}(t) - \mathbold{C}(t) = \sum_{k=1}^t (1 - \mathcal{O}_k(t))\, w_k(t)\, (\mathbold{y}_k - \tilde{\mathbold{y}}_k)\, \mathbold{p}_k^T
    \label{eq:bias}
\end{equation}
The residual is driven by the calibration error of $\tilde{\mathbold{y}}_k$. If $\mathbb{E}[\mathbold{y}_k \mid \mathbold{s}_k, \mathbold{p}_k] = \tilde{\mathbold{y}}_k$ (calibration) and $\mathcal{O}_k(t)$ is independent of $\mathbold{y}_k$ given $(\mathbold{s}_k,\mathbold{p}_k)$, the expected residual is zero and $\mathbold{C}(t)$ is conditionally unbiased for the oracle matrix. Equations~\eqref{eq:unified_cm} and~\eqref{eq:bias} are exact without eviction; eviction changes each entry of $\mathbold{C}(t)$ by less than $\varepsilon$ for each late label it discards, and by at most $\alpha^{s^\ast} < \varepsilon\, n_{\mathrm{eff}}$ in total (Eq.~\eqref{eq:truncation_bound}), when the entries of $\mathbold{y}_k\mathbold{p}_k^T - \tilde{\mathbold{y}}_k\mathbold{p}_k^T$ are bounded by one. Metrics that are ratios of cells inherit only first-order consistency from this. Post-hoc calibration (isotonic regression~\citep{zadrozny2002transforming}, Platt scaling~\citep{platt1999probabilistic}, temperature scaling~\citep{guo2017calibration}) reduces the calibration error that drives the residual, which links classifier calibration to evaluation bias under delayed labels.

\begin{example}[Running example: delayed labels, a stand-in and its correction]
\label{ex:running_delay}
Let the label of instance $k$ arrive at time $k+\Delta_k$ with $\Delta=(0,1,2,3,2,3,2,1)$, and let instance $7$ never be labelled, because the model flagged it as $C$ and the system requested no label ($A_7=0$, $\mathcal{E}_7=0$). At $t=8$ the window holds instances $5$ to $8$: the label of $5$ has arrived, those of $6$ and $8$ are pending (they arrive at $t=9$), and $7$ is never labelled. Only $6$ and $8$ enter the queue. The stand-in puts $\tfrac34$ on the predicted class and $\tfrac18$ on each other class. With weights $\tfrac14$ the unified state of Eq.~\eqref{eq:unified_cm} and the oracle state are
\[
\mathbold{C}(8)=\tfrac1{32}\begin{pmatrix}6&1&1\\1&14&1\\1&1&6\end{pmatrix},\qquad
\mathbold{C}_{\mathrm{oracle}}(8)=\tfrac1{32}\begin{pmatrix}8&0&0\\0&16&8\\0&0&0\end{pmatrix},
\]
and their difference, $\tfrac1{32}\bigl(\begin{smallmatrix}2&-1&-1\\-1&2&7\\-1&-1&-6\end{smallmatrix}\bigr)$, is the sum of $(\mathbold{y}_k-\tilde{\mathbold{y}}_k)\mathbold{p}_k^T/4$ over the instances without a label (Eq.~\eqref{eq:bias}).

The recall of $B$ is $\tfrac23$ in the oracle state, $\tfrac78$ in $\mathbold{C}(8)$ and $1$ if only the verified instance is used; the recall of $A$ is $1$, $\tfrac34$ and undefined.

For the stand-in of instance $7$ and the soft prediction $(\tfrac18,\tfrac18,\tfrac34)$ the marginal defect is $0$ for the product $t$-norm, $\tfrac32$ for the minimum and $1$ for the \L{}ukasiewicz $t$-norm; for the hard prediction it is $0$ for all three.

When the label of instance $6$ arrives, the correction adds $\tfrac14(\mathbold{y}_6\mathbold{p}_6^T-T_P(\tilde{\mathbold{y}}_6,\mathbold{p}_6))=\tfrac1{32}\bigl(\begin{smallmatrix}2&0&0\\-1&0&0\\-1&0&0\end{smallmatrix}\bigr)$. Under fading with $\alpha=\tfrac12$ the weights of instances $6$ and $8$ at $t=8$ are $\tfrac18$ and $\tfrac12$, and with $\varepsilon=\tfrac1{32}$ an entry is evicted once its age exceeds $s^\ast=5$ steps, which no queue entry has reached.

The stand-in also hides a void on the label side: it puts mass $\tfrac14$ in the row of $C$, so $\mathrm{Recall}_C=\tfrac34$ in $\mathbold{C}(8)$ although the oracle state has no support for $C$. The verified support is $\mathbold{u}^{\mathrm{ver}}=(0,1,0)$ and the coverage $\mathbold{\kappa}=(0,\tfrac12,0)$ for the label-side denominators, so the policies of \S\ref{sec:zero_division} treat $A$ and $C$ as unsupported for recall: the macro recall is $\tfrac78$ under $\mathtt{exclude}$, $\tfrac7{24}$ under $\mathtt{zero}$ and undefined under $\mathtt{nan}$, and the $\tfrac78$ is an estimate ($\kappa_B=\tfrac12$). The columns are exact: the column sums of $\mathbold{C}(8)$ equal those of the oracle, $(\tfrac14,\tfrac12,\tfrac14)$, so the precision of every class keeps $u^{\mathrm{int}}=(1,1,1)$.
\end{example}

%% ---------------------------------------------------------------------------
\section{Disentangling Concept Drift from Prior Probability Shifts}
\label{sec:drift_decomposition}

Non-stationarity in a stream has two statistical mechanisms. In \emph{real concept drift} the conditional distribution $P_t(Y \mid X)$ changes with the environment. In \emph{virtual drift} the feature distribution $P_t(X)$ or the class prior $P_t(Y) = \mathbold{\pi}(t)$ changes while the decision boundary stays valid. Metrics such as Accuracy, Precision and $F_1$ depend on the prior $\mathbold{\pi}(t)$, so a change in class balance moves them even when the classifier is unchanged, and drift detectors that monitor them or the error rate (DDM~\citep{gama2004learning}, ADWIN~\citep{bifet2007adwin}) can alarm on a prior shift alone.

This section works under label shift: a prior change means $P_t(Y)$ changes while $P_t(X \mid Y)$ stays fixed. We give the temporal taxonomy of drifts (\S\ref{sec:drift_taxonomy}), the canonical coordinates that are free of the prior and the conditions under which this holds in the binary and the multiclass case (\S\ref{sec:canonical_space}), a first-order decomposition of a metric change into a concept term and a prior term (\S\ref{sec:jacobian_decomposition}), and a monitoring scheme that detects on the canonical coordinates and attributes on the decomposition (\S\ref{sec:monitoring}).

\subsection{Temporal Drift Taxonomy and Dynamic Filtering Dynamics}
\label{sec:drift_taxonomy}
Drift events have different temporal profiles, which determine how quickly a learner must adapt and how $\Upsigma_{\mathrm{t}}$ filters the history.

\begin{enumerate}
    \item Abrupt drift is an instantaneous transition between two generating distributions $P_A$ and $P_B$ at time $t_0$, $P_t(X, Y) = (1 - H(t - t_0)) P_A + H(t - t_0) P_B$, with $H$ the Heaviside step. Under exponential fading, in expectation and if the state had settled at $\mathbold{C}_A$ before $t_0$, the state moves as
    \begin{equation}
    \mathbold{C}_\alpha(t) = \alpha^{t - t_0 + 1} \mathbold{C}_A + (1 - \alpha^{t - t_0 + 1}) \mathbold{C}_B, \qquad \forall t \ge t_0
    \end{equation}
    The adaptation timescale is the half-life $t_{1/2} = \ln 2 / \ln(1/\alpha) \approx 0.693\, n_{\mathrm{eff}}$, whereas a sliding window purges the old concept after exactly $W$ steps.

    \item Gradual drift is a probabilistic mixture over a finite duration $w_{\mathrm{drift}}$ centred at $t_0$, $P_t(X, Y) = (1 - s(t)) P_A + s(t) P_B$ with $s(t) = 1/(1 + \exp(-4 (t - t_0) / w_{\mathrm{drift}}))$. Samples from both concepts interleave during the transition.

    \item Incremental drift is a smooth parameter trajectory $\mathbold{\theta}(t) = \mathbold{\theta}_0 + \mathbold{v}_{\mathrm{drift}} (t - t_0)$. Under a steady velocity $\mathbold{v}_{\mathbold{C}} = d\mathbold{C}_{\mathrm{pop}}(t)/dt$ the fading estimator has a constant asymptotic tracking lag,
    \begin{equation}
    \mathbold{C}_\alpha(t) - \mathbold{C}_{\mathrm{pop}}(t) \approx - \left( \frac{\alpha}{1 - \alpha} \right) \mathbold{v}_{\mathbold{C}} = - (n_{\mathrm{eff}} - 1) \mathbold{v}_{\mathbold{C}}
    \end{equation}
    which is the bias half of the trade-off between smoothing noise and tracking latency.

    \item Reoccurring concepts are periodic or stochastic reappearances of earlier distributions $P_t \in \{P_1, \dots, P_K\}$. Because $\Upsigma_{\mathrm{t}}$ discounts by recency and not by identity, a returning concept enters $\mathbold{C}(t)$ like a new one, so reusing a specialised sub-model needs machinery outside $\Upsigma_{\mathrm{t}}$.

    \item Transient outliers are isolated single-instance perturbations $\delta(t - t_0)$. The fading operator attenuates such a shock to an amplitude of $(1-\alpha)$ and damps it as $\alpha^s$.
\end{enumerate}

\subsection{The Canonical Coordinates and Prior-Invariance}
\label{sec:canonical_space}
We call $(\mathrm{Recall}, \mathrm{Specificity})$ canonical because every binary measure is a function of them and of the prior. For a binary stream, or for the table of one class $j$, write $\pi_t$ for the positive-class prior at time $t$ and $\mathbold{R}(t) = [R_{11}(t), R_{00}(t)]^T = [\mathrm{Recall}(t), \mathrm{Specificity}(t)]^T$ for the canonical coordinates (the true positive and true negative rates of the classifier at time $t$). In terms of the cell masses of that table,
\begin{equation}
\label{eq:canonical_coords}
R_{11} = \frac{A}{A + C}, \qquad R_{00} = \frac{D}{D + B},
\end{equation}
where $A, B, C, D$ are the true positive, false positive, false negative and true negative masses. These are population quantities at time $t$ when the cells are the expected cell masses at $t$. A cell mass that is a time-weighted accumulation needs the care of Remark~\ref{rem:accumulated} below.

\begin{lemma}[A ratio of proportional cells is prior-invariant]
\label{lem:cell_ratio_prior_invariant}
If two nonnegative quantities $A, C$ satisfy $A = c \cdot r$ and $C = c \cdot (1-r)$ for some $c \ne 0$ and some $r$, then $A/(A+C) = r$, independently of $c$.
\end{lemma}
\begin{proof}
$A + C = c\cdot r + c\cdot(1-r) = c$, so $A/(A+C) = (c\cdot r)/c = r$.
\end{proof}

\begin{proposition}[Prior-invariance of the canonical coordinates at a fixed time]
\label{prop:prior_invariance}
Let $f_t: \mathcal{X} \to [0,1]$ be the classifier's decision function at time $t$ (distinct from the prediction vector $\mathbold{p}_t$, which is $f_t$'s output on one instance), and let $P_t(X \mid Y=1)$, $P_t(X \mid Y=0)$ be the class-conditional feature distributions. Both may vary arbitrarily with $t$: the classifier may be fixed, periodically retrained or continuously adapted. Then, for $0 < \pi_t < 1$, the canonical coordinates at time $t$ do not depend on the prior $\pi_t = P_t(Y=1)$:
\begin{equation}
\label{eq:prior_invariance_closed_form}
R_{11}(t) = \int f_t(x)\, dP_t(x \mid Y=1), \qquad R_{00}(t) = \int \bigl(1 - f_t(x)\bigr)\, dP_t(x \mid Y=0)
\end{equation}
Neither expression contains $\pi_t$, so $\partial R_{11}/\partial \pi_t = \partial R_{00}/\partial \pi_t = 0$. By contrast Precision and $F_1$ depend on $\pi_t$ even without concept drift, and so does Accuracy unless $R_{11} = R_{00}$.
\end{proposition}
\begin{proof}
By the law of total probability at the fixed time $t$, the expected cell masses are $A = \pi_t \int f_t\, dP_t(\cdot \mid Y=1)$ and $C = \pi_t \int (1-f_t)\, dP_t(\cdot \mid Y=1)$. Both are $\pi_t$ times a proportion of the positive-class distribution, so Lemma~\ref{lem:cell_ratio_prior_invariant} with $c = \pi_t$ gives $R_{11}(t) = \int f_t\, dP_t(\cdot \mid Y=1)$. The same lemma with $c = 1-\pi_t$ gives $R_{00}(t)$.
\end{proof}

\begin{remark}[Accumulated cells are not instantaneous coordinates]
\label{rem:accumulated}
Proposition~\ref{prop:prior_invariance} concerns the coordinates at one time. The cells of the fading state $\mathbold{C}_\alpha(t)$ accumulate over the memory of the operator, and the accumulated recall is
\begin{equation}
\widehat{R}^{\mathrm{acc}}_{11}(t) = \frac{\sum_{s} w_s\, \pi_s\, R_{11}(s)}{\sum_{s} w_s\, \pi_s},
\end{equation}
a prior-weighted average of past recalls. It is free of the prior if $R_{11}(s)$ is constant over the memory window or if $\pi_s$ is, and not in general otherwise. When the classifier changes while the prior changes, the accumulated coordinates mix the two. Call the cells of $\mathbold{C}_\alpha(t)$ the time-fading coordinates. To estimate the instantaneous coordinates without this mixing, decay by the arrivals of each class instead of by time (class-time fading): update $\widehat{R}_{11} \leftarrow \alpha\, \widehat{R}_{11} + (1-\alpha)\, \mathbb{I}[\text{prediction positive}]$ only on arrivals with $Y=1$, and $\widehat{R}_{00}$ only on arrivals with $Y=0$ (with the usual correction for the initial mass). Each estimate is then a function of its own class arrivals alone, so its mean is free of the prior by construction, and its memory is $n_{\mathrm{eff}}$ arrivals of that class instead of $n_{\mathrm{eff}}$ steps. Its variance still depends on the number of arrivals of the class, hence on the prior.
\end{remark}

\begin{proposition}[Prior-invariance in the multiclass case]
\label{prop:multiclass_invariance}
Let $c_{kl}(t) = P_t(\hat{Y}=l \mid Y=k)$ be the rows of the confusion matrix and $\pi_k(t) = P_t(Y=k)$ the class priors. Then
\begin{enumerate}
    \item the rows $c_k(t)$, and in particular $\mathrm{Recall}_j(t) = c_{jj}(t)$, do not depend on $\mathbold{\pi}(t)$;
    \item the one-vs-rest specificity is $\mathrm{Specificity}_j(t) = \sum_{k \ne j} \frac{\pi_k(t)}{1-\pi_j(t)}\,\bigl(1 - c_{kj}(t)\bigr)$, which depends on the prior only through the composition $\pi_k/(1-\pi_j)$, $k \ne j$, of the other classes. It is invariant to every prior change that preserves that composition (for example a change of $\pi_j$ alone, the other classes rescaled), and, if column $j$ of the confusion matrix is constant over $k \ne j$, to every prior change. Otherwise it moves with the priors of the other classes without any concept drift.
\end{enumerate}
\end{proposition}
\begin{proof}
Part 1 is the law of total probability, as in Proposition~\ref{prop:prior_invariance}. For part 2, $P(\hat{Y} \ne j \mid Y \ne j) = \sum_{k \ne j} P(Y=k \mid Y \ne j)\, P(\hat{Y} \ne j \mid Y=k)$ and $P(Y=k \mid Y \ne j) = \pi_k/(1-\pi_j)$.
\end{proof}
The binary case is the special case in which the rest is a single class. For multiclass streams the invariant coordinates are therefore the confusion rows $P(\hat{Y} \mid Y=k)$, one stream of outcomes per true class, and not the one-vs-rest pair $(\mathrm{Recall}_j, \mathrm{Specificity}_j)$.

\begin{example}[Running example: a label shift among the other classes]
\label{ex:running_shift}
Consider the population in which the classifier has the confusion rows $c_A=(\tfrac34,\tfrac14,0)$, $c_B=(0,\tfrac34,\tfrac14)$, $c_C=(\tfrac14,0,\tfrac34)$ (each class is mistaken for the next one), and the prior moves from $(\tfrac12,\tfrac14,\tfrac14)$ to $(\tfrac12,\tfrac18,\tfrac38)$, a label shift between $B$ and $C$ that leaves $A$ unchanged. The rows are unchanged, so $\mathrm{Recall}_A=\tfrac34$ before and after. The one-vs-rest specificity is $1-\tfrac{\pi_C}{1-\pi_A}c_{CA}$, which is $\tfrac78$ before and $\tfrac{13}{16}$ after, and $\mathrm{Precision}_A$ moves from $\tfrac67$ to $\tfrac45$. The classifier has not changed, yet $\mathrm{Specificity}_A$ moved, because the composition of the rest changed (Proposition~\ref{prop:multiclass_invariance}).

For Remark~\ref{rem:accumulated}, suppose that two steps ago the recall of $A$ fell from $\tfrac34$ to $\tfrac12$ while its prior fell from $\tfrac12$ to $\tfrac18$. With fading $\alpha=\tfrac12$ and a memory of four steps the accumulated recall is $\tfrac58$, against an instantaneous recall of $\tfrac12$ and a plain weighted mean of the two recalls of $\tfrac{11}{20}$: the accumulation over-weights the old phase, in which $A$ was frequent.
\end{example}

\begin{remark}[Continually adapting classifiers and empirical confounding]
\label{rem:continual_learning}
Proposition~\ref{prop:prior_invariance} places no restriction on $f_t$, so it holds whether the classifier is frozen, periodically retrained or updated after every instance. The Jacobian decomposition of \S\ref{sec:jacobian_decomposition} therefore needs no modification for continually learning classifiers: its concept term captures the combined rate of change from environmental concept drift ($P_t(X \mid Y)$ evolving) and classifier self-adaptation ($f_t$ evolving), and $\pi_t$ contributes nothing to that term. Two cautions follow. First, zero functional dependence does not imply zero sample correlation between $\mathrm{Recall}(t)$ and $\pi_t$ over a realised trajectory: if $f_t$ and $\pi_t$ are driven by the same non-stationary stream, as when a classifier adapts to recent label proportions, they can co-vary. Second, the accumulated coordinates mix the prior with the history of the recall (Remark~\ref{rem:accumulated}), which also produces residual correlations with the prior in a real stream.
\end{remark}

\subsection{Analytical Jacobian Drift Decomposition}
\label{sec:jacobian_decomposition}
A composite metric can be written as a differentiable function $M(\mathbold{R}, \pi)$ of the canonical coordinates and the prior (not to be confused with the decision function $f_t$). Differentiating along the trajectory of the stream gives, in continuous time,
\begin{equation}
\label{eq:jacobian_decomposition}
\frac{d M}{d t} = \underbrace{\mathbold{J}_{\mathrm{concept}}^T \frac{d\mathbold{R}}{dt}}_{\text{concept term}} + \underbrace{J_{\mathrm{prior}} \frac{d\pi}{dt}}_{\text{prior term}}
\end{equation}
where $\mathbold{J}_{\mathrm{concept}} = [\partial M/\partial R_{11}, \partial M/\partial R_{00}]^T$ and $J_{\mathrm{prior}} = \partial M/\partial \pi$. The decomposition reparametrises the change of $M$ in the coordinates $(R_{11}, R_{00}, \pi)$; the concept term isolates concept drift only to the extent that $\mathbold{R}$ is free of the prior (Proposition~\ref{prop:prior_invariance}). For the standard measures the Jacobians are, with $D$ the denominator of the measure in each case:
\begin{itemize}
    \item Precision, $M = \pi R_{11}/(\pi R_{11} + (1-\pi)(1-R_{00}))$:
    \begin{equation}
    \mathbold{J}_{\mathrm{concept}} = \begin{bmatrix} \frac{\pi(1-\pi)(1-R_{00})}{D^2} \\ \frac{\pi(1-\pi)R_{11}}{D^2} \end{bmatrix}, \qquad J_{\mathrm{prior}} = \frac{R_{11}(1-R_{00})}{D^2}
    \end{equation}
    \item $F_1$, $M = 2 \pi R_{11}/(\pi (1 + R_{11}) + (1 - \pi) (1 - R_{00}))$:
    \begin{equation}
    \mathbold{J}_{\mathrm{concept}} = \begin{bmatrix} \dfrac{2\pi\bigl[\pi + (1-\pi)(1-R_{00})\bigr]}{D^2} \\[1.2ex] \dfrac{2\pi(1-\pi)R_{11}}{D^2} \end{bmatrix}, \qquad J_{\mathrm{prior}} = \frac{2 R_{11} \bigl(1 - R_{00}\bigr)}{D^2}
    \end{equation}
    \item Accuracy, $M = \pi R_{11} + (1-\pi) R_{00}$: $\mathbold{J}_{\mathrm{concept}} = [\pi, 1-\pi]^T$ and $J_{\mathrm{prior}} = R_{11} - R_{00}$.
    \item Balanced accuracy, $M = (R_{11} + R_{00})/2$: $\mathbold{J}_{\mathrm{concept}} = [1/2, 1/2]^T$ and $J_{\mathrm{prior}} = 0$.
\end{itemize}

Here $\mathbold{R}$ is the time-fading state, whose prior-dependence is that of Remark~\ref{rem:accumulated}, which the path integration below partly absorbs. For a change between two states of the stream, the decomposition is a first-order statement. Write $\Delta M = M(\mathbold{R}_1, \pi_1) - M(\mathbold{R}_0, \pi_0)$. Evaluating the Jacobians at the midpoint gives
\begin{equation}
\label{eq:attribution_mid}
\Delta M \approx \mathbold{J}_{\mathrm{concept}}(\bar{\mathbold{R}}, \bar{\pi})^T \Delta\mathbold{R} + J_{\mathrm{prior}}(\bar{\mathbold{R}}, \bar{\pi})\, \Delta\pi,
\end{equation}
a quadrature of the line integral of $\nabla M$ whose error grows with the size of the change. It is exact for a metric that is bilinear in $(\mathbold{R}, \pi)$ (Accuracy) and inaccurate for large prior changes of $F_1$ and Precision, where $M$ is strongly curved in $\pi$. For large changes, summing the midpoint terms over intermediate states between the two end states (path integration) restores accuracy. When both $\mathbold{R}$ and $\pi$ change, the split of $\Delta M$ into a concept and a prior part depends on the order in which the two changes are applied (an interaction term), so the concept term of Eq.~\eqref{eq:attribution_mid} lies near the two orderings (exactly between them for Accuracy). The sensitivity of $F_1$ to specificity is $2\pi(1-\pi)R_{11}/D^2$, which vanishes as $\pi \to 0$ at fixed $R_{00} < 1$: a concept change in specificity moves $F_1$ little when positives are rare.

\begin{example}[Running example: attribution of the change]
\label{ex:jacobian}
For class $A$ in Example~\ref{ex:running_shift}, $F_1$ moves from $\tfrac45$ to $\tfrac{24}{31}$, a change of $-\tfrac4{155}\approx-0.0258$. The prior $\pi_A$ is unchanged, so the prior term is $0$, and the midpoint Jacobian gives a concept term of $-\tfrac{96}{3721}\approx-0.0258$ with a residual of $7\times10^{-6}$. The decomposition attributes the whole change to the concept although no concept drift occurred: the movement of $\mathrm{Specificity}_A$ comes from the other classes (Proposition~\ref{prop:multiclass_invariance}). Computed on the confusion rows, which do not move, the concept term is $0$.
\end{example}

\subsection{Monitoring the Canonical Coordinates and Attributing Alarms}
\label{sec:monitoring}
The decomposition answers a question about one metric: how much of an observed change of $M$ is due to the concept and how much to the prior. It is not a general detector of concept change. Its concept term combines the two coordinates with signs, so changes of $R_{11}$ and $R_{00}$ that offset in $M$ cancel in the term, and its weights shrink the sensitivity to specificity at low prior. Detection and attribution are therefore separated.

Detection acts on the canonical coordinates. For a binary stream let $\widehat{\mathbold{R}}(t)$ be the class-time fading estimates of Remark~\ref{rem:accumulated}, with $n_{\mathrm{ess}}^{+}$ and $n_{\mathrm{ess}}^{-}$ the effective numbers of positive and negative arrivals they average, and binomial variances $\widehat{R}(1-\widehat{R})/n_{\mathrm{ess}}$. With a reference $\mathbold{R}_{\mathrm{ref}}$ and its variances, taken as the average of the estimates over a warm-up window, the statistic
\begin{equation}
\label{eq:chi2_stat}
\chi^2(t) = \sum_{c \in \{11, 00\}} \frac{\bigl(\widehat{R}_{c}(t) - R_{\mathrm{ref},c}\bigr)^2}{\widehat{\mathrm{Var}}\bigl(\widehat{R}_{c}(t)\bigr) + \widehat{\mathrm{Var}}\bigl(R_{\mathrm{ref},c}\bigr)}
\end{equation}
is compared with a threshold $\tau$. For $m>2$ classes the same construction is applied to each confusion row $\widehat{c}_k(t)$ (with the multinomial covariance) and the maximum over rows is monitored. By Propositions~\ref{prop:prior_invariance} and~\ref{prop:multiclass_invariance} a label shift does not change these statistics in expectation. A per-step chi-square level controls the false alarm per step, not per run: the statistic is evaluated at every step on an autocorrelated estimator, so the run-level false alarm probability is much higher than nominal. We therefore calibrate $\tau$ by simulation on a stationary stream with the same $\alpha$, length and priors so that the run-level false alarm probability is $5\%$. The binomial variances also assume independent arrivals. On the streams built by resampling they hold, and in native order the dependence of the data multiplies the variance of $\widehat{R}$ by the factor $\kappa(\alpha)$ of Remark~\ref{rem:autocorr}, which we measured and did not use in the detector. The positives seen after a change limit the power on the recall coordinate at low prior. Section~\ref{sec:exp3} measures this, together with four alternatives: the Jacobian $z$-alarm, $\bigl|\mathbold{J}_{\mathrm{concept}}^T(\widehat{\mathbold{R}} - \mathbold{R}_{\mathrm{ref}})\bigr| / \sqrt{\mathbold{J}_{\mathrm{concept}}^T(\mathbold{\Sigma} + \mathbold{\Sigma}_{\mathrm{ref}})\mathbold{J}_{\mathrm{concept}}}$ on time-fading coordinates, with delta-method variances $\widehat{R}(1-\widehat{R})/(2n)$ and a reference frozen at the start of scoring; the chi-square of Eq.~\eqref{eq:chi2_stat} on time-fading coordinates; DDM on the error indicator; and DDM stratified by class, one DDM on the errors of the positive arrivals and another on those of the negative arrivals, alarming when either does. The excess false alarm reported below is the paired difference between the probability of at least one alarm in a run under a label shift and under a stationary stream. Stratified ADWIN and Page-Hinkley raised almost no alarm, and their results are in the data files.

Attribution acts on a metric that a conventional monitor already tracks. When a monitor raises an alarm on $M$ (for instance $F_1$), the decomposition between a reference state and the current state splits $\Delta M$ into a concept term and a prior term, and the alarm is classified as concept-driven if the concept term accounts for at least a fraction $\theta$ of $|\Delta M|$ (we use $\theta = 1/2$) and as prior-driven otherwise. In the experiments the monitor raises an alarm on $F_1$ when it leaves a band of three standard errors around a reference window, with the delta-method variance, and the decomposition is the path-integrated one between the reference window and the current state, over 50-step block means. This triage can suppress alarms caused by a prior shift. It cannot detect a concept change that the monitored metric does not register.

%% ---------------------------------------------------------------------------
\section{Experimental Evaluation}
\label{sec:experiments}

Each subsection below tests specific claims of the paper, which are of three kinds. Exact identities (the linear decomposability of Theorem~\ref{thm:temporal_decomposability}, the zero marginal defect of the product $t$-norm) can fail only if the proof or the code is wrong, so the experiment checks the implementation to floating-point precision. Quantitative predictions (prior invariance, detection under label shift, attribution accuracy, the bias of the interim estimator, the undefined-window rate, the complexity bounds) can fail, and we report where they do. For comparative claims (Pareto dominance, the zero-division policies) we report what holds and what does not. The synthetic experiments use fixed seeds, 30 to 200 per setting, and report means with standard deviations or intervals across seeds, except Tables~\ref{tab:exp1} and~\ref{tab:exp2} and the single-run Table~\ref{tab:exp6} (single runs) and Table~\ref{tab:exp7} (seven repetitions). The experiments on real streams are single runs of one stream, except the delay and censoring study of R3 and R4 in \S\ref{sec:exp2} (15 seeds of the audit and delay draws, Table~\ref{tab:real_delay}) and the comparison of learners on R4 in \S\ref{sec:exp4} (10 seeds of the adaptive tree, Table~\ref{tab:nslkdd}). The streams and the protocol are described once in \S\ref{sec:setup}, and Table~\ref{tab:claimmap} lists the claim tested on each stream.

\subsection{Datasets, Streams and Protocol}
\label{sec:setup}

Table~\ref{tab:synth} lists the synthetic streams S1 to S10, each with a known generating process, so that the oracle value of every quantity is available. Classifier outputs on these streams are simulated from specified error rates (S2, S4, S7), from a simulated score (S5) or from a simulated posterior (S6); no model is trained on them.

\begin{table}[!ht]
\centering
\footnotesize
\begin{tabular}{@{}l p{7.5cm} r l@{}}
\toprule
\textbf{ID} & \textbf{Stream and configuration} & \textbf{$n$} & \textbf{Used in} \\
\midrule
S1 & 3-class stream with continuous prior drift, $P(Y{=}j)=g(\sin(t/200))$; fading $\alpha=0.98$ and window $W=100$. & 1,500 & \S\ref{sec:exp1} \\
S2 & Binary stream with classifier errors drawn independently from a specified $(\mathrm{TPR},\mathrm{TNR})$, base prior $\pi_0 \in \{0.5, 0.2, 0.05\}$ and one change at $n/2$: a prior change $\pi_0 \to \pi_0/2$ or $\pi_0/5$ with the classifier fixed (label shift), or a concept change (an abrupt, gradual, incremental or reoccurring drop of specificity by 0.1 to 0.3 or of recall by 0.1 to 0.2), or both. Fading $\alpha \in \{0.95, 0.98, 0.99\}$. & 1,000 to 3,000 & \S\ref{sec:exp3} \\
S3 & SEA streams with the four drift profiles: abrupt ($\theta: 8.0\to4.0$ at $t=750$), gradual (mixing the two concepts), incremental ($\theta$ moving linearly from $8.0$ to $4.0$ over $t\in[500,1000]$) and reoccurring ($\theta$ alternating between $8.0$ and $4.0$ every $300$ steps). & 1,500 & \S\ref{sec:exp3} \\
S4 & Multiclass streams with 3 and 5 classes, confusion rows with diagonal $0.85$ and the errors spread uniformly, mildly concentrated (two thirds of the errors to the next class) or cyclic (all errors to the next class); base prior $(0.5,0.3,0.2)$ or $(0.4,0.25,0.15,0.1,0.1)$; label shifts and one-row concept changes at $n/2$. & 3,000 & \S\ref{sec:exp3} \\
S5 & Binary, $20\%$ positive class, a non-perfect scorer (score $\sim \mathrm{Beta}(4,2)$ for positives and $\mathrm{Beta}(2,4)$ for negatives, decision at $0.5$, the score as stand-in posterior), fading $\alpha=0.98$. Three latency regimes: constant $\Delta=30$; Poisson $\lambda=25$; class-conditional $\Delta=60$ for the positive (fraud) class and $\Delta=5$ for the negative class. & 1,200 & \S\ref{sec:exp2} \\
S6 & Credit/risk stream: $x\sim\mathcal{N}(0,1)$, $P(Y{=}1\mid x)=\sigma(1.5x-1.2)$. The stand-in $\tilde{\mathbold{y}}_k$ is the true probability plus Gaussian noise (standard deviation $0.1$, clipped to $[0.01,0.99]$), approximately calibrated; an instance is flagged when it exceeds $0.45$. Flagged instances are censored unless audited with a $10\%$ budget (Eq.~\eqref{eq:audit_composite}). Fading $\alpha=0.99$. Variants add a Poisson latency $\lambda=25$, a class-dependent latency (Poisson $60$ for positives, $5$ for negatives), a label-dependent audit ($0.30$ for flagged positives, $0.05$ for flagged negatives), a miscalibrated or online-calibrated stand-in, inverse-propensity weighting as a baseline, the sliding-window form with $W=100$, and the eviction step with tolerances $10^{-4}$ and $5\times10^{-3}$. & 2,000 & \S\ref{sec:exp2} \\
S7 & Three learners simulated from specified rates, with a $30\%$ positive class and fading $\alpha=0.98$: A $\mathrm{TPR}=0.90$, $\mathrm{TNR}=0.70$; B $\mathrm{TPR}=0.65$, $\mathrm{TNR}=0.95$; C $\mathrm{TPR}=0.92$, $\mathrm{TNR}=0.95$. & 1,200 & \S\ref{sec:exp4} \\
S8 & Two rare event classes whose arrivals follow independent on/off Markov burst processes (mean off-duration $\approx 167$ steps, mean burst duration $\approx 20$ to $40$ steps), with a per-class sliding window $W=60$. & 4,000 & \S\ref{sec:exp4} \\
S9 & Throughput benchmark: consecutive instances on one CPU core (Apple M3, Python 3.13.9, pure NumPy, no vectorisation across instances). & 300,000 & \S\ref{sec:exp7} \\
S10 & Three classes with priors $(0.6, 0.35, 0.05)$, a classifier that is correct with probability $0.8$, window $W=60$, labels audited with probability $b$ and delayed by a constant or Poisson delay; the stand-in puts $\tfrac34$ on the predicted class and $\tfrac18$ on the others. & 20,000 & \S\ref{sec:exp4} \\
\bottomrule
\end{tabular}
\caption{Synthetic streams. $n$ is the number of stream instances.}
\label{tab:synth}
\end{table}

Table~\ref{tab:real} lists the five real streams. Real labels are used throughout; the label-availability mechanisms are real only where stated and simulated otherwise.

\begin{table}[!ht]
\centering
\footnotesize
\begin{tabular}{@{}l p{7.5cm} r l@{}}
\toprule
\textbf{ID} & \textbf{Stream and configuration} & \textbf{$n$} & \textbf{Used in} \\
\midrule
R1 & Electricity (Elec2)~\citep{harries1999splice}: half-hourly electricity-market records with seasonal drift and a documented abrupt regime change from a 1996 market-rule expansion. A gradient-boosting classifier trained once on the first $10{,}000$ instances is used for the label-shift studies. & 45,312 & \S\ref{sec:exp3} \\
R2 & Airlines~\citep{vanschoren2014openml} (OpenML \texttt{data\_id=1169}, from U.S.\ DOT on-time performance records). For the latency study, a flight's scheduled duration (\texttt{Length}) sets $\Delta_k$, since its on-time or delayed outcome is only known once it has landed. A gradient-boosting classifier trained once on the first $50{,}000$ instances is used for the label-shift studies. & 539,383 & \S\ref{sec:exp3}, \S\ref{sec:exp2} \\
R3 & Credit Card Fraud~\citep{dalpozzolo2015calibrating}: $0.173\%$ fraud rate, chronologically ordered by \texttt{Time}; fraud arrivals cluster in time ($37.5\%$ of consecutive fraud pairs land within $60$ seconds). The censoring policy is simulated: approved instances ($\hat{c}_k=0$, threshold $0.5$) are always observed, flagged ones only with a $10\%$ audit budget (Eq.~\eqref{eq:audit_composite}). A plain Hoeffding tree collapses to the majority class at this base rate, so it is wrapped in random oversampling~\citep{wang2018systematic,aguiar2024survey}. Cumulative aggregation (no fading) for the censoring runs; the oversampler targets a $70/30$ class distribution. & 284,807 & \S\ref{sec:exp2}, \S\ref{sec:exp4} \\
R4 & NSL-KDD~\citep{tavallaee2009detailed}: the standard KDDTrain+ and KDDTest+ partitions concatenated into one stream, not a chronological order. Five coarse attack categories (normal $51.88\%$, DoS $35.95\%$, Probe $9.48\%$, R2L $2.52\%$, U2R $0.17\%$). Fading $\alpha=0.995$ in the Pareto runs. For the label-shift studies a gradient-boosting classifier, binary (normal against attack) or with the five classes, is trained once on a random $30\%$. The binary version with an online Hoeffding tree is also used for the delay and censoring study of \S\ref{sec:exp2}. & 148,517 & \S\ref{sec:exp2}, \S\ref{sec:exp3}, \S\ref{sec:exp4} \\
R5 & Covertype (\texttt{sklearn.datasets.fetch\_covtype}): $54$ features, positive class Spruce/Fir (cover type 1, rate $0.36$); the file order is geographic, not chronological. Used only for the label-shift studies, with a gradient-boosting classifier trained once on a random $30\%$. & 581,012 & \S\ref{sec:exp3} \\
\bottomrule
\end{tabular}
\caption{Real streams. $n$ is the number of stream instances.}
\label{tab:real}
\end{table}

The real-stream experiments of \S\ref{sec:exp2} and \S\ref{sec:exp4} use an online Hoeffding tree from \texttt{river}~\citep{montiel2021river} run prequentially (test-then-train, one instance at a time) over the dataset's native row order. The model, and where used the calibration map $g_t$ of \S\ref{sec:product_tnorm}, are trained only on labels that have arrived. The label-shift studies of \S\ref{sec:exp3} use a different protocol, because a label shift requires control of the class priors. A scikit-learn gradient-boosting classifier is trained once on an initial chunk of R1 or R2, or on a random $30\%$ of R4 or R5, and frozen, so that $P(\hat{Y} \mid X)$ is fixed. Streams are built by class-conditional resampling of the held-out pool: at each step the class is drawn from a prior path, then an instance of that class is drawn uniformly with replacement (a clean label shift), or taken in the original order within the class (native order, which keeps the real drift of $P(X \mid Y)$). Concept drift is injected by permuting, across the stream instances after the change point, the values of the most important features of the frozen classifier, which lowers its recall and specificity by an amount measured on the pool.

\begin{table}[!ht]
\centering
\footnotesize
\begin{tabular}{@{}p{5.6cm} l l l@{}}
\toprule
\textbf{Claim} & \textbf{Kind} & \textbf{Streams} & \textbf{Where} \\
\midrule
Linear decomposability of the temporal state (Theorem~\ref{thm:temporal_decomposability}) & exact identity & S1 & \S\ref{sec:exp1} \\
Prior invariance of the canonical coordinates (Propositions~\ref{prop:prior_invariance}, \ref{prop:multiclass_invariance}) & quantitative & S2, S4, R1, R2, R4 & \S\ref{sec:exp3} \\
Drift profiles & quantitative & S3 & \S\ref{sec:exp3} \\
Attribution by the Jacobian decomposition & quantitative & S2, R1, R2, R4 & \S\ref{sec:exp3} \\
Detection on the canonical coordinates under label shift & quantitative & S2, S4, R1, R2, R4, R5 & \S\ref{sec:exp3} \\
Triage of alarms & quantitative & S2 & \S\ref{sec:exp3} \\
Zero marginal defect of the product $t$-norm; bounded queue & exact identity & S5 & \S\ref{sec:exp2} \\
Bias follows calibration error (Eq.~\eqref{eq:bias}); correction under censoring and class-dependent delay & quantitative & S6, R2, R3, R4 & \S\ref{sec:exp2} \\
The four label-availability settings; the window form and the eviction step & quantitative & S6 & \S\ref{sec:exp2} \\
Pareto dominance over time & comparative & S7, R4 & \S\ref{sec:exp4} \\
Zero-division policies & comparative & S8, R3, R4 & \S\ref{sec:exp4} \\
Undefined-window probability (Proposition~\ref{prop:undefined_window}, Corollary~\ref{cor:undefined_latency}) & quantitative & R3, S8, S10 & \S\ref{sec:exp4} \\
Resource complexity (Propositions~\ref{prop:complexity}, \ref{prop:complexity_window}) & quantitative & S9 & \S\ref{sec:exp7} \\
\bottomrule
\end{tabular}
\caption{Which claim is tested on which stream.}
\label{tab:claimmap}
\end{table}

\subsection{Decomposability of the Temporal State}
\label{sec:exp1}

Theorem~\ref{thm:temporal_decomposability} says that, at every time step and for both operators, the measure of the pooled one-vs-rest cells equals the denominator-weighted average of the per-class measures for the linear-fractional measures, and its proof rests on the commutation of $\Upsigma_{\mathrm{t}}$ with the selection matrices $\mathbold{S}_j$. On S1 we keep a global multiclass state and, separately, one binary evaluator per class fed only $(y=j, p=j)$ under the same operator. We compare $\mathbold{S}_j\mathbold{C}(t)\mathbold{S}_j^T$ with the binary evaluator's table, the pooled measure with the weighted average for Accuracy, Recall, Precision, $F_1$ and Jaccard, and, as a contrast, the pooled MCC and balanced accuracy with the equal-weight average, since these are not linear-fractional. A violation above floating-point level would count against the theorem or the implementation. Table~\ref{tab:exp1} shows that commutation and decomposition hold to rounding error under both operators, and that MCC and balanced accuracy do not decompose. This is a test of an algebraic identity, so it checks the implementation and not the claim's empirical content.

\begin{table}[!ht]
\centering
\footnotesize
\begin{tabular}{@{}lcccc@{}}
\toprule
Operator & Commutation & Decomposition & MCC contrast & Bal.\ acc.\ contrast \\
\midrule
Fading ($\alpha = 0.98$) & $5.0 \times 10^{-14}$ & $\le 4.4 \times 10^{-16}$ & $0.102$ & $0.044$ \\
Window ($W = 100$) & $0$ & $\le 2.2 \times 10^{-16}$ & $0.090$ & $0.039$ \\
\bottomrule
\end{tabular}
\caption{Largest difference over $1{,}500$ steps of S1. Decomposition is the largest of the five linear-fractional measures. Commutation is $\max|\mathbold{S}_j\mathbold{C}(t)\mathbold{S}_j^T - \text{binary evaluator}|$. The last two columns are the largest difference between the pooled value and the equal-weight average of the per-class values.}
\label{tab:exp1}
\end{table}

\subsection{Prior Invariance, Detection and Attribution of Drift}
\label{sec:exp3}

We test five claims. At a fixed time the canonical coordinates do not depend on the prior, while the pooled metrics do (Proposition~\ref{prop:prior_invariance}). In the multiclass case the one-vs-rest specificity is not invariant and the confusion rows are (Proposition~\ref{prop:multiclass_invariance}). A detector on the class-time coordinates raises no more false alarms under label shift than under stationarity, once its threshold is calibrated for the run-level false alarm. The Jacobian decomposition attributes a metric change to concept and prior within a first-order error. An alarm on $F_1$ caused by a prior shift can be told from one caused by a concept change. A false-alarm excess under label shift, an attribution residual above $10\%$ outside the large-shift regime, or a prior-driven alarm classified as concept-driven would count against them.

On S2, Table~\ref{tab:phase} gives the phase means over 200 seeds. A pure prior shift moves $F_1$ by $-0.49$ while Accuracy, Recall and Specificity stay within $0.01$ (the classifier has $\mathrm{TPR}=\mathrm{TNR}$, so Accuracy is prior-free here). A concept drift of specificity moves $F_1$ by $-0.60$ and Accuracy by $-0.24$. A monitor of $F_1$ therefore cannot tell the two apart, which is the motivation for the canonical coordinates.

\begin{table}[!ht]
\centering
\footnotesize
\fitbox{\input{tables/tab_phase.tex}}
\caption{Phase means, mean (sd) over 200 seeds, for a pure prior shift (A: $P(Y{=}1)$ from $0.50$ to $0.05$, $\mathrm{TPR}=\mathrm{TNR}=0.85$, S2) and an abrupt SEA concept drift (B, S3). The change is the difference of the phase means. The SEA boundary change of B also lowers the positive rate from $0.32$ to $0.08$, so part of the drop of $F_1$ in B is a prior effect; Recall and Specificity are not affected by it.}
\label{tab:phase}
\end{table}

On R1 (Elec2) with a Hoeffding tree frozen after the first $20{,}000$ instances and evaluated on the remaining $25{,}312$, Table~\ref{tab:elec2} gives the correlations with the prior $\pi(t)$, with moving-block bootstrap intervals (blocks of $100$ steps, about two memory lengths, $2{,}000$ resamples) and a null band from circular shifts of $\pi(t)$, which shows how large the correlation of two unrelated series with the same autocorrelation can be (up to $0.13$ to $0.19$). $F_1$ ($0.548$) and Accuracy ($0.446$) lie far outside the band ($p=0.002$). Recall and Specificity, whether computed on the accumulated cells ($0.111$ and $0.124$) or with class-time fading ($0.138$ and $0.123$), lie inside it ($p$ between $0.16$ and $0.30$), so the data show no dependence of the canonical coordinates on the prior, and they do not show that class-time fading removes any: the paired difference of the absolute correlation between the two estimators is $+0.027$ for recall (interval $[-0.003, +0.061]$) and $-0.001$ for specificity ($[-0.017, +0.012]$). With a classifier that keeps learning the correlations of the accumulated coordinates with the prior of the fading state are $0.316$ for $F_1$, $-0.077$ for Accuracy, $0.174$ for Recall and $-0.090$ for Specificity, and class-time fading correlates less for both coordinates ($0.122$ against $0.174$ for recall, $0.052$ against $0.090$ for specificity). Elec2 has real drift of $P(X \mid Y)$, so a correlation with $\pi(t)$ in this stream cannot by itself refute Proposition~\ref{prop:prior_invariance}, and with blocks of $1{,}000$ steps the intervals are wider than those of the table.

\begin{table}[!ht]
\centering
\footnotesize
\fitbox{\input{tables/tab_elec2.tex}}
\caption{Elec2 (R1), tree frozen after $20{,}000$ instances, fading $\alpha=0.98$: correlation of each series with the prior $\pi(t)$ of the fading state, a moving-block bootstrap interval, and the $95\%$ band of $|\mathrm{corr}|$ under circular shifts of $\pi(t)$ with its $p$-value.}
\label{tab:elec2}
\end{table}

\begin{table}[!ht]
\centering
\footnotesize
\fitbox{\input{tables/tab_taxonomy.tex}}
\caption{Four drift profiles on SEA streams (S3), mean (sd) over 200 seeds. The classifier is static.}
\label{tab:taxonomy}
\end{table}

Table~\ref{tab:taxonomy} gives the four drift profiles of S3. The abrupt, gradual and incremental profiles end in the same concept and lose $0.25$ to $0.26$ of specificity, with differences within one standard deviation across seeds. A single run that showed the incremental profile with the largest loss does not reproduce: it has the largest loss in none of 200 seeds. The reoccurring profile loses less because the post-drift window falls partly on a returned concept. The classifier is static, so no tracking lag is involved.

Table~\ref{tab:attribution} reports, for each metric and Jacobian estimator, the share of cells (drift shape, classifier and $\alpha$) of S2 in which the first-order residual is below $10\%$ of the observed change of the metric. The population residual of the midpoint rule for $F_1$ under a prior change from $0.5$ is $0.9\%$, $4\%$, $12\%$, $27\%$ and $38\%$ for final priors of $0.40$, $0.30$, $0.20$, $0.10$ and $0.05$ ($\alpha=0.98$), so the midpoint rule is accurate for modest changes and not for large ones, whatever the noise. Path integration over 50-step block means brings the sampled residual to $3\%$ to $5\%$ for final priors of $0.20$ or lower, for the $0.85/0.85$ classifier. For a concept change the population residual of the midpoint rule is below $1\%$ for drops of specificity up to $0.4$ and of recall up to $0.3$ from a base prior of $0.5$, and up to $2\%$ for $F_1$ and $5\%$ for Precision from a base prior of $0.2$; the sampled residual is dominated by estimation noise when the change is small, and the estimated concept term is close to the counterfactual (for a specificity drop of $0.2$, $-0.078$ against $-0.077$). When both change, the counterfactual depends on the order of the changes (for the $0.85/0.85$ classifier, an interaction of $-0.03$ to $-0.11$), and the estimate lies between the two orderings. Accuracy is bilinear in $(\mathbold{R},\pi)$ and the midpoint rule is exact for it.

\begin{table}[!ht]
\centering
\footnotesize
\fitbox{\input{tables/tab_attribution.tex}}
\caption{Share of S2 cells in which the first-order residual of the decomposition is below $10\%$ of the observed change of the metric (96 cells for concept only, 48 and 24 for prior only from base prior $0.5$ and $0.2$, 96 for both; Accuracy has valid prior-only residuals only for the asymmetric classifier, and its $83\%$ from base prior $0.2$ is 6 cells whose residual is sampling noise).}
\label{tab:attribution}
\end{table}

On the frozen-classifier streams the attribution reproduces the population values. For a fivefold drop of the prior, the prior term is $-0.149$ against $-0.156$ on R1 and $-0.202$ against $-0.216$ on R2, and the concept term is at most $0.006$ in absolute value. For corruption only, the concept term is $-0.284$, $-0.159$ and $-0.397$ against $-0.286$, $-0.161$ and $-0.400$ on R1, R2 and R4. When the prior drops and the classifier is corrupted, the sum matches the observed change and the split differs from the one-order counterfactual by the interaction term.

For detection on the binary streams of S2, Table~\ref{tab:detect_synth} compares the detector on the class-time coordinates (Eq.~\eqref{eq:chi2_stat}) with the Jacobian $z$-alarm, the same chi-square on time-fading coordinates and the standard river detectors. The nominal per-step threshold does not control the run-level false alarm: on the null stream, the default threshold gives a run-level false alarm of $0.08$ to $0.68$ depending on $\alpha$, because the post-reference period holds several hundred to two thousand autocorrelated checks. We therefore calibrate each threshold on the null so that the run-level false alarm is $5\%$, using 200 calibration seeds and evaluating on 200 disjoint seeds. The calibrated threshold grows with the stream length and decreases with $\alpha$ (for the class-time chi-square at $\alpha=0.98$ and $\pi_0=0.5$ with $(\mathrm{TPR},\mathrm{TNR})=(0.85,0.85)$ it is $12.0$ for $n=1000$ and $20.8$ for $n=3000$, against the nominal $11.83$). Over 144 cells of prior, classifier, length and $\alpha$, the excess of the run-level false alarm under a pure label shift over the null is $-0.002$ and $+0.001$ for the two class-time detectors, with the largest single cell at $0.045$, within the resolution of 200 paired seeds ($\pm 0.04$). It is $+0.055$ for the Jacobian alarm and $+0.064$ for the chi-square on time-fading coordinates, up to $+0.50$ at low prior, long streams and low $\alpha$. DDM on the error indicator has no excess with a symmetric classifier and an excess of up to $+0.70$ with $(\mathrm{TPR},\mathrm{TNR})=(0.90,0.70)$, because its error rate moves with the prior. The stratified DDM has no excess but a null false alarm of $0.21$ because it is not calibrated.

\begin{table}[!ht]
\centering
\footnotesize
\setlength{\tabcolsep}{3.5pt}\fitbox{\input{tables/tab_detect_synth.tex}}
\caption{Detection on S2 over all cells of the sweep. Null FA is the run-level false alarm under the null. Excess FA is the paired difference between the run-level false alarm under a pure label shift and under the null, averaged over cells; in parentheses the number of cells whose interval lies above and below zero (144 cells for the calibrated detectors, 48 for the two DDM variants). Power is for an abrupt drop of specificity or recall of $0.2$. Calibrated thresholds give a $5\%$ null false alarm; DDM variants use the river defaults.}
\label{tab:detect_synth}
\end{table}

At the calibrated threshold, the class-time chi-square detects an abrupt specificity drop of $0.2$ with probability $0.86$ and a recall drop of $0.2$ with $0.53$, against $0.44$ and $0.53$ for the Jacobian alarm. A specificity drop is detected from the negatives, so its power hardly depends on the prior ($0.87$, $0.87$ and $0.85$ for $\pi_0=0.05$, $0.2$ and $0.5$, for an abrupt drop of $0.2$). A recall drop needs positives: for an abrupt recall drop of $0.2$ at $n=1000$ the power rises from $0.19$ to $0.85$ with $\pi_0$, for about $25$ to $250$ positives after the change. Gradual and incremental drifts cost a few points of power, reoccurring drift more ($0.70$ against $0.86$). For an abrupt specificity drop of $0.2$ the median delay is $119$ steps at $n=1000$ and $144$ at $n=3000$ ($155$ and $334$ pooled over the four drift shapes).

\begin{table}[!ht]
\centering
\footnotesize
\setlength{\tabcolsep}{3.5pt}\fitbox{\input{tables/tab_detect_real.tex}}
\caption{Label shift on frozen-classifier streams built from R1, R2, R4 and R5, thresholds calibrated on the shuffled null. Excess FA is the paired excess of the run-level false alarm under a prior-only change, averaged over datasets, $\alpha \in \{0.98, 0.99\}$ and the two prior drops; in parentheses the number of cells (12 for R1, R2 and R4 together, 4 for R5) whose interval lies above zero. Power is the probability of an alarm after a corruption-only change and none before it, $\alpha=0.98$.}
\label{tab:detect_real}
\end{table}

Table~\ref{tab:detect_real} repeats the test on real instances and a real classifier, whose errors depend on $x$ and are autocorrelated, on four streams (R1, R2, R4 and the Covertype stream R5). Calibrated on the shuffled null, the excess is $+0.054$ for the Jacobian alarm and $+0.072$ for the time-fading chi-square, against $-0.010$ and $-0.005$ for the class-time detectors, none of whose 12 cells has an interval above zero. The first two rise on R1 and R2, where the frozen classifier is asymmetric, and stay flat on R4, where it is nearly symmetric. On R5, a fourth stream with an asymmetric classifier, the excess is $+0.024$ for the Jacobian alarm and $+0.115$ for the time-fading chi-square (4 of 4 cells above zero), against $-0.013$ and $-0.019$ for the class-time detectors, and the power after the corruption is $0.99$ for the class-time chi-square. Power after a corruption-only change is high for the class-time chi-square on all three streams ($0.90$ to $0.98$), with a median delay of about $120$ steps on R1 against about $230$ for the Jacobian alarm. In native order the real drift of $P(X \mid Y)$ dominates: thresholds calibrated on the shuffled null give a null false alarm of $0.68$ to $0.99$ on R1 and R2, so alarms there cannot be read as label shift or concept drift, and thresholds calibrated on native-order null streams remove the false alarms at the price of almost no power against the corruption. To test whether the dependence of Remark~\ref{rem:autocorr} explains these false alarms, we divided each $z$ statistic of the class-time detectors by $\sqrt{\hat\kappa}$ and kept the threshold calibrated on the shuffled null. With $\hat\kappa$ computed on the whole pool sequence after removing a moving average of $1{,}001$ arrivals, the native-order null false alarm of the class-time chi-square falls from $0.99$ to $0.22$--$0.25$ on Elec2 and from $0.73$--$0.78$ to $0.45$--$0.51$ on Airlines, and the power after the corruption falls from $0.91$--$0.97$ to $0.09$--$0.29$ and from $0.91$--$0.95$ to $0.60$--$0.81$. With $\hat\kappa$ computed on the $600$-step warm-up window of the stream, which misses the long lags, the false alarm stays at $0.56$--$0.74$. Without the moving average, the false alarm and the power both fall to nearly zero. On the resampled streams the warm-up estimate is close to one ($1.06$ to $1.10$) and nothing changes, and on NSL-KDD $\hat\kappa \approx 1$. Dependence therefore explains part of the false alarms in native order. The rest is a variation of the mean that no variance inflation absorbs without removing the power, which is the real drift of $P(X \mid Y)$ that this experiment cannot separate from the effect under test. No class-time detector shows a significant excess in native order.

\begin{table}[!ht]
\centering
\footnotesize
\fitbox{\input{tables/tab_multiclass.tex}}
\caption{Excess of the run-level false alarm under label shift (mean over the label-shift scenarios) for one-vs-rest detectors on the pair $(\mathrm{Recall}_j, \mathrm{Specificity}_j)$ and row-wise detectors on the confusion rows, thresholds calibrated for a $5\%$ family-wise false alarm. S4 for $m=3$ and $5$ classes; the last row is R4 with five classes.}
\label{tab:multiclass}
\end{table}

For $m>2$ classes the one-vs-rest specificity moves with the priors of the other classes, as Proposition~\ref{prop:multiclass_invariance} says. Table~\ref{tab:multiclass} shows the consequence on S4. With a uniform confusion, where the proposition predicts invariance, no detector has an excess. With a cyclic confusion the one-vs-rest detectors have an excess of $+0.40$ ($m=3$) and $+0.33$ ($m=5$) at $\alpha=0.98$ and up to $+0.74$ at $\alpha=0.99$, while the row-wise detectors stay within $\pm 0.03$. On R4 with five classes, a fivefold increase of the priors of U2R and R2L lowers the population specificity of the normal class from $0.955$ to $0.880$, and the one-vs-rest excess is $+0.115$ at $\alpha=0.98$ and $+0.605$ at $\alpha=0.99$ for that shift, against $0.000$ and $-0.015$ for the rows. The row-wise detectors also detect a concept change in the smallest row more often than the one-vs-rest ones ($0.48$ to $0.75$ against $0.22$ to $0.25$ for $m=5$ at $\alpha=0.98$), although the family-wise calibration costs power: an oracle that monitors only the changed row reaches $0.80$ to $0.93$. When a label shift starves the changed row of arrivals, every detector is blind, and the U2R row has about five arrivals per $3{,}000$ steps. The confusion structures of S4 are our choice, and the size of the effect depends on them.

\begin{table}[!ht]
\centering
\footnotesize
\fitbox{\input{tables/tab_triage.tex}}
\caption{Triage of $F_1$ alarms from a 3-sigma monitor on S2 with the path-integrated decomposition and $\theta=1/2$. Prior-driven alarms suppressed is the share of alarms in scenarios with only a prior change that are classified as prior-driven; concept alarms kept is the share of alarms caused by a concept change that are classified as concept-driven.}
\label{tab:triage}
\end{table}

For triage, Table~\ref{tab:triage} classifies the alarms of a 3-sigma monitor on $F_1$. At $\alpha=0.98$, $93\%$ of the alarms caused by a prior shift are classified as prior-driven and $86\%$ of those caused by a concept change as concept-driven, with a pooled accuracy of $0.90$; the pooled accuracy is $0.81$ at $\alpha=0.95$ and $0.95$ at $\alpha=0.99$, and $\theta=1/2$ balances the two kinds of error better than $1/4$ or $3/4$. Triage is not a detector. After a prior shift to $\pi=0.05$ the monitor on $F_1$ rarely fires on a later concept change (in $3\%$ to $19\%$ of runs), and an alarm caused by noise alone is classified as concept-driven, because noise lands in the concept term. The attribution has been tested on a binary stream with a fixed classifier, independent errors, fully labelled data and an abrupt prior change, and the other settings are not covered.

\subsection{Verification Latency, Selective Feedback, and Auditing}
\label{sec:exp2}

Five claims are tested. With the product $t$-norm the interim stand-in has zero marginal defect (Eq.~\eqref{eq:defect}) and the verification queue stays bounded. The residual bias of the unified estimator follows the calibration error of the stand-in (Eq.~\eqref{eq:bias}). Under decision-dependent censoring with an audit budget the naive estimator is biased and the corrected one removes most of the bias. The same holds in the four label-availability settings of Table~\ref{tab:settings}. A non-zero defect, a bias that does not follow the calibration error, or a corrected estimate no closer to the oracle than the naive one would count against them. Throughout, the naive estimate is computed on the verified instances only, the corrected estimate is the unified state of Eq.~\eqref{eq:unified_cm}, in which the stand-in replaces the label of every pending or unaudited instance, and the oracle is the state obtained when every label is available at arrival.

On S5, with a non-perfect scorer, the verified recall differs from the oracle recall under label-dependent latency (Table~\ref{tab:exp2}): with positives confirmed after $60$ steps and negatives after $5$, the verified recall is $0.859$ against an oracle of $0.711$, because the verified set over-represents the class confirmed first. The product $t$-norm has zero defect for soft and for hard predictions, and the minimum and \L{}ukasiewicz $t$-norms have a defect of about $1.17$ for soft predictions and coincide with the product for hard ones. The queue is bounded by construction when $\Delta$ is finite.

\begin{table}[!ht]
\centering
\footnotesize
\setlength{\tabcolsep}{3pt}
\begin{tabular}{@{}lccccc@{}}
\toprule
Latency regime & Queue, mean / max & Verified & Recall (oracle) & Defect $T_P$ & Defect $T_M$, $T_L$ \\
\midrule
Constant ($\Delta = 30$) & 28.7 / 29 & 1,171 / 1,200 & 0.796 (0.800) & 0.000 & 1.171 \\
Poisson ($\lambda = 25$) & 23.5 / 29 & 1,176 / 1,200 & 0.923 (0.942) & 0.000 & 1.169 \\
Class-dependent ($60$ / $5$) & 13.3 / 20 & 1,182 / 1,200 & 0.859 (0.711) & 0.000 & 1.179 \\
\bottomrule
\end{tabular}
\caption{Latency regimes on S5, one run. The defect is the mean marginal defect (Eq.~\eqref{eq:defect}) of the stand-in and the soft prediction; it is zero for all three $t$-norms with hard predictions. The queue holds up to $\Delta-1$ entries because an instance is recorded before the clock advances.}
\label{tab:exp2}
\end{table}

Airlines (R2) replaces this synthetic latency schedule with a real, covariate-dependent one (mean $\Delta_k = 132$ min, maximum 655 min, mean pending queue size 347). With an uncalibrated Hoeffding tree, the accuracy of the naive estimator is $0.0897$ below that of the corrected one (mean absolute difference over the stream $0.0986$). No oracle is evaluated, so this is the difference between the two estimators and not their bias. It is close to the calibration gap of the raw classifier, which is overconfident by $0.0906$ (mean confidence $0.7284$ against accuracy $0.6378$), as Eq.~\eqref{eq:bias} suggests. After online Platt-scaling recalibration the calibration gap is $-0.0100$ and the difference is $-0.0165$ (mean absolute $0.0448$). Two points do not establish proportionality; Table~\ref{tab:censor} and the paragraph that follows test it directly.

Table~\ref{tab:censor} reports 200 seeds of S6 and its variants. In the immediate partially labelled setting with audit by decision, the naive recall bias is $-0.434$, inverse-propensity weighting (IPW), which weights each verified instance by the inverse of its audit probability, is nearly unbiased with the true probability ($-0.030$) with a standard deviation five times that of the corrected estimator (RMSE $0.144$ against $0.028$), and the corrected estimator has bias $+0.013$. IPW is also fragile: with a label-dependent audit and a misspecified, decision-only probability it is biased by $+0.227$.

The bias of the corrected estimator is linear in the calibration error of the stand-in measured on the unverified instances: over eight stand-ins (logit shifts of $-1$ to $+1$, including the calibrated one, and temperatures of $0.5$ to $3$) the pooled slope over $1{,}600$ runs is $0.346$ (interval $[0.340, 0.353]$) with $R^2 = 0.99$ across the eight configuration means; with delayed labels the single-regressor fit is weaker ($R^2 = 0.54$ across configuration means) and the bias is linear only when the calibration errors of the flagged and the approved unverified instances enter separately ($R^2 = 0.98$), and it reproduces the residual of the calibrated stand-in, which is not exactly calibrated among flagged instances. Online Platt scaling on the logit of the raw score gives $-0.018$ in this setting, and a version that uses the raw probability as the feature is undertrained on a 2,000-step stream ($-0.202$).

A label-dependent audit with a six-to-one ratio leaves a corrected bias of $+0.031$, and the correction is partial under label-dependent delay: with the delay of S5, $+0.094$ in the delayed and partially labelled setting, which removes $79\%$ of the naive bias but does not recover the oracle, and $+0.055$ in the fully labelled delayed setting, where the naive recall is nearly unbiased ($-0.003$) and the correction makes it worse.

\begin{table}[!ht]
\centering
\footnotesize
\setlength{\tabcolsep}{3pt}\scriptsize\fitbox{\input{tables/tab_censor.tex}}
\caption{Recall bias against the oracle at the end of the stream, mean (sd) over 200 seeds of S6 and its variants. Audit by decision is the $10\%$ budget for flagged instances; audit by label is $0.30$ for flagged positives and $0.05$ for flagged negatives; class-dependent delay is Poisson $60$ for positives and $5$ for negatives. Naive uses verified labels only; IPW weights verified instances by the inverse of the true audit probability, or of a probability that depends on the decision only; corrected uses a calibrated stand-in, online Platt scaling on the logit of the score, or the raw score. IPW ignores delay, so it is shown only where the audit is the only imperfection.}
\label{tab:censor}
\end{table}

The Credit Card Fraud stream (R3) reproduces the censoring-plus-audit composite (Eq.~\eqref{eq:audit_composite}) on real transaction features and labels, with the censoring policy simulated on top (\S\ref{sec:setup}). At this base rate a plain Hoeffding tree predicted the majority class throughout (0 fraud predictions across 284,807 transactions), which is why R3 uses the oversampling wrapper. With the raw score as stand-in the corrected $F_1$ was $0.9469$ against an oracle of $0.0119$, because every censored instance is flagged by construction and its raw score is self-consistent with the decision. With online Platt-scaling recalibration trained only on observed labels, the recall bias goes from $-47.3$ percentage points (naive) to $-0.8$ points (corrected) and the specificity bias from $+22.5$ points to $-0.01$ points, on a stream in which $23.4\%$ of the labels are censored. The corrected $F_1$ is $0.0109$ against the oracle's $0.0119$, a relative error of $8\%$, and the naive $F_1$ is $0.0099$, an error of $17\%$. That run has one audit seed.

Table~\ref{tab:real_delay} repeats the test with 15 seeds, adds delay to the censoring, and adds a second real stream, NSL-KDD collapsed to a binary problem (normal against attack, R4, file order) with a plain Hoeffding tree. The audit budget is $10\%$, the delay is none, Poisson $300$ or class-dependent (Poisson $900$ for positives and $100$ for negatives), the operator is cumulative, and the model and the calibrator learn only from labels that have arrived. Without delay, the first seed reproduces the run above. The naive estimator is badly biased on both streams, in recall ($-0.42$ on R3, $-0.24$ to $-0.27$ on R4) whatever the delay. Inverse-propensity weighting with the exact audit probability, which is available only because we simulated the audit, is nearly unbiased in recall (within $0.005$) under the cumulative operator. The corrected estimator recovers recall on R4 (bias $-0.004$ for every delay) and on R3 without delay ($-0.010$), but it is only partial elsewhere. On R3 with delay its recall bias is $-0.14$ and $F_1$ is underestimated by about two thirds ($-0.010$ against an oracle of about $0.014$); on R4 it underestimates specificity ($-0.074$), precision ($-0.09$) and $F_1$ ($-0.05$) with or without delay. In a diagnostic run (R4, seed 0, no delay) the calibrated stand-in averaged $0.862$ on the censored flagged instances against a true rate of $0.960$; we did not diagnose the other cases, and the stand-ins disagree among themselves (Platt scaling on the logit gives $+0.04$ in recall on R3, and the raw score is far off, with a precision bias of $+0.89$). So the statement that the correction removes the censoring bias holds for recall on R4 and on R3 without delay, and not for R3 with delay or for precision, specificity and $F_1$ on R4. Under fading with $\alpha=0.999$ IPW is noisy on R3 (recall bias $-0.29$, sd $0.41$, with about $1.7$ effective frauds in memory). With the eviction tolerance $10^{-4}$, the queue holds on average $243$ (R3) and $170$ (R4) entries under Poisson $300$, and $80$ and $110$ under the class-dependent delay, when only audited instances are queued, and $528$ to $1{,}154$ entries when every instance is queued; eviction removed no audited entry in any run.

\begin{table}[!ht]
\centering
\scriptsize
\setlength{\tabcolsep}{3pt}
\fitbox{\input{tables/tab_real_delay.tex}}
\caption{Bias against the oracle at the end of the stream, mean (sd) over 15 seeds, cumulative operator, audit budget $0.10$, on R3 (Credit Card Fraud, oversampled Hoeffding tree) and R4 collapsed to a binary problem (plain Hoeffding tree, file order). The delay is simulated; class-dependent is Poisson $900$ for positives and $100$ for negatives. IPW uses the exact audit probability. The oracle precision is about $0.007$ and the oracle $F_1$ about $0.014$ on R3, and the oracle $F_1$ is about $0.96$ on R4.}
\label{tab:real_delay}
\end{table}

Table~\ref{tab:exp12} runs the four settings on S6. Without censoring, the naive estimator is already close to the oracle, because a latency of about $25$ steps is small against the memory $n_{\mathrm{eff}}=100$, so delay alone shifts the estimate little and the interim correction mainly reduces its variability (recall bias sd $0.037$ against $0.052$). Censoring produces the large bias ($-0.43$ in recall for the naive estimator), and the corrected estimator removes it with or without delay. Adding delay to censoring changes the bias of neither estimator and increases the spread of the corrected one (recall bias sd $0.048$ against $0.027$ without delay), presumably because the stand-in stays in place until the label arrives; we did not isolate this. Where the system decides which labels to request, the queue holds only the selected instances still pending ($19$ on average in setting (iv)), whereas holding every instance until eviction needs $115$ entries. These queue sizes are counted from the arrival schedule, and Table~\ref{tab:window} reports them again with the eviction step executed.

\begin{table}[!ht]
\centering
\footnotesize
\setlength{\tabcolsep}{1.8pt}
\begin{tabular}{@{}lccccc@{}}
\toprule
 & \multicolumn{2}{c}{\textbf{Recall bias}} & \multicolumn{2}{c}{\textbf{$F_1$ bias}} & \textbf{Queue size} \\
\cmidrule(lr){2-3}\cmidrule(lr){4-5}\cmidrule(lr){6-6}
\textbf{Setting} & Naive & Corrected & Naive & Corrected & decides / unknown \\
\midrule
(i) immediate, full & $0.000$ & $0.000$ & $0.000$ & $0.000$ & $0$ / $0$ \\
(ii) delayed, full & $+0.007\ (0.052)$ & $+0.008\ (0.037)$ & $+0.005\ (0.040)$ & $+0.009\ (0.032)$ & $24.8$ / $24.8$ \\
(iii) immediate, partial & $-0.431\ (0.058)$ & $+0.013\ (0.027)$ & $-0.394\ (0.074)$ & $+0.020\ (0.045)$ & $0$ / $96.2$ \\
(iv) delayed, partial & $-0.431\ (0.069)$ & $+0.014\ (0.048)$ & $-0.396\ (0.085)$ & $+0.021\ (0.053)$ & $19.0$ / $115.2$ \\
\bottomrule
\end{tabular}
\caption{Bias against the all-labels-immediately oracle at the end of the stream in the four label-availability settings of \citet{gomes2022survey} (full or partial labelling; immediate or delayed labels); bias mean (sd) over $30$ seeds. Last column: mean queue size when the system decides which labels to request (only selected, pending instances are enqueued) and when eligibility is unknown (every pending instance is held until its label arrives or its fading weight drops below $\varepsilon=10^{-4}$).}
\label{tab:exp12}
\end{table}

Table~\ref{tab:window} runs the sliding-window form of the interim estimator ($W=100$) and the eviction step of Algorithm~\ref{alg:update} (fading $\alpha=0.99$, tolerance $\varepsilon=10^{-4}$, every instance queued) on S6 over 100 seeds. The windowed corrected estimator removes the censoring bias as the fading one does (recall bias $+0.014$ against $-0.444$ for the naive windowed estimator in setting~(iii)). It also does when the delay is longer than the window (Poisson $150$, setting~(v)): every instance leaves the window unverified, the state consists of interim terms only, and its recall bias of $+0.008$ is the quality of the stand-in itself. The queue sizes of fading with eviction ($96.3$ and $115.3$) reproduce the counts of Table~\ref{tab:exp12}. With $\varepsilon=10^{-4}$ eviction (after $s^\ast=458$ steps) changes the state in no run, because no label arrives that late. With $\varepsilon=5\times10^{-3}$ ($s^\ast=69$ steps) and the delay of setting~(v), eviction does discard late corrections: it changes the state by at most $2.4$ cell units on a total mass of about $100$ (mean over seeds of the largest change $1.5$), the recall bias is $+0.011$ against $+0.010$ without eviction, and the queue shrinks from $348$ to $68$ entries.

\begin{table}[!ht]
\centering
\scriptsize
\setlength{\tabcolsep}{1.8pt}
\fitbox{\input{tables/tab_window.tex}}
\caption{Recall bias against the oracle (mean (sd) over 100 seeds) and queue size (mean and mean of the maximum) of the fading and sliding-window estimators on S6, with every instance queued. Fading uses $\alpha=0.99$ and eviction tolerance $10^{-4}$; the window is $W=100$.}
\label{tab:window}
\end{table}

\subsection{Multiclass Structural Diagnostics and Zero-Division Auditing}
\label{sec:exp4}

Three claims are examined. Pareto dominance over time separates learners that a scalar metric ranks arbitrarily; this is a demonstration of the framework and not a falsifiable prediction. We tested the conjecture that the $\mathtt{exclude}$ policy gives the smoothest macro trajectory, and it does not hold across regimes; what we claim instead is the audit property, that $\mathtt{exclude}$ and $\mathtt{nan}$ never report a number for a class without verified support. Except for S10 below, the experiments of this subsection have no latency, where the verified and the interim supports coincide. The undefined-window probability of Proposition~\ref{prop:undefined_window} and the direction of its bursty extension (Remark~\ref{rem:bursty_undefined}) are tested on a real transaction timeline. Pareto dominance is defined at a time step: learner $a$ dominates $b$ if $R_{11}^a(t) \ge R_{11}^b(t)$ and $R_{00}^a(t) \ge R_{00}^b(t)$ with at least one strict inequality, and the two are incomparable if neither dominates.

On S7, over 100 seeds, Models A and B are incomparable in $98.8\%$ of the windows (standard deviation $2.2$ points) as their generating rates are, so any ranking between them by $F_1$ or Accuracy is an artifact of metric scalarization. The generating rates of Model C dominate those of Model A on both coordinates, yet the fading estimates ($n_{\mathrm{eff}}=50$) show C dominating A in only $60.7\%$ of the windows (standard deviation $12.3$ points), because of sampling noise at this memory, which the Pareto relation, unlike a scalar ranking, makes visible. A single seed gave $54.4\%$. The relation is that of the ROC convex hull~\citep{provost2001robust}, extended to continuously updated temporal states.

NSL-KDD (R4) is the only multiclass real benchmark of this paper, and its row order is not chronological (\S\ref{sec:setup}). Applying the $\mathbold{S}_j$ projection (Eq.~\eqref{eq:ovr_projection}) to its five classes and comparing a plain with an adaptive Hoeffding tree gave, in one run (native order, adaptive seed $2027$), a pattern in which the adaptive tree dominates more often on the rare classes (R2L: $39.2\%$ against $16.3\%$ of windows; U2R: $41.9\%$ against $16.1\%$) and the plain tree on normal ($20.7\%$ against $9.8\%$). Table~\ref{tab:nslkdd} shows that this run does not represent the other seeds. Over 10 seeds of the adaptive tree the adaptive advantage on normal, DoS and Probe is stable (DoS in 10 of 10 seeds), the single run was at the favourable end for R2L and U2R, and for normal it had the opposite sign to the mean. For R2L and U2R the sign is not stable across seeds (6 and 5 of 10), and when the rows are permuted within blocks of $1{,}000$ it reverses (the adaptive tree ahead in 2 and 4 of 10 orders). The Pareto relation therefore separates the learners per class, but for the rare classes of this stream the sign of the difference depends on the seed and the order.

\begin{table}[!ht]
\centering
\footnotesize
\setlength{\tabcolsep}{3pt}
\fitbox{\input{tables/tab_nslkdd.tex}}
\caption{NSL-KDD (R4): fraction of windows in which the adaptive tree dominates the plain tree minus the fraction in which the plain tree dominates the adaptive one, mean over 10 runs with a bootstrap interval, and the number of runs in which the adaptive tree is ahead. Only the adaptive tree has a seed; in the second block the rows are permuted within blocks of $1{,}000$.}
\label{tab:nslkdd}
\end{table}

On S8, whose rare event classes arrive in bursts, Table~\ref{tab:exp6} shows the macro-precision trajectory under the policies of \S\ref{sec:zero_division}. In this single run $\mathtt{exclude}$ reduces the total variation of the trajectory by $34\%$ against the $\mathtt{zero}$ default ($17.66$ against $26.85$) while flagging $26.4\%$ of steps as partly undefined, and $\mathtt{one}$, which counts $0/0$ as $1.0$ ($11.85$), is not reliably worse than $\mathtt{exclude}$. Over 100 seeds the reduction is $21.6\%$ (standard deviation $18.0$ points), but the total variation is summed over different numbers of steps (the policies leave different shares of the stream undefined); per defined step it is $-4.7\%$ (standard deviation $26$ points, interval $[-9.8, +0.4]$), so the smoothness advantage disappears once the number of steps is accounted for.

\begin{table}[!ht]
\centering
\footnotesize
\begin{tabular}{@{}llccc@{}}
\toprule
\textbf{Convention} & \textbf{$0/0$ Action} & \textbf{Undef.} & \textbf{TV} & \textbf{Mean $\pm$ Std} \\
\midrule
$\mathtt{zero}$ (sklearn / MOA)      & $0/0 \to 0.0$               & $0.0\%$  & 26.85 & $0.387 \pm 0.308$ \\
$\mathtt{one}$ (alt.\ default)       & $0/0 \to 1.0$               & $0.0\%$  & 11.85 & $0.926 \pm 0.170$ \\
\textbf{$\mathtt{exclude}$}          & avg.\ over supported        & $26.4\%$ & \textbf{17.66} & $0.852 \pm 0.298$ \\
$\mathtt{nan}$ (strict)              & any $0/0 \to$ undefined     & $81.2\%$ & 4.05  & $0.787 \pm 0.233$ \\
\bottomrule
\end{tabular}
\caption{Macro-temporal precision ($\Sigma_m \circ \Upsigma_{\mathrm{t}}$) over $2$ bursty rare event classes under each zero-division policy of \S\ref{sec:zero_division}. TV is computed over each convention's own defined (non-undefined) subsequence; ``Undefined'' is the fraction of steps the convention leaves without a numeric value.}
\label{tab:exp6}
\end{table}

We repeated the protocol on two real streams, Credit Card Fraud (R3), whose fraud arrivals cluster in time, with a single binary metric and $W=200$, and NSL-KDD's four event classes (R4; DoS, Probe, R2L, U2R, excluding the normal background) with $W=300$. Table~\ref{tab:zerodiv_crosscut} gives the outcome: the smoothing effect is negligible on real binary data ($+2\%$) and reverses on the real multiclass stream ($-35\%$). In $79\%$ of the NSL-KDD windows three or four of the four event classes are supported, so this is not the regime of mostly empty support; the fluctuating membership of $\mathtt{exclude}$ shifts the average each time a class enters or leaves the support, a larger jump than the fixed composition of $\mathtt{zero}$ with occasional zero padding, since the rare classes' true precision is often close to zero. These total variations are raw sums and have the step-count issue noted above.

\begin{table}[!ht]
\centering
\small
\begin{tabular}{@{}lccrrr@{}}
\toprule
\textbf{Setting} & \textbf{Real?} & \textbf{Classes} & \textbf{TV (zero)} & \textbf{TV (excl.)} & \textbf{$\Delta$} \\
\midrule
Synthetic (Table~\ref{tab:exp6}) & No  & 2 & 26.85 & 17.66  & $+34\%$ \\
Credit Card Fraud                 & Yes & 1 (fraud) & 242.74 & 236.74 & $+2\%$ \\
NSL-KDD                           & Yes & 4 (event) & --- & --- & $-35\%$ \\
\bottomrule
\end{tabular}
\caption{The $\mathtt{exclude}$-vs-$\mathtt{zero}$ smoothness comparison across one synthetic and two real streams ($\Delta{>}0$: $\mathtt{exclude}$ smoother). Sign and magnitude are regime-dependent, not a fixed property of the policy; NSL-KDD's raw TV values are omitted as they are not directly comparable in scale to the other two rows' single-metric TV (four-class macro composition vs.\ one).}
\label{tab:zerodiv_crosscut}
\end{table}

The smoothness claim does not generalise, and we do not carry it forward. What holds in all three settings of Table~\ref{tab:zerodiv_crosscut} is the audit property: $\mathtt{exclude}$ and $\mathtt{nan}$ never report a number for a class without verified support, while $\mathtt{zero}$ and $\mathtt{one}$ always do, whether or not their constant happens to be close to the truth. With latency the property needs the verified support of \S\ref{sec:zero_division}, which only the experiment on S10 below exercises.

On S10 (three classes, window $W=60$, the rare class with prior $0.05$), Table~\ref{tab:support} compares the probability that no instance of the rare class in the window is verified with the formula of Corollary~\ref{cor:undefined_latency}: $0.161$ against $0.166$ for a constant delay of $25$, $0.409$ against $0.412$ with an audit probability of $0.5$, $0.163$ and $0.422$ against $0.167$ and $0.413$ with Poisson delays, and $1$ for a delay of $70 \ge W$. The probability that the class is absent from the window is $0.042$ throughout. The verified support never reports a value for a class with no true instance, whereas the interim support does so in $4.2\%$ of the steps, those at which the class is absent, because the stand-in always puts mass on it. The price is accuracy: the absolute error of the macro recall under $\mathtt{exclude}$ against the oracle is $0.012$ to $0.041$ with the verified support and $0.006$ with the interim support, because the verified support drops a class that is present but whose labels are pending, and with a delay beyond the window it leaves the macro recall undefined.

\begin{table}[!ht]
\centering
\footnotesize
\fitbox{\input{tables/tab_support.tex}}
\caption{Verified support under latency on S10 (window $W=60$, rare class prior $0.05$; 4 runs of 20{,}000 steps per row). Formula is the probability of Corollary~\ref{cor:undefined_latency} that no instance of the rare class is verified; Empirical is its frequency; Class absent is the frequency with which the class has no true instance in the window. MAE is the absolute error of the macro recall under $\mathtt{exclude}$ against the oracle when the support is the verified or the interim one.}
\label{tab:support}
\end{table}

On how often a policy trigger fires, Table~\ref{tab:undefined_realdata} compares the empirical undefined-window rate on the chronologically ordered Credit Card Fraud stream, at the class's marginal rate $p=0.001727$, with the i.i.d.\ prediction of Eq.~\eqref{eq:undefined_window_iid} for three window sizes. The real rate exceeds the i.i.d.\ baseline at every $W$, by a factor of $1.1$ at $W=200$, $1.4$ at $W=500$ and $2.2$ at $W=1{,}000$, which is the direction that Remark~\ref{rem:bursty_undefined} predicts for clustered arrivals. The Markov formula of that remark predicts $56.4\%$ for the generator of S8 and a simulation of $4\times10^5$ steps gives $57.6\%$ on average over three seeds and two classes (standard deviation $0.7$ points).

\begin{table}[!ht]
\centering
\small
\begin{tabular}{@{}cccc@{}}
\toprule
\textbf{Window ($W$)} & \textbf{i.i.d.\ prediction, Eq.~\eqref{eq:undefined_window_iid}} & \textbf{Empirical (real data)} & \textbf{Gap} \\
\midrule
200   & 70.8\% & 78.7\% & $+7.9$ pts \\
500   & 42.1\% & 59.1\% & $+17.0$ pts \\
1,000 & 17.7\% & 38.1\% & $+20.3$ pts \\
\bottomrule
\end{tabular}
\caption{Undefined-window rate for the fraud class on the real, chronologically-ordered Credit Card Fraud stream ($p=0.001727$, $n=284{,}807$), against the i.i.d.\ prediction of Eq.~\eqref{eq:undefined_window_iid} at the same marginal rate. The growing gap is the real-data signature of the same clustering mechanism Remark~\ref{rem:bursty_undefined} quantifies on the synthetic bursty generator.}
\label{tab:undefined_realdata}
\end{table}

\subsection{Computational Throughput and Memory Footprint}
\label{sec:exp7}

Propositions~\ref{prop:complexity} and~\ref{prop:complexity_window} claim an update cost that does not depend on the stream length for both operators, with constant memory for fading and memory growing with $W$ for the window. Table~\ref{tab:exp7} gives the median of seven repetitions on S9 for fading and for windows from $W=100$ to $W=100{,}000$. The time per instance of the window operator is about $1.1$ to $1.3$ times that of fading and does not grow with $W$ over three orders of magnitude, while its memory grows linearly with $W$. The measured memory is about five times the analytic buffer size (two vectors of $m$ doubles per stored instance) because of Python object overhead. The evaluators measured here are those of the package (\S\ref{sec:software}), which validates its input at every update. Absolute times depend on the hardware; the claim is carried by the independence of $W$ for time and the proportionality to $W$ for memory.

\begin{table}[!ht]
\centering
\footnotesize
\setlength{\tabcolsep}{3pt}\fitbox{\input{tables/tab_throughput.tex}}
\caption{Throughput and memory on S9 (300,000 instances, $m=2$, 7 repetitions). Analytic memory is the buffer of the window or the state of fading; measured memory is the allocation in use after filling the window, traced by \texttt{tracemalloc}.}
\label{tab:exp7}
\end{table}

%% ---------------------------------------------------------------------------
%% ---------------------------------------------------------------------------
\section{Software}
\label{sec:software}

The framework is implemented as the experimental subpackage \texttt{aggmetrics.stream} of the \texttt{aggmetrics} Python package for indicator-matrix evaluation~\citep{prati2026aggmetricspaper}, from version 0.4.0 (\url{https://github.com/rcprati/aggmetrics}; DOI of the version: \url{https://doi.org/10.5281/zenodo.24291757}). It is installed with \texttt{pip install aggmetrics}, needs only NumPy, and a notebook of the package (gallery notebook 17) shows its main functions on simulated streams. A \texttt{TemporalState} holds the state $\mathbold{C}(t)$ under exponential fading, a sliding window or no forgetting, and evaluates any measure of the package's registry on it with the three aggregation operators, the zero-division policies of \S\ref{sec:zero_division} and the audit tuple $(\mathbold{u}, m_{\mathrm{active}})$; with a reference prior it returns the value of a measure at that prior, the quantity monitored in \S\ref{sec:monitoring}. A \texttt{ClassTimeState} gives the class-time estimates of Remark~\ref{rem:accumulated}. A \texttt{DelayedEvaluator} is Algorithm~\ref{alg:update} in both of its forms, with the interim term, the correction, the eviction and the window FIFO, and reports the verified and the interim support and the coverage $\kappa_j$ of \S\ref{sec:zero_division}. \texttt{attribute} splits the change of a measure between two states into a concept and a prior part, by the counterfactual, the symmetric or the path-integrated Jacobian form of \S\ref{sec:jacobian_decomposition}, and triages an alarm. The closed forms of the paper (the undefined-window probability of Proposition~\ref{prop:undefined_window}, the window size of Corollary~\ref{cor:window_sizing}, the queue bound of Eq.~\eqref{eq:truncation_bound} and the factor $\kappa(\alpha)$ of Remark~\ref{rem:autocorr}) are in \texttt{stream.sizing}, so that a window or a fading factor can be sized before deployment. The chi-square monitor of \S\ref{sec:monitoring} and the calibration of its threshold are included and marked experimental, with the limits stated there.

\begin{small}
\begin{verbatim}
ev = DelayedEvaluator(2, alpha=0.98, eps=1e-3)  # Algorithm 1, fading
k = ev.arrive(proba, y_pred)        # interim term, product t-norm
ev.label(k, y_true)                 # correction when the label comes
ev.compute("recall")                # value under the verified support
ev.coverage()                       # verified share of each class
attribute("f1", ref.snapshot(), cur.snapshot()).triage()
\end{verbatim}
\end{small}

The states are tested against the package's batch evaluation on the same data, for the cumulative and the window states, against the decomposability of Theorem~\ref{thm:temporal_decomposability} for every decomposable measure, against the values of the running example of Appendix~\ref{app:running}, and against simulation for the closed forms. It has full branch coverage. The experiments of \S\ref{sec:experiments} run on the package: the evaluators of their scripts are thin adapters over it. A class with less than $10^{-12}$ of the mass of the state is reported as undefined, since a ratio of such masses is a number made of rounding residue. The subpackage is experimental, and its interface may change in a later release.

\section{Conclusions and Future Directions}
\label{sec:conclusions}

We extended the indicator-matrix framework of classification evaluation~\citep{prati2026unified} to data streams. The temporal operator $\Upsigma_{\mathrm{t}}$ gives exponential-fading and sliding-window evaluation with an update cost that does not depend on the stream length, and linear decomposability between pooled and per-class measures holds under it. Delayed and selective feedback form an observation mask whose four corners are the four label-availability settings of the literature. An interim estimate built with the product $t$-norm fills the gap left by missing labels, and its bias is linear in the calibration error of the estimate when labels are not delayed. Zero-division voids are handled by audited conventions. Prior shift and concept drift are separated by monitoring prior-free coordinates and attributing a metric change with a first-order Jacobian decomposition.

The experiments changed some of our expectations. Detectors on class-time coordinates raise no excess false alarms under label shift on synthetic streams, on four frozen-classifier real streams and with several classes, while the Jacobian alarm we first proposed as a detector, and DDM on the error indicator, do. The Jacobian decomposition is useful for attribution and for triage of alarms, not as a general detector. The one-vs-rest specificity is not prior-free with several classes, so the confusion rows are the coordinates to monitor. The smoothness advantage we expected from excluding unsupported classes depends on the stream, and what we claim for the zero-division policies is that, with the verified support, they report no value for a class that has none. The correction of selection bias is only partial when the delay depends on the label and, on the real streams tested, for precision, specificity and $F_1$ on NSL-KDD and for recall on Credit Card with delay.

The evidence has limits. The detection and attribution studies use simulated streams, and frozen classifiers on real streams with the label shift imposed by resampling; in native order the real drift of $P(X \mid Y)$ cannot be separated from the effect under test. A calibrated threshold that depends on $\alpha$ and on the stream length is needed for a given run-level false alarm, and the power for a recall change is low at low prior. The variance model of the detector assumes independent arrivals, which holds for the resampled streams and not for native order, where Remark~\ref{rem:autocorr} measures an inflation that the detector does not correct. The sliding-window form of the interim estimate is run on one synthetic stream only, the eviction step of Algorithm~\ref{alg:update} on that stream and on two real ones, the interim estimate on the synthetic streams is a noisy oracle and not a learned model, and the real-stream delay and censoring experiments simulate both on real features and labels. Future work includes interim estimators that adapt to calibration drift, tracked online with class-conditional calibration errors. Twenty-five statements of the paper were checked in Lean~4 against Mathlib. Each covers the algebraic or analytic core of the statement, with probabilistic steps taken as hypotheses or left out. They are the decomposability of Theorem~\ref{thm:temporal_decomposability}, for any nonnegative temporal weighting, including the commutation of $\Upsigma_{\mathrm{t}}$ with the selection matrices and the classes with a zero denominator; the fading and sliding-window recurrences (Eqs.~\eqref{eq:fading_update} and~\eqref{eq:window_update}); the exact split of a change of the pooled measure into an intrinsic and a prevalence part (Eq.~\eqref{eq:drift_decomposition}); the marginal consistency of the product $t$-norm (Eqs.~\eqref{eq:row_marg} and~\eqref{eq:col_marg}), with the marginal defects of the minimum and the \L{}ukasiewicz $t$-norm on the soft prediction of the running example (Example~\ref{ex:running_delay}) and their agreement with the product on hard predictions; the bias identity of Eq.~\eqref{eq:bias} and its zero expectation under calibration; the invariant of Algorithm~\ref{alg:update} for exponential fading, which is the unified state of Eq.~\eqref{eq:unified_cm} without eviction and, with eviction, that state minus the discarded corrections, whose effect on each entry is bounded by the tail mass of Eq.~\eqref{eq:truncation_bound} and by $\varepsilon$ per discarded correction; the bound on the queue, with the explicit age $s^\ast$ above; the independence of the column sums of the unified state from the stand-in; the audit property of the policies $\mathtt{exclude}$ and $\mathtt{nan}$ (no number for a class without support), with the value of $\mathtt{zero}$ as the $\mathtt{exclude}$ average shrunk by the share of supported classes; the cell-ratio lemma and Proposition~\ref{prop:prior_invariance}, with the recall and the specificity as integrals of the decision function against the class-conditional laws, free of the prior, and the dependence of the precision on it, shown on a numerical example; the undefined-window probability of Proposition~\ref{prop:undefined_window}, exact for the sliding window and bounded on both sides under fading, derived from the independence of the arrivals (Mathlib's \texttt{iIndepSet}), with the two ages of the fading bounds taken through their defining inequalities; the latency corollary (Corollary~\ref{cor:undefined_latency}), with its value for a constant delay and the window size it requires; the multiclass invariance of Proposition~\ref{prop:multiclass_invariance}, with the specificity formula and the values of the running example; the window-sizing rule of Corollary~\ref{cor:window_sizing}; the expectation and the variance of the fading estimator in Proposition~\ref{prop:variance_scaling}, for every $t$ and from the independence of the entries, with their limits; the variance of the fading estimator for autocorrelated entries (Remark~\ref{rem:autocorr}), exact for every $t$ and converging to Eq.~\eqref{eq:variance_inflation}, with $\kappa=1$ for independent entries and $\kappa=(1+\alpha r)/(1-\alpha r)$ for $\rho_d=r^d$; the chain-rule form of the decomposition of Eq.~\eqref{eq:jacobian_decomposition}; the closed-form Jacobians of the precision, $F_1$ and the accuracy and the exactness of the mid-point rule for the accuracy (\S\ref{sec:jacobian_decomposition}); and the state after an abrupt drift with its half-life, the tracking lag under a steady velocity and the damping of a transient outlier (\S\ref{sec:drift_taxonomy}). The values of the running example are also recomputed in Lean. The bursty extension of the undefined-window probability (Remark~\ref{rem:bursty_undefined}), the sliding-window form of the algorithm with eviction, the complexity bounds, the Jacobian of the balanced accuracy, the error of the mid-point rule for $F_1$ and the precision, and the gradual and reoccurring drift profiles are not formalized. The formalization is at \url{https://github.com/rcprati/aggmetrics-datastream-lean}, and the implementation of \S\ref{sec:software} is in the \texttt{aggmetrics} package, version 0.4.0 or later~\citep{prati2026aggmetricspaper}.

%% ---------------------------------------------------------------------------

%% ---------------------------------------------------------------------------
\section*{Statements and Declarations}

\textbf{Funding.} This work was supported by the S\~ao Paulo Research Foundation (FAPESP) and the National Council for Scientific and Technological Development (CNPq, grant 310534/2025-9).

\textbf{Competing interests.} The authors have no relevant financial or non-financial interests to disclose.

\textbf{Ethics approval and consent to participate.} Not applicable. The study uses public benchmark data sets and simulated streams, with no human participants or animals.

\textbf{Data availability.} All data sets are public: Electricity (Elec2)~\citep{harries1999splice}, Airlines~\citep{vanschoren2014openml}, Credit Card Fraud~\citep{dalpozzolo2015calibrating}, NSL-KDD~\citep{tavallaee2009detailed} and Covertype (distributed with scikit-learn). The synthetic streams are generated by the code described under Code availability.

\textbf{Code availability.} The experiment scripts, the tables, the figures and the generated result files are at \url{https://github.com/rcprati/data-stream-evaluation}; the real data sets are fetched by a script of the repository. The Lean~4 formalization is at \url{https://github.com/rcprati/aggmetrics-datastream-lean}. The implementation of \S\ref{sec:software} is in the \texttt{aggmetrics} package (\url{https://github.com/rcprati/aggmetrics}), version 0.4.0 or later, released on GitHub and archived on Zenodo (\url{https://doi.org/10.5281/zenodo.24291757})~\citep{prati2026aggmetricspaper}.

\textbf{Author contributions.} Ronaldo C. Prati: conceptualization, methodology, formal analysis, software, investigation, data curation, validation, visualization, writing -- original draft, writing -- review and editing, project administration. Elaine Ribeiro de Faria: conceptualization, validation, writing -- review and editing. All authors approved the final manuscript.

\textbf{Use of generative AI.} The authors used Claude (Anthropic), through Claude Code, to help write and review code, run and analyse the experiments, develop the Lean~4 formalization, and draft and revise parts of the text. The authors specified the research questions and the design, reviewed and verified the outputs, and take full responsibility for the content.

\bibliography{refs}

\begin{appendices}
\section{Notation}
\label{app:notation}

\begin{table}[!ht]
\centering
\small
\caption{Summary of notation used throughout the paper.}
\label{tab:notation}
\begin{tabular}{@{}p{3cm}p{9.2cm}@{}}
\toprule
\multicolumn{2}{@{}l}{\textit{Stream and instances}} \\
\midrule
$t$ & Discrete evaluation time step, $t \in \{1,2,\dots\}$ \\
$k$ & Index of a specific instance's arrival/generation time (may differ from the current evaluation time $t$) \\
$m$ & Number of classes \\
$n$ & Number of instances in a batch (stationary setting only) \\
$\mathbold{x}_t, \mathbold{y}_t, \mathbold{p}_t$ & Feature vector, true label vector, and prediction vector at time $t$ \\
$\mathbold{e}_j$ & $j$-th standard basis (one-hot) vector \\
$\Delta^{m-1}$ & Probability simplex in $\mathbb{R}^m$ \\
\midrule
\multicolumn{2}{@{}l}{\textit{Temporal aggregation ($\Upsigma_{\mathrm{t}}$)}} \\
\midrule
$\Upsigma_{\mathrm{t}}$ & Temporal reduction operator (general; encompasses both realizations below) \\
$\Upsigma_{\mathrm{t}}^\alpha$ & Exponential-fading realization, fading factor $\alpha \in (0,1)$ \\
$\Upsigma_{\mathrm{t}}^W$ & Sliding-window realization, window size $W$ (a scalar, not the matrix $\mathbold{W}(t)$ below) \\
$\mathbold{W}(t)$ & Diagonal temporal weighting matrix, $\operatorname{diag}(w_1(t),\dots,w_t(t))$ \\
$w_k(t)$ & Weight assigned to instance $k$ at evaluation time $t$ \\
$\mathbold{C}(t)$ & Temporal state (contingency) matrix; $\mathbold{C}_\alpha(t)$, $\mathbold{C}_W(t)$ denote each realization \\
$n_{\mathrm{eff}} = 1/(1-\alpha)$ & Memory horizon of exponential fading (mean age of the weights, plus one); the effective sample size for variance is $(1+\alpha)/(1-\alpha)$ \\
\midrule
\multicolumn{2}{@{}l}{\textit{Base framework recap (\citealp{prati2026unified})}} \\
\midrule
$\mathbold{Y}, \hat{\mathbold{Y}}$ & Batch indicator matrices (labels, predictions), $n \times m$ \\
$\Sigma_1$ & Micro reduction: pool all cells globally, then evaluate $M$ once \\
$\Sigma_m$ & Macro reduction: evaluate $M$ on each class, then average \\
\midrule
\multicolumn{2}{@{}l}{\textit{Structural (per-class) projections (\S\ref{sec:ovr})}} \\
\midrule
$\mathbold{S}_j$ & Linear per-class selection matrix for class $j$ \\
$\mathbold{C}_j(t)$ & $2\times 2$ per-class contingency matrix, $\mathbold{S}_j\mathbold{C}(t)\mathbold{S}_j^T$ \\
$\mathrm{TP}_j(t), \mathrm{FP}_j(t)$ & Entries of $\mathbold{C}_j(t)$ (true/false positive counts for class $j$) \\
$\mathrm{FN}_j(t), \mathrm{TN}_j(t)$ & Entries of $\mathbold{C}_j(t)$ (false negative/true negative counts for class $j$) \\
\midrule
\multicolumn{2}{@{}l}{\textit{Latency, censoring, and interim estimation (\S\ref{sec:delayed_labels})}} \\
\midrule
$\Delta_k$ & Verification latency of instance $k$ (possibly random) \\
$\mathcal{O}_k(t)$ & Dynamic per-instance observation mask (calligraphic; unrelated to the $O(\cdot)$ complexity notation of Proposition~\ref{prop:complexity}) \\
$T_P$ & Product $t$-norm, $T_P(a,b)=a\cdot b$ \\
$A_k$ & Indicator that a label is ever supplied for instance $k$ \\
$\mathcal{E}_k$ & Eligibility flag: the evaluator does not know at arrival that no label will be requested \\
$\mathcal{Q}(t)$ & Verification queue of pending instances \\
$\tilde{\mathbold{y}}_k$ & Interim estimate of the label (the stand-in), $g_t(\mathbold{s}_k)$ \\
$g_t$ & Calibration map fitted on the labels that have arrived \\
$\varepsilon$ & Tolerance for eviction and for the support indicator under fading \\
$n_{\mathrm{ess}} = (1+\alpha)/(1-\alpha)$ & Effective sample size of exponential fading \\
\midrule
\multicolumn{2}{@{}l}{\textit{Zero-division auditing (\S\ref{sec:zero_division})}} \\
\midrule
$u_j(t)$ & Support indicator for class $j$ (equals $1$ iff class $j$ has non-zero support at $t$) \\
$m_{\mathrm{active}}(t)$ & Number of classes with non-zero support at $t$ \\
$u^{\mathrm{ver}}_j(t), u^{\mathrm{int}}_j(t)$ & Verified and interim support indicators of class $j$ (equal without latency) \\
$\kappa_j(t)$ & Coverage: share of the denominator of class $j$ that comes from verified instances \\
\midrule
\multicolumn{2}{@{}l}{\textit{Canonical coordinates and drift decomposition (\S\ref{sec:drift_decomposition})}} \\
\midrule
$\pi(t)$ & Scalar positive-class prior of a per-class table (not bold) \\
$\mathbold{\pi}(t)$ & Multiclass prior vector, $P_t(Y)$ (bold) \\
$A(t), B(t), C(t), D(t)$ & Per-class cell masses (TP, FP, FN, TN) (not the bold state matrix $\mathbold{C}(t)$ above) \\
$R_{11}(t), R_{00}(t)$ & Canonical coordinates, $=\mathrm{Recall}(t), \mathrm{Specificity}(t)$ \\
$M(t)$ & A generic composite performance metric \\
$\mathbold{J}_{\mathrm{concept}}(t), J_{\mathrm{prior}}(t)$ & Concept- and prior-sensitivity Jacobians of $M(t)$ \\
\bottomrule
\end{tabular}
\end{table}

\section{The Running Example in Full}
\label{app:running}

\begin{sloppypar}

All values below are produced by \texttt{experiments/running\_example.py} in exact fractions. The stream has $8$ instances, with true labels $A\,B\,A\,C\,B\,A\,B\,B$ and predictions $A\,B\,B\,C\,B\,A\,C\,B$. The label of instance $k$ arrives at time $k+\Delta_k$, that is at $t=(1,3,5,7,7,9,9,9)$, and instance $7$ is never labelled.

\emph{Temporal state (Example~\ref{ex:running}).} Under fading with $\alpha=\tfrac12$, $\mathbold{C}_\alpha(4)$ has rows $(\tfrac1{16},\tfrac14,0)$, $(0,\tfrac18,0)$, $(0,0,\tfrac12)$ and mass $\tfrac{15}{16}$. Under the window $W=4$, $\mathbold{C}_W(8)$ has rows $(\tfrac14,0,0)$, $(0,\tfrac12,\tfrac14)$, $(0,0,0)$.

\emph{Decomposition and the void (Examples~\ref{ex:running_decomp} and~\ref{ex:running_void}).} The per-class tables $(\mathrm{TP},\mathrm{FN},\mathrm{FP},\mathrm{TN})$ are $(\tfrac14,0,0,\tfrac34)$, $(\tfrac12,\tfrac14,0,\tfrac14)$, $(0,0,\tfrac14,\tfrac34)$ and the pooled table is $(\tfrac34,\tfrac14,\tfrac14,\tfrac74)$. Precision is $(1,1,0)$ per class, with weights $(\tfrac14,\tfrac12,\tfrac14)$, pooled and weighted macro $\tfrac34$, unweighted macro $\tfrac23$. Recall is $(1,\tfrac23,\text{undefined})$ with weights $(\tfrac14,\tfrac34,0)$, pooled and weighted macro $\tfrac34$; the macro recall is $\tfrac56$ (exclude), $\tfrac59$ (zero), undefined (nan); $\mathbold{u}=(1,1,0)$.

\emph{Delayed labels (Example~\ref{ex:running_delay}).} In the window, instance $5$ is verified, $6$ and $8$ are pending, and $7$ is never labelled. The verified part is $\tfrac1{32}(0,0,0;\,0,8,0;\,0,0,0)$, the interim part is $\tfrac1{32}(6,1,1;\,1,6,1;\,1,1,6)$, so that $\mathbold{C}(8)=\tfrac1{32}(6,1,1;\,1,14,1;\,1,1,6)$ with mass $1$; the oracle is $\tfrac1{32}(8,0,0;\,0,16,8;\,0,0,0)$ and the difference is $\tfrac1{32}(2,-1,-1;\,-1,2,7;\,-1,-1,-6)$, equal to the right-hand side of Eq.~\eqref{eq:bias}. Recalls (oracle, $\mathbold{C}(8)$, verified only) are $A$: $1$, $\tfrac34$, undefined; $B$: $\tfrac23$, $\tfrac78$, $1$; $C$: undefined, $\tfrac34$, undefined. For the label-side denominators the verified support is $(0,1,0)$ and the coverage $(0,\tfrac12,0)$; the macro recall under the verified support is $\tfrac78$ (exclude), $\tfrac7{24}$ (zero), undefined (nan). The column sums of $\mathbold{C}(8)$ and of the oracle are both $(\tfrac14,\tfrac12,\tfrac14)$. The marginal defect of the stand-in of instance $7$ is $(0,\tfrac32,1)$ for (product, minimum, \L{}ukasiewicz) against the soft prediction $(\tfrac18,\tfrac18,\tfrac34)$, and $(0,0,0)$ against the hard one. The correction for instance $6$ is $\tfrac1{32}(2,0,0;\,-1,0,0;\,-1,0,0)$ under the window and $\tfrac1{64}(2,0,0;\,-1,0,0;\,-1,0,0)$ under fading at $t=8$; the fading weights of instances $6$ and $8$ are $\tfrac18$ and $\tfrac12$, and the eviction age for $\varepsilon=\tfrac1{32}$ is $s^\ast=5$.

\emph{Label shift (Examples~\ref{ex:running_shift} and~\ref{ex:jacobian}).} With rows $(\tfrac34,\tfrac14,0)$,\allowbreak\ $(0,\tfrac34,\tfrac14)$,\allowbreak\ $(\tfrac14,0,\tfrac34)$ and the prior $(\tfrac12,\tfrac14,\tfrac14)\to(\tfrac12,\tfrac18,\tfrac38)$, the values of $(\mathrm{Recall}_A, \mathrm{Specificity}_A, \mathrm{Precision}_A, F_1^A)$ are $(\tfrac34,\tfrac78,\tfrac67,\tfrac45)$ before and\allowbreak\ $(\tfrac34,\tfrac{13}{16},\tfrac45,\tfrac{24}{31})$ after. The change of $F_1^A$ is $-\tfrac4{155}$, and the midpoint concept term is $-\tfrac{96}{3721}$ with prior term $0$. The accumulated recall of the example in Example~\ref{ex:running_shift} is $\tfrac58$.
\end{sloppypar}
\end{appendices}

\end{document}

%% file: tables/tab_phase.tex
\begin{tabular}{@{}llcccc@{}}
\toprule
Scenario & Phase & $F_1$ & Accuracy & Recall & Specificity \\
\midrule
A: pure prior shift & Balanced & $0.848\ (0.022)$ & $0.849\ (0.020)$ & $0.849\ (0.029)$ & $0.850\ (0.027)$ \\
 & Imbalanced & $0.355\ (0.073)$ & $0.851\ (0.019)$ & $0.841\ (0.099)$ & $0.851\ (0.019)$ \\
 & Change & $-0.493$ & $+0.001$ & $-0.008$ & $+0.002$ \\
\midrule
B: concept drift & Pre-drift & $1.000$ & $1.000$ & $1.000$ & $1.000$ \\
 & Post-drift & $0.397\ (0.055)$ & $0.760\ (0.024)$ & $1.000\ (0.000)$ & $0.738\ (0.026)$ \\
 & Change & $-0.603$ & $-0.240$ & $+0.000$ & $-0.262$ \\
\bottomrule
\end{tabular}

%% file: tables/tab_elec2.tex
\begin{tabular}{@{}lccc@{}}
\toprule
Series & Correlation with $\pi(t)$ & 95\% interval & Null band, $|\mathrm{corr}|$ ($p$) \\
\midrule
$F_1$ & $0.548$ & $[0.470, 0.651]$ & $0.13$ ($0.002$) \\
Accuracy & $0.446$ & $[0.364, 0.537]$ & $0.14$ ($0.002$) \\
Recall, time fading & $0.111$ & $[0.028, 0.218]$ & $0.15$ ($0.176$) \\
Specificity, time fading & $0.124$ & $[0.031, 0.209]$ & $0.18$ ($0.268$) \\
Recall, class-time & $0.138$ & $[0.049, 0.258]$ & $0.18$ ($0.158$) \\
Specificity, class-time & $0.123$ & $[0.026, 0.210]$ & $0.19$ ($0.303$) \\
\bottomrule
\end{tabular}

%% file: tables/tab_taxonomy.tex
\begin{tabular}{@{}lcccc@{}}
\toprule
Profile & Spec.\ before & Spec.\ after & $\Delta$ Spec. & $F_1$ after \\
\midrule
Abrupt & $1.000\ (0.000)$ & $0.738\ (0.026)$ & $0.262\ (0.026)$ & $0.397\ (0.055)$ \\
Gradual & $1.000\ (0.000)$ & $0.747\ (0.026)$ & $0.253\ (0.026)$ & $0.426\ (0.052)$ \\
Incremental & $1.000\ (0.000)$ & $0.749\ (0.026)$ & $0.251\ (0.026)$ & $0.444\ (0.055)$ \\
Reoccurring & $0.885\ (0.019)$ & $0.774\ (0.025)$ & $0.112\ (0.031)$ & $0.516\ (0.047)$ \\
\bottomrule
\end{tabular}

%% file: tables/tab_attribution.tex
\begin{tabular}{@{}llcccc@{}}
\toprule
Metric & Jacobian & Concept only & Prior only, base 0.5 & Prior only, base 0.2 & Both \\
\midrule
F1 & midpoint & $100\%$ & $31\%$ & $62\%$ & $66\%$ \\
 & path-integrated & $100\%$ & $88\%$ & $88\%$ & $98\%$ \\
Precision & midpoint & $86\%$ & $67\%$ & $83\%$ & $98\%$ \\
 & path-integrated & $84\%$ & $94\%$ & $92\%$ & $100\%$ \\
Accuracy & midpoint & $100\%$ & $100\%$ & $83\%$ & $100\%$ \\
 & path-integrated & $100\%$ & $100\%$ & $83\%$ & $100\%$ \\
\bottomrule
\end{tabular}

%% file: tables/tab_detect_synth.tex
\begin{tabular}{@{}lccccc@{}}
\toprule
Detector & Threshold & Null FA & Excess FA & Power, spec.\ $-0.2$ & Power, recall $-0.2$ \\
\midrule
Jacobian $z$-alarm & calibrated & $0.052$ & $+0.055$ (61/2) & $0.44$ & $0.53$ \\
$\chi^2$ on time-fading coordinates & calibrated & $0.052$ & $+0.064$ (73/0) & $0.74$ & $0.53$ \\
$\chi^2$ on class-time coordinates & calibrated & $0.054$ & $-0.002$ (3/6) & $0.86$ & $0.53$ \\
max-$z$ on class-time coordinates & calibrated & $0.053$ & $+0.001$ (3/9) & $0.87$ & $0.50$ \\
DDM on the error indicator & default & $0.065$ & $+0.045$ (8/0) & $0.93$ & $0.41$ \\
DDM stratified by class & default & $0.208$ & $-0.033$ (0/21) & $0.96$ & $0.79$ \\
\bottomrule
\end{tabular}

%% file: tables/tab_detect_real.tex
\begin{tabular}{@{}lccccc@{}}
\toprule
Detector & Excess FA (R1, R2, R4) & Excess FA (R5) & Power, R1 & Power, R2 & Power, R4 \\
\midrule
Jacobian $z$-alarm & $+0.054$ (6/12) & $+0.024$ (1/4) & $0.67$ & $0.30$ & $0.99$ \\
$\chi^2$ on time-fading coordinates & $+0.072$ (5/12) & $+0.115$ (4/4) & $0.97$ & $0.39$ & $0.99$ \\
$\chi^2$ on class-time coordinates & $-0.010$ (0/12) & $-0.013$ (0/4) & $0.98$ & $0.90$ & $0.97$ \\
max-$z$ on class-time coordinates & $-0.005$ (0/12) & $-0.019$ (0/4) & $0.99$ & $0.78$ & $0.97$ \\
\bottomrule
\end{tabular}

%% file: tables/tab_multiclass.tex
\begin{tabular}{@{}llcccc@{}}
\toprule
$m$ & Confusion & \multicolumn{2}{c}{Excess FA, $\alpha=0.98$} & \multicolumn{2}{c}{Excess FA, $\alpha=0.99$} \\
\cmidrule(lr){3-4}\cmidrule(lr){5-6}
 & & one-vs-rest & rows & one-vs-rest & rows \\
\midrule
3 & uniform & $-0.009$ & $-0.001$ & $-0.022$ & $-0.003$ \\
3 & mild & $+0.006$ & $-0.015$ & $+0.019$ & $-0.003$ \\
3 & cyclic & $+0.402$ & $-0.026$ & $+0.736$ & $-0.005$ \\
5 & uniform & $-0.001$ & $+0.007$ & $+0.006$ & $-0.001$ \\
5 & mild & $+0.064$ & $-0.023$ & $+0.131$ & $-0.004$ \\
5 & cyclic & $+0.330$ & $-0.006$ & $+0.487$ & $+0.014$ \\
\midrule
5 & NSL-KDD & $+0.075$ & $-0.005$ & $+0.318$ & $-0.007$ \\
\bottomrule
\end{tabular}

%% file: tables/tab_triage.tex
\begin{tabular}{@{}lcccc@{}}
\toprule
$\alpha$ & Prior-driven alarms suppressed & Concept alarms kept & Pooled accuracy & Balanced accuracy \\
\midrule
$0.95$ & $0.74$ & $0.96$ & $0.81$ & $0.85$ \\
$0.98$ & $0.93$ & $0.86$ & $0.90$ & $0.89$ \\
$0.99$ & $0.98$ & $0.89$ & $0.95$ & $0.94$ \\
\bottomrule
\end{tabular}

%% file: tables/tab_censor.tex
\begin{tabular}{@{}lcccc@{}}
\toprule
Estimator & (iii) audit by decision & (iii) audit by label & (ii) class-dependent delay & (iv) class-dependent delay \\
\midrule
Naive & $-0.434\ (0.062)$ & $-0.277\ (0.050)$ & $-0.003\ (0.066)$ & $-0.438\ (0.074)$ \\
IPW, true probability & $-0.030\ (0.141)$ & $-0.006\ (0.069)$ & --- & --- \\
IPW, decision only & --- & $0.227\ (0.056)$ & --- & --- \\
Corrected, calibrated & $0.013\ (0.025)$ & $0.031\ (0.024)$ & $0.055\ (0.033)$ & $0.094\ (0.039)$ \\
Corrected, Platt & $-0.018\ (0.035)$ & $0.036\ (0.028)$ & $0.047\ (0.039)$ & $0.066\ (0.049)$ \\
Corrected, raw & $0.027\ (0.026)$ & $0.042\ (0.024)$ & $0.075\ (0.035)$ & $0.122\ (0.042)$ \\
\bottomrule
\end{tabular}

%% file: tables/tab_real_delay.tex
\begin{tabular}{@{}lllcccc@{}}
\toprule
Stream & Delay & Estimator & Recall & Specificity & Precision & $F_1$ \\
\midrule
R3 & none & naive & $-0.426\ (0.041)$ & $0.221\ (0.031)$ & $-0.000\ (0.001)$ & $-0.000\ (0.002)$ \\
 &  & IPW & $-0.005\ (0.014)$ & $0.000\ (0.002)$ & $-0.000\ (0.001)$ & $-0.000\ (0.002)$ \\
 &  & corrected & $-0.010\ (0.007)$ & $-0.000\ (0.000)$ & $-0.001\ (0.000)$ & $-0.001\ (0.001)$ \\
 & Poisson 300 & naive & $-0.416\ (0.042)$ & $0.183\ (0.024)$ & $-0.000\ (0.001)$ & $-0.001\ (0.002)$ \\
 &  & IPW & $-0.004\ (0.013)$ & $0.000\ (0.002)$ & $-0.000\ (0.001)$ & $-0.000\ (0.002)$ \\
 &  & corrected & $-0.140\ (0.024)$ & $-0.001\ (0.000)$ & $-0.005\ (0.001)$ & $-0.010\ (0.002)$ \\
 & class-dependent & naive & $-0.422\ (0.038)$ & $0.196\ (0.023)$ & $-0.000\ (0.001)$ & $-0.000\ (0.002)$ \\
 &  & IPW & $-0.003\ (0.014)$ & $0.000\ (0.001)$ & $-0.000\ (0.001)$ & $-0.000\ (0.002)$ \\
 &  & corrected & $-0.137\ (0.021)$ & $-0.001\ (0.000)$ & $-0.005\ (0.001)$ & $-0.009\ (0.002)$ \\
\midrule
R4 (binary) & none & naive & $-0.235\ (0.007)$ & $0.036\ (0.004)$ & $-0.001\ (0.003)$ & $-0.133\ (0.005)$ \\
 &  & IPW & $-0.000\ (0.000)$ & $-0.000\ (0.002)$ & $-0.001\ (0.003)$ & $-0.000\ (0.002)$ \\
 &  & corrected & $-0.004\ (0.000)$ & $-0.074\ (0.003)$ & $-0.089\ (0.004)$ & $-0.049\ (0.002)$ \\
 & Poisson 300 & naive & $-0.248\ (0.011)$ & $0.029\ (0.004)$ & $-0.000\ (0.002)$ & $-0.143\ (0.008)$ \\
 &  & IPW & $0.000\ (0.001)$ & $-0.000\ (0.002)$ & $-0.000\ (0.002)$ & $-0.000\ (0.001)$ \\
 &  & corrected & $-0.004\ (0.000)$ & $-0.074\ (0.003)$ & $-0.090\ (0.004)$ & $-0.049\ (0.002)$ \\
 & class-dependent & naive & $-0.266\ (0.011)$ & $0.034\ (0.005)$ & $-0.001\ (0.003)$ & $-0.155\ (0.008)$ \\
 &  & IPW & $0.000\ (0.001)$ & $-0.000\ (0.002)$ & $-0.001\ (0.003)$ & $-0.000\ (0.001)$ \\
 &  & corrected & $-0.004\ (0.000)$ & $-0.075\ (0.003)$ & $-0.091\ (0.004)$ & $-0.049\ (0.002)$ \\
\bottomrule
\end{tabular}

%% file: tables/tab_window.tex
\begin{tabular}{@{}lcccccc@{}}
\toprule
 & \multicolumn{2}{c}{Recall bias, fading} & \multicolumn{2}{c}{Recall bias, window} & \multicolumn{2}{c}{Queue, mean (max)} \\
\cmidrule(lr){2-3}\cmidrule(lr){4-5}\cmidrule(lr){6-7}
Setting & Naive & Corrected & Naive & Corrected & Fading & Window \\
\midrule
(ii) delayed, full & $-0.003\ (0.045)$ & $0.002\ (0.031)$ & $0.001\ (0.062)$ & $0.002\ (0.037)$ & $24.9$ ($30$) & $24.9$ ($30$) \\
(iii) immediate, partial & $-0.445\ (0.063)$ & $0.012\ (0.023)$ & $-0.444\ (0.092)$ & $0.014\ (0.037)$ & $96.3$ ($125$) & $23.1$ ($35$) \\
(iv) delayed, partial & $-0.447\ (0.065)$ & $0.010\ (0.032)$ & $-0.444\ (0.100)$ & $0.012\ (0.044)$ & $115.3$ ($145$) & $42.1$ ($54$) \\
(v) delay 150, partial & $-0.442\ (0.074)$ & $0.010\ (0.051)$ & $-0.441\ (0.116)$ & $0.008\ (0.075)$ & $206.5$ ($240$) & $97.5$ ($100$) \\
\bottomrule
\end{tabular}

%% file: tables/tab_nslkdd.tex
\begin{tabular}{@{}lcccc@{}}
\toprule
 & \multicolumn{2}{c}{Native order, 10 seeds} & \multicolumn{2}{c}{Blocks of 1,000 permuted, 10 orders} \\
\cmidrule(lr){2-3}\cmidrule(lr){4-5}
Class & Difference [95\% interval] & Adaptive ahead & Difference [95\% interval] & Adaptive ahead \\
\midrule
normal & $+0.162$ $[+0.085, +0.234]$ & 8 of 10 & $+0.299$ $[+0.221, +0.364]$ & 9 of 10 \\
DoS & $+0.152$ $[+0.092, +0.212]$ & 10 of 10 & $+0.298$ $[+0.225, +0.356]$ & 10 of 10 \\
Probe & $+0.137$ $[+0.086, +0.183]$ & 10 of 10 & $+0.105$ $[+0.023, +0.186]$ & 7 of 10 \\
R2L & $+0.032$ $[-0.088, +0.142]$ & 6 of 10 & $-0.078$ $[-0.221, +0.071]$ & 2 of 10 \\
U2R & $+0.009$ $[-0.113, +0.138]$ & 5 of 10 & $-0.104$ $[-0.265, +0.051]$ & 4 of 10 \\
\bottomrule
\end{tabular}

%% file: tables/tab_support.tex
\begin{tabular}{@{}lcccccc@{}}
\toprule
Delay & $b$ & Formula & Empirical & Class absent & MAE, verified & MAE, interim \\
\midrule
immediate & $1.0$ & $0.046$ & $0.042$ & $0.042$ & $0.000$ & $0.000$ \\
constant 25 & $1.0$ & $0.166$ & $0.161$ & $0.042$ & $0.012$ & $0.006$ \\
constant 25 & $0.5$ & $0.412$ & $0.409$ & $0.042$ & $0.039$ & $0.006$ \\
Poisson 25 & $1.0$ & $0.167$ & $0.163$ & $0.042$ & $0.012$ & $0.006$ \\
Poisson 25 & $0.5$ & $0.413$ & $0.422$ & $0.042$ & $0.041$ & $0.006$ \\
constant 70 ($\ge$ W) & $1.0$ & $1.000$ & $1.000$ & $0.042$ & undefined & $0.006$ \\
\bottomrule
\end{tabular}

%% file: tables/tab_throughput.tex
\begin{tabular}{@{}lrrrrr@{}}
\toprule
Operator & $W$ & Time, median [min, max] (s) & Inst./s & Memory (analytic) & Memory (measured) \\
\midrule
Fading ($\alpha=0.999$) & --- & 1.10 [1.06, 1.13] & 271,572 & 32 B & 1,224 B \\
Window & 100 & 1.41 [1.30, 1.69] & 213,198 & 3,232 B & 18 KB \\
Window & 1,000 & 1.36 [1.28, 1.56] & 219,937 & 31 KB & 165 KB \\
Window & 10,000 & 1.29 [1.27, 1.36] & 232,948 & 313 KB & 1,644 KB \\
Window & 100,000 & 1.19 [1.19, 1.23] & 251,689 & 3,125 KB & 16,431 KB \\
\bottomrule
\end{tabular}